%% file: main.tex
\documentclass{article}

\usepackage[preprint]{neurips_2026}

\usepackage[utf8]{inputenc} % allow utf-8 input
\usepackage[T1]{fontenc}    % use 8-bit T1 fonts
\usepackage{hyperref}       % hyperlinks
\usepackage{url}            % simple URL typesetting
\usepackage{booktabs}       % professional-quality tables
\usepackage{amsfonts}       % blackboard math symbols
\usepackage{nicefrac}       % compact symbols for 1/2, etc.
\usepackage{microtype}      % microtypography
\usepackage{xcolor}     % colors
\usepackage{graphicx}
\usepackage{subcaption}
\usepackage{algorithmicx}
\usepackage{amsmath}
\usepackage{amssymb}
\usepackage{bbm}
\usepackage{booktabs}
\usepackage{multirow}
\usepackage{xspace}

\usepackage{algorithm}
\usepackage{algpseudocode}
\usepackage{caption}
\usepackage{wrapfig}
\usepackage{float}
\usepackage{tcolorbox}
\tcbuselibrary{skins, breakable}

\makeatletter
\newenvironment{breakablealgorithm}
  {% \begin{breakablealgorithm}
   \begin{center}
   \refstepcounter{algorithm}
   \hrule height.8pt depth0pt \kern2pt
   \renewcommand{\caption}[2][\relax]{%
     {\raggedright\textbf{\ALG@name~\thealgorithm} ##2\par}%
     \ifx\relax##1\relax
       \addcontentsline{loa}{algorithm}{\protect\numberline{\thealgorithm}##2}%
     \else
       \addcontentsline{loa}{algorithm}{\protect\numberline{\thealgorithm}##1}%
     \fi
     \kern2pt\hrule\kern2pt
   }}
  {% \end{breakablealgorithm}
   \kern2pt\hrule\relax
   \end{center}
  }
\makeatother

\usepackage{fontawesome5}
\usepackage{array}

\newtcolorbox{exampleblock}{
    colback=lightgrayblue,
    colframe=black!20,
    boxrule=0.4pt,
    arc=2pt,
    left=4pt,
    right=4pt,
    top=3pt,
    bottom=3pt,
    boxsep=2pt,
    before skip=0.4em,
    after skip=0.4em
}

\usepackage[toc,page]{appendix}
\usepackage{titletoc}
\newcommand{\method}{\textsc{MoDA}\xspace}
\newcommand{\methodcolor}{\textcolor{themeblue}{\textsc{M}}\textcolor{themeblue2}{\textsc{O}}\textcolor{themeblue3}{\textsc{D}}\textcolor{themeblue4}{\textsc{A}}\xspace}

\newcommand{\ssot}{\text{SSoT}}
\usepackage{pifont}
\newcommand{\cmark}{\ding{51}}
\newcommand{\xmark}{\ding{55}}

\definecolor{lightgrayblue}{rgb}{0.961,0.973,0.988}
\definecolor{darkyellow}{rgb}{0.980, 0.65, 0}
\definecolor{themeblue}{rgb}{0.400, 0.600, 1.000}
\definecolor{themeblue2}{rgb}{0.102,0.459,1.000}
\definecolor{themeblue3}{rgb}{0.000,0.200,0.800}
\definecolor{themeblue4}{rgb}{0.000, 0.200, 0.600}
\definecolor{darkred}{rgb}{0.800, 0.000, 0.000}

\newcommand\liweiaddressed[1]{}

\newcommand{\iconstar}{{\raisebox{0.2ex}{\scalebox{0.7}{\faStar}}}}
\newcommand{\iconfire}{{\raisebox{0.2ex}{\scalebox{0.7}{\faHotjar}}}}

\newcommand{\github}{\raisebox{-1.5pt}{\includegraphics[height=1.05em]{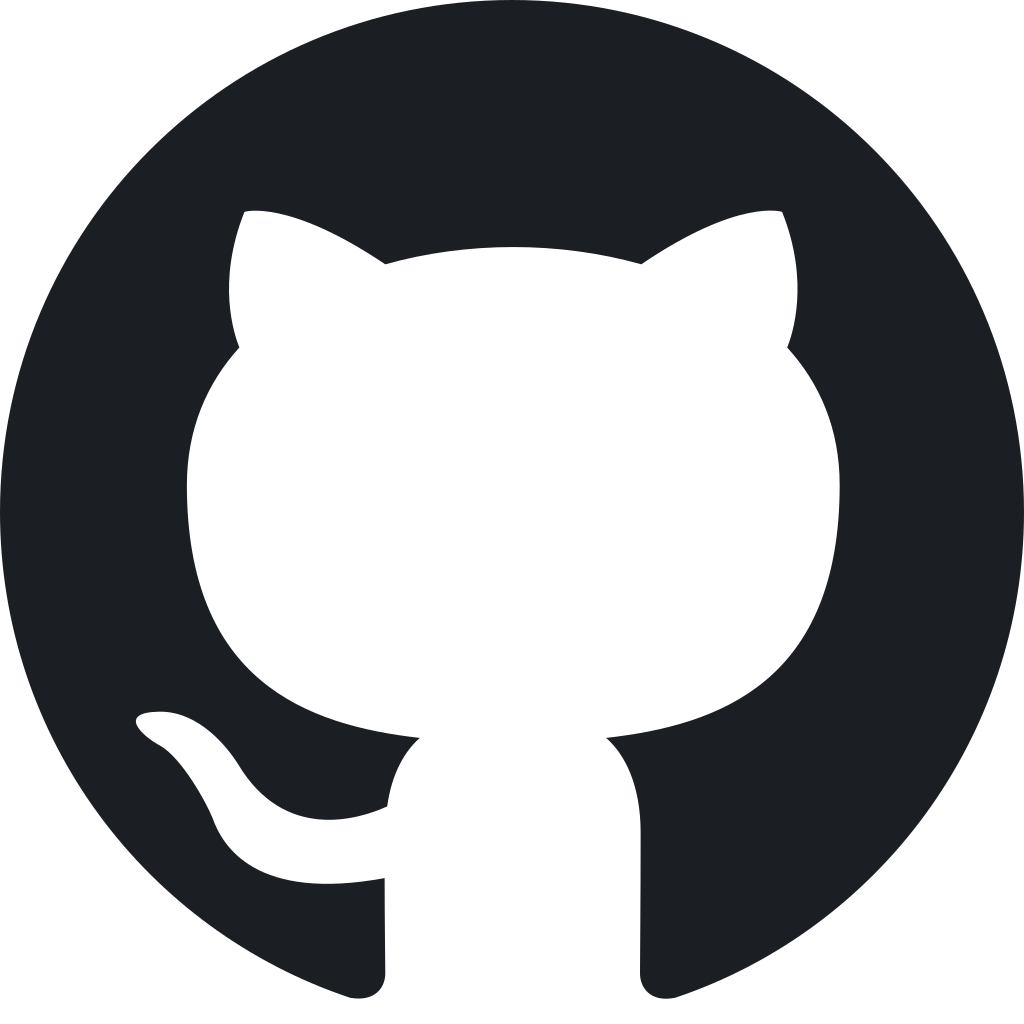}}\xspace}

\title{Forty Shades of \textcolor{themeblue}{B}\textcolor{themeblue2}{l}\textcolor{themeblue3}{u}\textcolor{themeblue4}{e}: Quality-Diversity Alignment \\ via Mode-Conditioned Reinforcement Learning}

\author{ \begin{tabular}{c} \textcolor{themeblue}{\iconstar} Jiayi Yuan$^{1}$ \quad \textcolor{themeblue2}{\iconstar} Hangoo Kang$^{2}$ \quad \textcolor{themeblue3}{\iconstar} James Jihao Liu$^{2}$ \\ [0.2em] Yejin Choi$^{2}$ \quad Vikram Iyer$^{1}$ \quad \textcolor{darkred}{\iconfire} Liwei Jiang$^{1}$ \quad \textcolor{darkred}{\iconfire} Natasha Jaques$^{1}$  \end{tabular} \\ [1.4em] $^{1}$University of Washington \quad $^{2}$Stanford University \\ [0.4em]\texttt{yuancarrieyjy@cs.washington.edu} \quad  \texttt{\{hangook,jihaoliu\}@stanford.edu} \\ [0.4em] \textcolor{themeblue4}{\iconstar} Equal first author \quad \textcolor{darkred}{\iconfire} Equal senior author \\ [0.4em] \github Code: \url{github.com/yuanjiayiy/mode-conditioned-diversity-alignment}}
\begin{document}

\maketitle

\input{notes_arxiv/0_abstract}
\input{notes_arxiv/1_introduction}

\input{notes_arxiv/6_related_work}
\input{notes_arxiv/3_method}

\input{notes_arxiv/4_experiments}
\input{notes_arxiv/5_discussions}
\input{notes_arxiv/z_acknowledgement}

\newpage
\bibliography{reference}
\bibliographystyle{plain}
\input{notes_arxiv/appendix}

\end{document}

%% file: notes_arxiv/0_abstract.tex
\begin{abstract}

A notable byproduct of LLM alignment training is \textit{mode collapse}: the progressive loss of output diversity that narrows a model's expressivity at inference time. This degradation is especially limiting for applications requiring open-ended exploration and pluralistic perspectives, such as scientific ideation and creative writing. We present \textbf{\methodcolor}~(\textbf{MO}de-conditioned \textbf{D}iversity \textbf{A}lignment), an online post-training RL algorithm that jointly optimizes generation quality and diversity, inspired by the coordination perspective in multi-agent reinforcement learning (MARL). \method trains a single shared LLM policy conditioned on abstract numbered mode tokens, where each mode acts as an agent that competes to produce outputs distinct from the others. This MARL-inspired formulation encourages mode-conditioned agents to explore complementary regions of the high-quality output space without requiring hand-crafted personas or architectural modifications. \method{} employs a prompt-adaptive quality gating mechanism that calibrates quality thresholds against a frozen reference policy and grants diversity rewards only to responses that meet them, preventing reward-hacking behaviors such as language switching and verbosity.
To study quality-diversity tradeoffs, we evaluate \method{} on a comprehensive benchmark spanning seven general capability tasks and four domain-specific diversity tasks in scientific ideation and creative writing. \method{} improves SBERT diversity by \textbf{265\%} on the Infinite-Chat held-out prompts while increasing general capability performance by \textbf{10.3\%} over the Qwen3-8B baseline. Compared with the strongest DivPO baseline, \method{} improves SBERT diversity from 0.274 to 0.482 (\textbf{+75.9\%}) and E-Vendi from 2.86 to 4.4 (\textbf{+53.8\%}), while improving average general capability pass@1 by \textbf{7.0\%}. Overall, \method{} provides a drop-in alternative to standard post-training methods that preserves and expands the model's expressive output space while improving quality.

\end{abstract}

%% file: notes_arxiv/1_introduction.tex
\begin{figure*}[t!]
    \centering
    \begin{subfigure}[t]{0.78\linewidth}
        \centering
        \includegraphics[width=\textwidth]{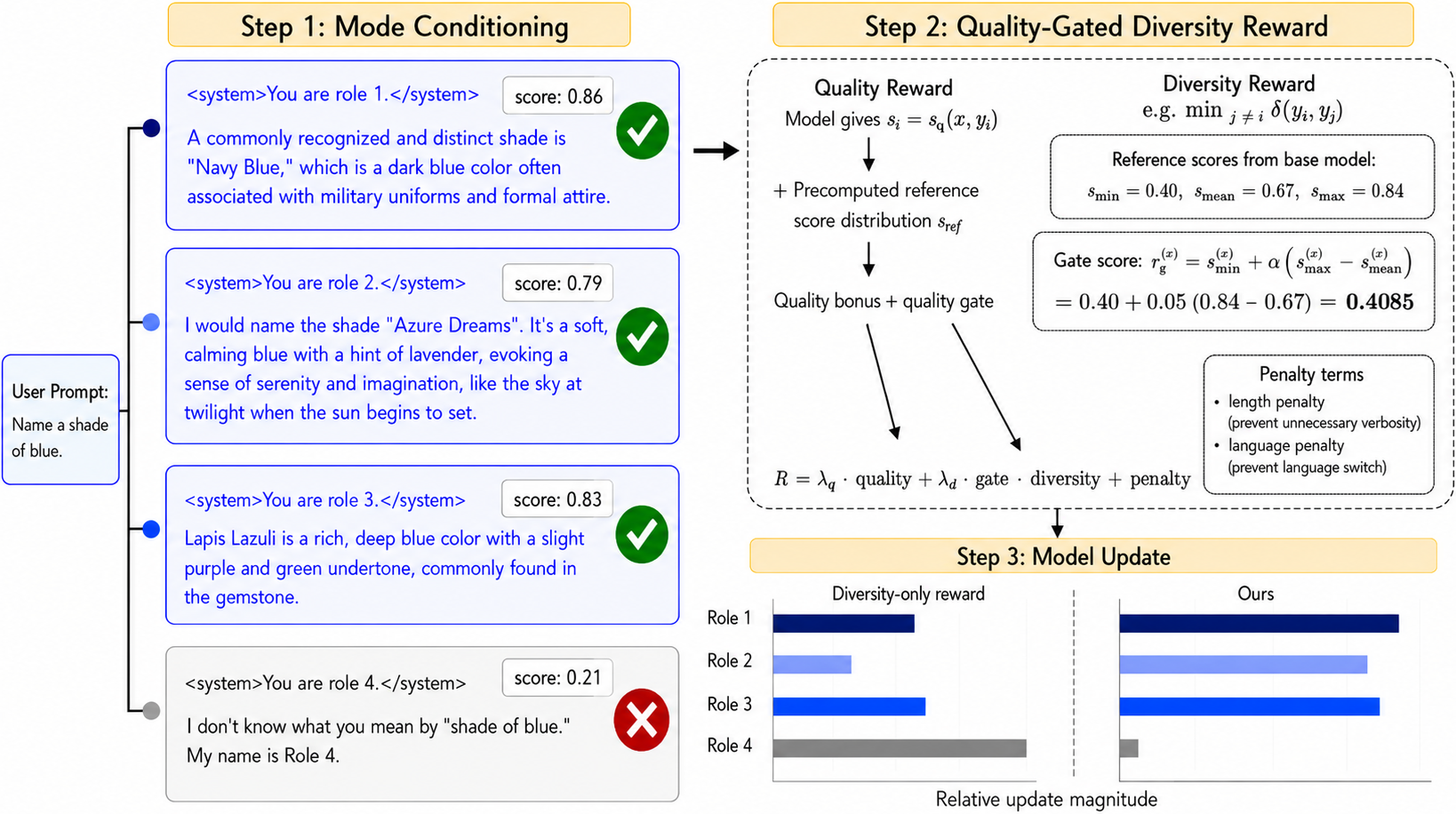}
        \label{fig:architecture}
    \end{subfigure}\hfill
    \begin{subfigure}[t]{0.2\linewidth}
        \centering
        \includegraphics[width=\linewidth]{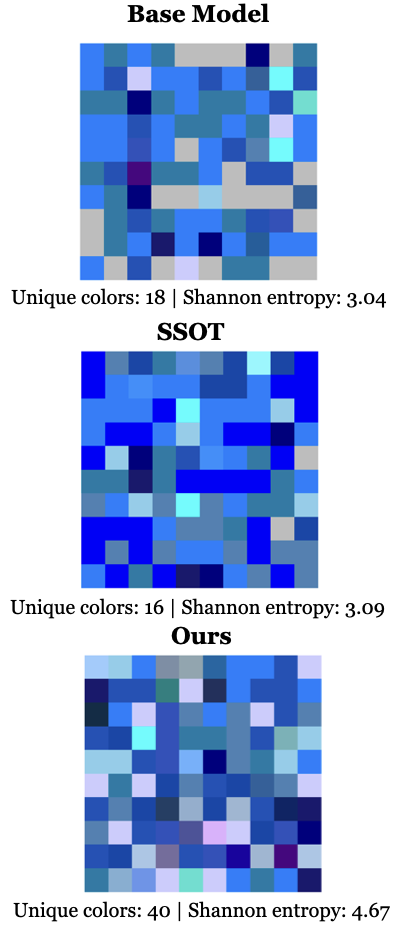}
        \label{fig:40blue}
    \end{subfigure}
    \vspace{-0.2cm}
\caption{\textbf{Left}: Overview of \method. \method{} uses mode conditioning to elicit diverse candidate responses, then applies a quality-gated diversity reward with penalty terms to promote high-quality diversity. Each candidate is labeled with its quality score $s_i$ relative to the prompt-adaptive quality threshold $\tau_q^{(x)}$. Candidates with $s_i<\tau_q^{(x)}$ have $g(x,y_i)=0$ and therefore receive no diversity credit. Step 3 shows the resulting signed group-relative advantages $\widehat{A}_i$; positive advantages promote a response, whereas negative advantages suppress it during the policy update. Matching colors track the same response through Steps 1--3. \textbf{Right}: 100 responses to the prompt \textbf{``Name a shade of blue''} from the Qwen3-8B initial policy (\texttt{Qwen/Qwen3-8B}), String Seed of Thought (\ssot)~\citep{Misaki2026}, and \method{} under diverse-mode inference. \method{} produces 40 distinct shades of blue, 122\% more than the Qwen3-8B initial policy and 150\% more than \ssot. \liweiaddressed{It's unclear what the two bar charts of Model update mean? I think the issue is that, it's not intuitive which of the four generations in Step 1 is considered low quality, and hence needs penalty. One possibility is to add in the quality score (mock-up is fine), to the upper-right corner of each generation box, plus, provide a reference score. So that it makes it more clear which response should be suppressed during model update. Also the four bars in Step 3 seem to have different colors compared to the four boxes in Step 1. }
}
\label{fig:moda + 40blue}
\end{figure*}

\section{Introduction}
\label{sec:introduction}

\begin{quote}
\itshape
``In the course of evolution nature has gone to endless trouble to see that every individual is unlike every other individual.''

\hfill --- Aldous Huxley, \textit{Brave New World Revisited}
\end{quote}

Post-training alignment is essential for improving the capability and safety of large language models (LLMs), but it also introduces a concerning side effect: \textit{mode collapse}. Aligned models increasingly favor a narrow set of responses (the ``mode'') over the broader space of plausible outputs, resulting in a substantial loss of generation diversity~\citep{jiang2025artificialhivemindopenendedhomogeneity, west2025base, huang2024self, sourati2026homogenizingeffectlargelanguage}. Yet in many domains, diversity is not merely desirable but essential. In AI-assisted scientific discovery, progress relies on exploring a broad and plausible hypothesis space rather than prematurely converging on a single line of inquiry~\citep{gruver2023proteindesignguideddiscrete, romera2024mathematical}. Likewise, generative diversity is critical for applications requiring creativity and open-ended exploration~\citep{Shumailov2024AIMC, abdulhai2026llmsdistortwrittenlanguage}. While inference-time interventions~\citep{Misaki2026, Zhang2025} can partially mitigate this loss of diversity, a more fundamental solution is to redesign post-training algorithms so that expressivity is preserved throughout the alignment process by construction.

We introduce \method~(\textbf{MO}de-conditioned \textbf{D}iversity \textbf{A}lignment), an online post-training RL algorithm that jointly optimizes the quality and diversity of generated responses by formulating alignment through the lens of multi-agent reinforcement learning (MARL), where agents develop diverse behaviors through competition and coordination~\citep{eysenbach2018diversityneedlearningskills,liang2024learning,li2021celebratingdiversitysharedmultiagent}. \method{} introduces \textit{role conditioning} to instantiate multiple lightweight agents within a single shared policy. Specifically, we prepend abstract role tokens to the system prompt, enabling each role to generate a response to the same user query while independently optimizing a group-relative diversity reward. This creates competitive reward dynamics among roles, encouraging them to specialize in distinct regions of the high-quality response space and collectively produce a diverse set of outputs.

A key design choice of \method{} is a \textit{prompt-adaptive quality gate} that grants diversity rewards only to responses exceeding a prompt-specific quality threshold computed from a frozen reference policy. Among responses that satisfy the gate, diversity is rewarded according to semantic distance from the nearest neighboring response, encouraging exploration within the space of sufficiently high-quality generations. By coupling diversity rewards with quality, this mechanism prevents reward hacking, in which models maximize diversity by producing unusual but irrelevant outputs~\citep{li2025jointlyreinforcingdiversityquality,wan2025enhancingpersonalizedmultiturndialogue}.

Assessing quality-diversity tradeoffs requires evaluating both standard capabilities and applications that benefit from diverse outputs. To this end, we curate a comprehensive 11-task evaluation suite that jointly measures general capability retention and diversity-focused applications where multiple high-quality generations are useful or necessary. The suite includes seven general capability benchmarks (GSM8K, MMLU, GPQA, BoolQ, HellaSwag, TruthfulQA, and IFEval) and four open-ended domain application benchmarks spanning scientific ideation and creative writing such as HypoBench~\citep{liu2026hypobenchsystematicprincipledbenchmarking}, PreScience~\citep{Ajith2026}, NoveltyBench~\citep{zhang2025noveltybenchevaluatinglanguagemodels}, and a held-out split of Infinite-Chat~\citep{jiang2025artificialhivemindopenendedhomogeneity}. We evaluate output diversity using complementary metrics: \textit{semantic-level} (SBERT), \textit{entropy-level} (E-Vendi), and \textit{entailment-level} (a learned pairwise discriminator). Under the same thinking-disabled setting, \method{} improves SBERT diversity and E-Vendi by 155.5\% and 94.0\%, respectively, over the Qwen3-8B initial policy on the diversity-focused benchmark suite, while simultaneously increasing average pass@1 by 10.3 percentage points over the initial policy, and by 7.0 percentage points over the strongest DivPO baseline on general capability benchmarks.

Compared with prior diversity-oriented post-training methods~\citep{li2025jointlyreinforcingdiversityquality, lanchantin2025diverse, chung2025modifyinglargelanguagemodel, Chen2025PosttrainingLL, slocum2025diversepreferencelearningcapabilities, Puri2026}, \method{} conditions diversity on an explicit mode token rather than baking it unconditionally into the model. As a result, it naturally supports two inference modes: supplying mode tokens yields a diverse set of responses, while omitting them recovers the base model's standard behavior, preserving single-response capability. More broadly, we hope \method{} motivates future work that brings richer MARL principles into LLM post-training.

%% file: notes_arxiv/6_related_work.tex
\input{notes_arxiv/table/related_works}

\section{Related work}
\label{sec:related_work}

\paragraph{Why Does Generation Diversity Matter?}
Language models often collapse toward a narrow set of high-probability responses, resulting in an ``Artificial Hivemind'' characterized by substantial intra- and inter-model homogeneity on open-ended prompts \cite{jiang2025artificialhivemindopenendedhomogeneity}. Zhang et al.~\cite{Zhang2025} further show that post-training alignment reduces diversity through biases in preference data, limiting performance in domains that require diverse outputs.
In scientific reasoning, automated research agents benefit from a wider exploration breadth \citep{zou2026fmlbenchbenchmarkingmachinelearning}, and mode collapse causes reduced exploration in solution spaces \citep{yuan2026agenticredevolvingagenticsystems}. In creative tasks, reduced diversity in LLM outputs also diminishes the creativity of LLM-assisted writing \citep{abdulhai2026llmsdistortwrittenlanguage,Anderson2024HomogenizationEO,padmakumar2023does}. In mathematical proofs, strategy diversity is critical since repeated sampling is only useful when samples explore distinct reasoning paths \citep{Wu2025, Cao2025TowardsAM}. These findings motivate methods that preserve quality while expanding the range of model outputs.

% Language models often collapse toward a narrow set of high-probability responses. Jiang et al.~\cite{jiang2025artificialhivemindopenendedhomogeneity} characterize this as an ``Artificial Hivemind,'' demonstrating substantial intra- and inter-model homogeneity on open-ended prompts, while Zhang et al.~\cite{Zhang2025} show that post-training alignment further amplifies this effect through biases in preference data. This lack of diversity is particularly problematic in domains requiring broad exploration. Diverse outputs improve exploration in scientific reasoning and research agents \citep{Zou2025, yuan2026agenticredevolvingagenticsystems}, enhance creativity in LLM-assisted writing \citep{abdulhai2026llmsdistortwrittenlanguage,Anderson_2024,padmakumar2023does}, and increase the effectiveness of repeated sampling for mathematical reasoning by exploring distinct proof strategies \citep{Wu2025, Cao2025TowardsAM}. These findings motivate methods that preserve quality while expanding the diversity of model outputs.

\paragraph{Measuring Diversity of LLMs.}
Diversity of LLM-generated content has been measured along several dimensions. Lexical metrics include distinct n-grams \citep{ippolito2019comparison}, Self-BLEU \citep{zhu2018texygen}, Measure of Textual Lexical Diversity (MTLD) \citep{McCarthy2010MTLDVA}, and compression-based homogeneity scores \citep{shaib2024standardizing}. Semantic metrics build on sentence embeddings \citep{reimers2019sentencebertsentenceembeddingsusing, wieting2018paranmt} and information-theoretic constructions such as the Vendi Score \citep{friedman2023vendiscorediversityevaluation} and its conditional variant \citep{jalali2024conditionalvendiscoreinformationtheoretic}, as well as gradient-based measures like G-Vendi \citep{jung2025prismaticsynthesisgradientbaseddata}. A complementary line of work studies how diversity collapses through the training pipeline \citep{kirk2024understandingeffectsrlhfllm, omahony2024attributingmodecollapsefinetuning, padmakumar2023does, west2025base, dang2025assessing}, and recent benchmarks evaluate humanlike or effective semantic diversity directly \citep{zhang2025noveltybenchevaluatinglanguagemodels, shypula2025evaluatingdiversityqualityllm, guo2025benchmarking, Shahid2025}. However, most evaluations assess diversity or quality in isolation, whereas we jointly evaluate capability and diversity across real-world applications to ensure models preserve multiple plausible answers without mode collapse \citep{Misaki2026,Puri2026} or catastrophic forgetting \citep{luo2025empiricalstudycatastrophicforgetting}.

% \liwei{Can we tighten the following paragraph a bit more? It's a bit too long.}
\paragraph{Improving Diversity during Training and Inference.}
Recent work has explored improving diversity through prompting \citep{Zhang2025, lau2025dipperdiversitypromptsproducing, lagzian2025multinoveltyimprovediversitynovelty, wang2025multilingualpromptingimprovingllm, Kim2026, Wu2025}, decoding \citep{holtzman2020curiouscaseneuraltext, nguyen2024turningheatminpsampling, ruan-etal-2025-g2, wang2024lpoadaptivedecodinglatent}, and training \citep{li2025jointlyreinforcingdiversityquality, lanchantin2025diverse, chung2025modifyinglargelanguagemodel, Chen2025PosttrainingLL, slocum2025diversepreferencelearningcapabilities, Puri2026} (Table \ref{tab:related-work}). At inference time, Verbalized Sampling \citep{Zhang2025} prompts models to express a distribution over possible responses, while String Seed of Thought \citep{Misaki2026} injects random strings as entropy sources. DIPPER \citep{lau2025dipperdiversitypromptsproducing} and Multi-Novelty \citep{lagzian2025multinoveltyimprovediversitynovelty} construct diverse prompt ensembles, and multilingual, persona, or chain-of-thought prompting can elicit variations \citep{wang2025multilingualpromptingimprovingllm, Wu2025}. At decoding time, min-$p$ sampling \citep{nguyen2024turningheatminpsampling} adaptively truncates low-probability tokens based on model confidence, while G2 \citep{ruan-etal-2025-g2} and adaptive temperature methods \citep{wang2024lpoadaptivedecodinglatent} guide generation with auxiliary diversity modules or learned per-instance decoding parameters. In post-training, RL has emerged as a powerful method for enhancing diversity. Multi-answer RL \citep{Puri2026} trains models to output multiple diverse answers within one response; Soft Preference Learning decouples entropy from KL-penalty to improve lexical and semantic variety \citep{slocum2025diversepreferencelearningcapabilities}; DivPO \citep{lanchantin2025diverse} applies DPO to responses that are both diverse and exceed a quality threshold. 

Several recent works share our goal of jointly optimizing quality and diversity through RL. DARLING \citep{li2025jointlyreinforcingdiversityquality} balances quality and diversity by rewarding a multiplicative aggregation of quality and diversity metrics; DQO \citep{Chen2025PosttrainingLL} uses a determinant-based group diversity reward, but assigns the same reward to all responses in the group. GRPO-Unlikeliness \citep{he2025rewardingunlikelyliftinggrpo} and GAPO \citep{anschel2025group} both encourage diversity through frequency- and likelihood-based rewards. However, although these methods explore different reward designs, none of them allow a single model to switch between producing one high-quality answer and a diverse set of responses.
\method, on the other hand, approaches quality-diversity alignment from a MARL-inspired perspective, treating diverse generation as a competition-and-coordination problem among role-conditioned agents within a shared policy. This formulation enables explicit per-role credit assignment under quality constraints, encouraging specialization across the high-quality response space while keep the model's standard behavior intact. 

%% file: notes_arxiv/table/related_works.tex
\begin{wrapfigure}[18]{r}{0.45\textwidth}
\vspace{-1.2em}
\centering
\scriptsize
\setlength{\tabcolsep}{3pt}
\renewcommand{\arraystretch}{0.85}
\begin{tabular}{llc}
\toprule
\textbf{Method} & \textbf{Mechanism} & \textbf{Code} \\
\midrule
\multicolumn{3}{l}{\textsc{\underline{Inference-time}}} \\
Verbalized Sampling~\citep{Zhang2025} & Prompting & \cmark \\
String Seed of Thought~\citep{Misaki2026} & Prompting & \cmark \\
DIPPER~\citep{lau2025dipperdiversitypromptsproducing} & Prompting & \xmark \\
Multi-Novelty~\citep{lagzian2025multinoveltyimprovediversitynovelty} & Prompting & \xmark \\
Multilingual Prompting~\citep{wang2025multilingualpromptingimprovingllm} & Prompting & \cmark \\
Nucleus Sampling~\citep{holtzman2020curiouscaseneuraltext} & Decoding & \cmark \\
min-$p$ Sampling~\citep{nguyen2024turningheatminpsampling} & Decoding & \cmark \\
G2~\citep{ruan-etal-2025-g2} & Decoding & \xmark \\
Adaptive Decoding (LPO)~\citep{wang2024lpoadaptivedecodinglatent} & Decoding & \cmark \\
\midrule
\multicolumn{3}{l}{\textsc{\underline{Training-time}}} \\
DARLING~\citep{li2025jointlyreinforcingdiversityquality} & RL & \cmark \\
DivPO~\citep{lanchantin2025diverse} & DPO & \xmark \\
DQO~\citep{Chen2025PosttrainingLL} & RL & \cmark \\
Multi-answer RL~\citep{Puri2026} & RL with multiple answers& \cmark \\
Soft Preference Learning~\citep{slocum2025diversepreferencelearningcapabilities} & Entropy Decoupling & \xmark \\
\midrule
\textbf{\method (ours)} &
\textbf{\shortstack[l]{Mode-Conditioned\\MARL}} &
\cmark \\
\bottomrule
\end{tabular}
\caption{\textbf{Diversity-oriented methods.} \method{} conditions diversity on explicit mode tokens, inspired by MARL.}
\label{tab:related-work}
\vspace{-1.2em}
\end{wrapfigure}

%% file: notes_arxiv/3_method.tex
\section{\methodcolor: Mode-Conditioned Diversity Alignment for LLMs}
\label{sec:method}

\begin{figure}[t!]
    \centering
    \includegraphics[width=\textwidth]{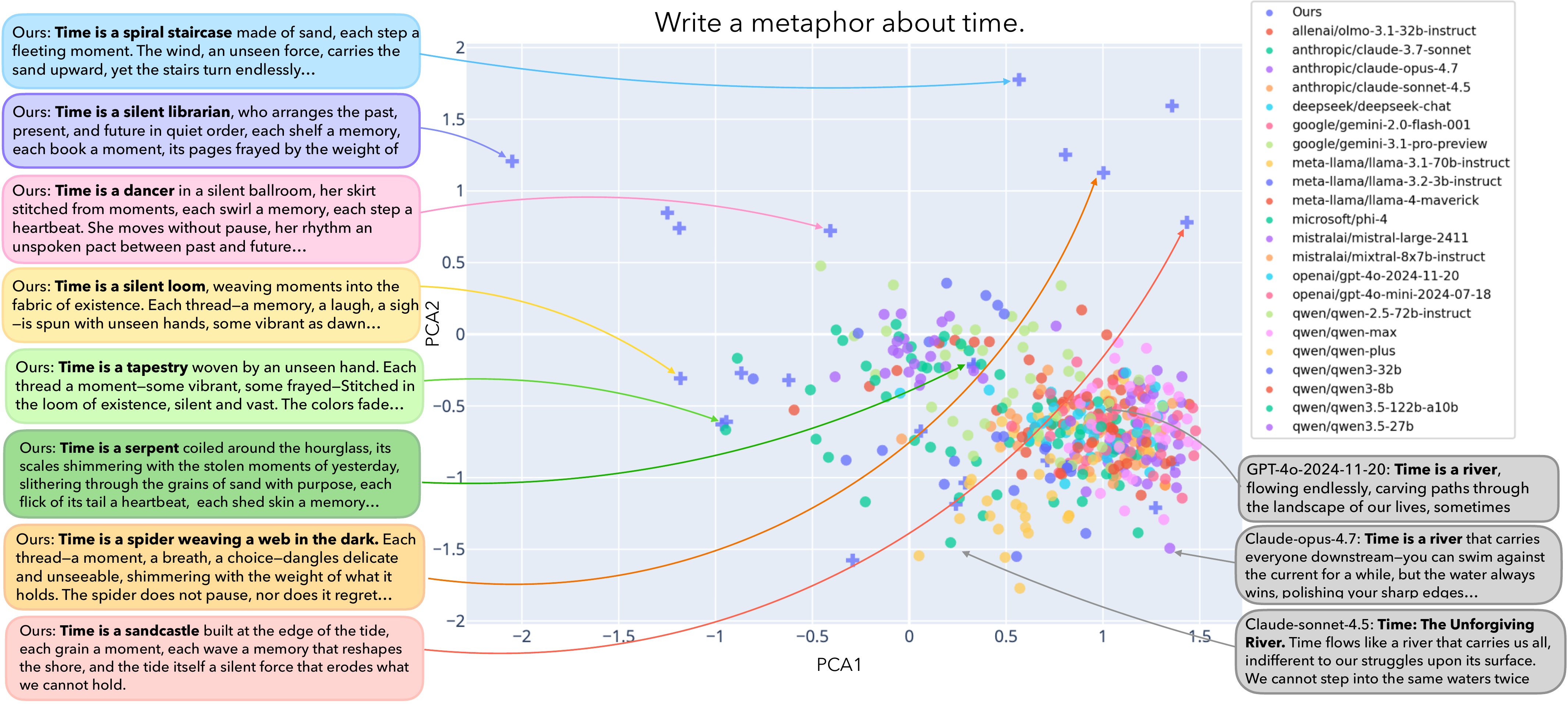}
    \caption{Responses to the query \textbf{``Write a metaphor about time''} clustered by applying PCA to reduce sentence embeddings to two dimensions. Each of the 23 off-the-shelf models and our trained non-thinking model generates 20 responses using top-$p$ sampling ($p = 1.0$) and temperature $= 1.0$.}
    \label{fig:time}
\end{figure}

\subsection{Preliminaries: LLM RL Post-Training and Mode Collapse}
\label{ssec:preliminaries}

Recent RL post-training has substantially improved the utility and safety of LLMs \citep{shao2024deepseekmath,Guo_2025}. However, optimizing only the expected reward often leads to \emph{mode collapse}, where the learned policy focuses on a small set of high-reward responses while suppressing equally valid alternatives \citep{kirk2024understandingeffectsrlhfllm}. As a result, generated responses become repetitive despite maintaining high quality. \textbf{Our goal is to learn policies that generate responses that are simultaneously high-quality, diverse, and non-redundant.}

Let $\mathcal{S}$ denote the space of natural language token sequences. Given a prompt $x \in \mathcal{S}$, a language model $\pi(\cdot \mid x)$ defines a probability distribution over responses in $\mathcal{S}$, where $\pi(y \mid x)$ denotes the probability of generating response $y \in \mathcal{S}$. For each prompt, we independently sample $k$ responses, $y_1, \ldots, y_k \sim \pi(\cdot \mid x)$, forming a \emph{response group} $\mathcal{Y}=\{y_i\}_{i=1}^{k}$.

Conventional RL post-training maximizes the expected reward
\begin{equation}
\max_{\theta}
\;
J_{\mathrm{RL}}(\theta)
=
\mathbb{E}_{x \sim \mathcal{D}}
\left[
\mathbb{E}_{y \sim \pi_{\theta}(\cdot \mid x)}
r(x,y)
\right],
\end{equation}
where $r(x,y)$ is the reward assigned to response $y$ for prompt $x$. Optimizing this objective alone progressively concentrates the policy on a small set of high-reward responses, reducing the policy entropy, which is defined as: 
\begin{equation}
H(\pi_{\theta}(\cdot \mid x))
=
-
\sum_{y \in \mathcal{S}}
\pi_{\theta}(y \mid x)
\log
\pi_{\theta}(y \mid x),
\end{equation}
and causing the model to repeatedly generate only a few dominant responses despite the existence of many alternatives with comparable rewards. We refer to this phenomenon as \emph{mode collapse}.

\subsection{Multi-Agent Reinforcement Learning (MARL)-Inspired Mode Conditioning}

\paragraph{MARL Motivation.} Classical MARL studies how populations of agents can learn complementary behaviors by explicitly encouraging behavioral diversity through mechanisms such as information sharing, skill specialization, KL divergence maximization, and adversarial objectives \citep{eysenbach2018diversityneedlearningskills, liang2024learning, li2021celebratingdiversitysharedmultiagent}. We bring this perspective to LLM post-training by introducing \textit{mode conditioning}, where a single policy is conditioned on different role tokens and each role is treated as an abstract agent. Agents must learn how to produce responses that are unlike those of the other agents, introducing competitive dynamics that drive the agents to continually diversify their responses. Unlike standard alignment, which learns a single behavioral interface, mode conditioning provides a \textit{richer interface} that enables the model to internalize multiple high-quality behavioral modes.

\paragraph{Mode Conditioning via System Role Identifiers.} We instantiate mode conditioning by prepending role identifiers into the system prompt. In our primary setting, we adopt an intentionally minimal identity-based conditioning: \texttt{You are role $i$}. We use numbered roles instead of hand-crafted personas to avoid injecting domain-specific assumptions about desirable perspectives. Although our ablations show that crafted personas (e.g., ``innovative thinker,'' ``passionate educator,'' or ``methodical experimenter'') and token-based conditioning (e.g. ``Start with your response with APPLE / BANANA / ORANGE.") are also effective, we adopt the minimal identity-based conditioning to encourage emergent specialization, where each agent adaptively learns how to play its role in order to  maximize the diversity objective.

Formally, let $x \in \mathcal{X}$ denote a user prompt and $\mathcal{Z}={z_1,\ldots,z_N}$ a fixed set of role identifiers. A shared policy $\pi_\theta(y\mid z_i, x)$ is conditioned on role $z_i$, with all roles sharing parameters $\theta$. During training, each role generate one response for $K$ prompts within the batch, forming a response group $\mathcal{Y}_x=\{y_{i,k}: i=1,\ldots,N; k=1,\ldots,K\}$. Each response receives a Quality-Gated Diversity Reward (\S~\ref{subsec:qgdr}), and the shared policy is optimized with Group Relative Policy Optimization (GRPO; \citep{shao2024deepseekmath}). Algorithm~\ref{alg: moda short} summarizes the training procedure, with full pseudocode provided in Appendix~\ref{alg:moda}.

\paragraph{Dual Inference Settings.} \label{sec:inference_modes} Not all prompts benefit from diverse responses. For example, factual queries such as ``\textit{What is Albert Einstein's birthday?}'' have a single correct answer. A key advantage of mode conditioning is that omitting the role instruction recovers the model's standard behavior, allowing \method{} to support both inference modes. Under the \textbf{standard mode}, the model generates a single response without a role identifier. Under the \textbf{diverse mode}, we prompt the model under multiple role identifiers to produce a candidate set ${y_1,\ldots,y_N}$. The resulting set can be returned directly, ranked by a quality model, or aggregated by a downstream model into a final answer.

\subsection{Quality-Gated Diversity Reward}
\label{subsec:qgdr}

Preserving quality while increasing diversity is critical: a diversity-only objective is vulnerable to reward hacking; it can be exploited by irrelevant, malformed, unnecessarily verbose, or superficially different responses. We therefore optimize a four-component reward consisting of a quality reward, a prompt-adaptive quality gate, a response-level diversity score, and explicit reward-hacking penalties:
\begin{equation}
\begin{aligned}
r_i
={}&
\underbrace{\lambda_q r_{\mathrm{qual}}(x,y_i)}_{\text{quality reward}}
+
\underbrace{\lambda_d g(x,y_i)}_{\text{quality gate}}
\underbrace{d(x,y_i,\mathcal{Y})}_{\text{diversity reward}} \\
&+
\underbrace{
r_{\mathrm{len}}(y_i)
+r_{\mathrm{lang}}(x,y_i)
+r_{\mathrm{fmt}}(y_i)
}_{\text{penalties}} .
\end{aligned}
\label{eq:qgdr}
\end{equation}

\paragraph{Prompt-Adaptive Quality Scoring.}
To determine whether our training preserves the original model's capability, we need a reference baseline: how the initial policy perform on the same prompt. To instantiate this baseline, for each prompt $x$, we score five responses sampled from the frozen initial policy and denote their minimum, mean, and maximum scores by $s_{\min}^{(x)}$, $\bar{s}^{(x)}$, and $s_{\max}^{(x)}$. Let $s_i=S_\phi(x,y_i)$ be the reference score assigned by the reward model. We define the prompt-adaptive threshold  $\tau_q$ and quality reward magnitude factor $\sigma_q$ as
\begin{equation}
    \tau_q^{(x)} = s_{\min}^{(x)} + \alpha \bigl(s_{\max}^{(x)} - \bar{s}^{(x)}\bigr),
    \qquad
    \sigma_q^{(x)} = \gamma \bigl(\bar{s}^{(x)} - s_{\min}^{(x)} + \epsilon\bigr),
\end{equation}
where $\alpha$ shifts the quality threshold above the reference minimum by a fraction of the reference score spread, $\gamma$ controls the sharpness of the scoring transition, and $\epsilon > 0$ ensures numerical stability.

The \textbf{binary quality gate} and\textbf{ quality reward}  for a generated response $y_i$ are
\begin{equation}
g(x, y_i) = \mathbf{1}\!\left[s_i \geq \tau_q^{(x)}\right]
,
\quad
    r_{\mathrm{qual}}(x, y_i)
    =
    \mu \tanh\!\left(
    \frac{s_i - \tau_q^{(x)}}{\sigma_q^{(x)}}
    \right)
\end{equation}
where $\mu$ controls the magnitude of the quality reward. This quality gate ensures that the model receives diversity reward only when the response meets the prompt-adaptive quality threshold $\tau_q^{(x)}$. As a result, responses that are merely more unusual, off-topic, or malformed compared to the initial policy would not receive diversity credit, preventing spurious solutions which hack the diversity reward.
Meanwhile, the quality reward ensures that responses above the threshold receive a positive bonus, while responses below the prompt-adaptive threshold still receive a quality-learning signal but receive no diversity credit. The bounded $\tanh$ transformation prevents the quality term from dominating optimization once the response already exceeds the threshold by a large margin. 

\paragraph{Diversity Scoring.}
For each prompt, the response group 
$\mathcal{Y}=\{y_1,\ldots,y_6\}$ contains one candidate from each of the six modes and best fits on 4 GPUs: one for the reward model and three for inference. We embed the final answer text using
\texttt{sentence-transformers/all-MiniLM-L6-v2}~\citep{wang2020minilmdeepselfattentiondistillation}.
% and $\ell_2$-normalize the resulting 384-dimensional embeddings:
% \begin{equation}
% \mathbf{e}_i
% =
% \frac{E_{\mathrm{SBERT}}(y_i)}
% {\lVert E_{\mathrm{SBERT}}(y_i)\rVert_2}.
% \end{equation}

The diversity reward directly optimized during training is the normalized semantic distance to the closest alternative within the response group:
\begin{equation}
d(x,y_i,\mathcal{Y})
=
\min_{j\neq i}
\frac{1-\mathbf{e}_i^\top\mathbf{e}_j}{2}.
\label{eq:training-sbert-diversity}
\end{equation}
Here, $y_i$ and every $y_j$ are responses to the same prompt but are generated under different roles, and each response is therefore compared with the other five candidates in its group. For thinking-enabled models, the hidden thinking trace is removed before embedding. This nearest-neighbor formulation rewards a response only if it is sufficiently distinct from its closest alternative, preventing multiple responses from collapsing to the same mode while a single outlier dominates the group-level diversity score. Consequently, role-conditioned agents are incentivized to specialize in different regions of the response space. Although we use semantic distance by default, the framework is agnostic to the diversity metric, and lexical, syntactic, or learned semantic measures can be substituted directly.

\paragraph{Reward-Hacking Penalties.}
The terms $r_{\mathrm{len}}$, $r_{\mathrm{lang}}$, and $r_{\mathrm{fmt}}$ are non-positive penalties for excessive length, language mismatch, and formatting or role-label leakage, respectively. These terms suppress superficial ways of increasing measured diversity without improving the usefulness of the response.
\subsection{Policy Optimization}
\label{subsec:policy_optimization}

We optimize the shared policy with on-policy reinforcement learning. For each prompt, all role-conditioned samples form one reward group. We optimize $\pi_\theta$ using Group Relative Policy Optimization
(GRPO~\citep{shao2024deepseekmath}), which estimates advantages from
within-group reward comparisons without a separate value network:
\begin{equation}
    \mathcal{L}_{\mathrm{GRPO}}
    = \mathbb{E}_{x,\{y_i\}}\!\left[
        \frac{1}{k} \sum_{i=1}^{k}
        \min\!\left(
            \rho_i \hat{A}_i,\;
            \mathrm{clip}(\rho_i,\, 1{-}\varepsilon,\, 1{+}\varepsilon)\,\hat{A}_i
        \right)
        - \beta\,\mathbb{D}_{\mathrm{KL}}\!\left(\pi_\theta \,\|\, \pi_{\mathrm{ref}}\right)
    \right]
    \label{eq:grpo}
\end{equation}
where $\rho_i = \pi_\theta(y_i \mid x) / \pi_{\mathrm{ref}}(y_i \mid x)$ is the
importance ratio \citep{kloek1978bayesian}, $\hat{A}_i = (r_i - \mathrm{mean}(\mathbf{r})) /
\mathrm{std}(\mathbf{r})$ is the group-normalized advantage over
$\mathbf{r} = \{r(x, y_i, \mathcal{Y})\}_{i=1}^{k}$
from Eq.~\ref{eq:qgdr}, and $\mathbb{D}_{\mathrm{KL}}\!(\pi_\theta \,\|\, \pi_{\mathrm{ref}})$ controls divergence from the base language model \citep{jaques2017sequencetutorconservativefinetuning}.
The reward is computed over the set of role-conditioned generations, but the trainable model remains a single shared policy.

\begin{algorithm}
\caption{\textsc{MoDA}: \textbf{Mo}de-Conditioned \textbf{D}iversity \textbf{A}lignment }
\label{alg: moda-stage2}
\small
\begin{algorithmic}[1]
\Require Dataset $\mathcal{D}$ with precomputed $(s_{\min},s_{\max},\mu_{\rm ref})$; policy $\pi_\theta$; reference policy $\pi_{\rm ref}$; reward model $S_\phi$; abstract roles $\mathcal{M}=\{m_i\}_{i=1}^N$; diversity metric $\delta$; weights $\lambda_q,\lambda_d$; constants $\alpha,\gamma,\epsilon,\eta$
\For{training iteration $t=1,\ldots,T$}
    \State Sample minibatch $\mathcal{B}\subset\mathcal{D}$
    \ForAll{$x\in\mathcal{B}$}
        \State Retrieve $(s_{\min}(x),s_{\max}(x),\mu_{\rm ref}(x))$
        \State $\tau_q^{(x)}\gets s_{\min}+\alpha(s_{\max}-\mu_{\rm ref})$,\quad
        $ \sigma_q^{(x)}\gets \gamma(\mu_{\rm ref}-s_{\min}+\epsilon)$
        \State Generate mode-conditioned responses with abstract roles
        $\mathcal{Y}(x)=\{y_i\sim\pi_\theta(\cdot\mid x,m_i)\}_{i=1}^N$
        \For{$i=1,\ldots,N$}
            \State $q_i\gets S_\phi(x,y_i)$ \quad \# Reward model assess response $y_i$
            \quad
            \State $R_{\rm quality}\gets \mu\tanh\!\left((q_i- \tau_q^{(x)})/ \sigma_q^{(x)}\right)$ \quad \# Compute quality bonus
            \State $G_{\rm qual}\gets \mathbf{1}\{q_i\ge  \tau_q^{(x)}\}$,\quad
            $R_{\rm div}\gets \min_{j\ne i}\delta(y_i,y_j)$ \# Quality gate and diversity reward
            \State $R_i\gets \lambda_q R_{\rm qual}
            +\lambda_d G_{\rm qual}R_{\rm div}
            +R_{\rm penalty}(x,y_i)$
        \EndFor
        \State $A_i\gets \bigl(R_i-\mathrm{mean}(R_{1:N})\bigr)/
        \bigl(\mathrm{std}(R_{1:N})+\eta\bigr)$ for $i=1,\ldots,N$
    \EndFor
    \State Update $\pi_\theta$ with GRPO using advantages $\{A_i\}$ and KL regularization to $\pi_{\rm ref}$
\EndFor
\State \Return trained policy $\pi_\theta$
\end{algorithmic}
\label{alg: moda short}
\end{algorithm}

%% file: notes_arxiv/4_experiments.tex
\section{Experiments}
\label{sec:experiments}
We implement \method~using the verl codebase \citep{sheng2024hybridflow}, using vLLM \citep{kwon2023efficient} for inference and FSDP \citep{zhao2023pytorch} for training. We use Qwen3-8B-Instruct (Qwen/Qwen3-8B) \citep{yang2025qwen3technicalreport} as the base model. We sample 5 responses using the initial policy for each prompt, precompute the reference responses using the reward model \texttt{Skywork-Reward-V2-Llama-3.1-8B-40M} \citep{liu2025skywork}, and store them as part of the dataset. During training, we sample 6 role-response pairs from the model for each prompt as a response group, and embed them using \texttt{all-MiniLM-L6-v2} \citep{wang2020minilmdeepselfattentiondistillation}. Our training has two generation settings: thinking-enabled and thinking-disabled. Quality and diversity metrics are computed only over the final answer, with the thinking trace masked out, throughout the training-time reward computation and evaluation. We include more implementation details in Appendix \ref{appx:implementation}.

\subsection{Baselines}
\label{subsec:baselines}
Given that \method{} is a training-based method, the most direct comparisons are training-time approaches. Additionally, we include inference-time methods to showcase how they compare to training-time methods. On the training-time side, both \textbf{DARLING}~\citep{li2025jointlyreinforcingdiversityquality} and \textbf{DivPO} \citep{lanchantin2025diverse} update model parameters to jointly optimize for diversity and quality. \textbf{DivPO} is a preference-optimization (DPO) method that constructs response preference pairs by selecting rare, high-quality responses as preferred and common, low-quality ones as rejected. \textbf{DARLING} optimizes an online RL objective with a learned diversity signal. We trained two versions of DARLING: one using our mixture data specified in Sec \ref{subsec:training data} and one using their specified recipe (10k prompts subsampled from WildChat). On the inference-time side,  we repeatedly sample the base model $k$ times per prompt without role conditioning, testing whether stochastic decoding alone suffices. \textbf{\ssot}~\citep{Misaki2026} is a representative prompting method that induces diversity by injecting random string seeds in the thinking tokens as an entropy source. Notably, inference-time and training-time methods are complementary. We can apply inference-time methods to \method{} at decoding time to further amplify diversity improvements.

%\textbf{Naive Role Prompting} applies the same role codes to the frozen model without role-diversity training, isolating the contribution of training from role labels alone. \textbf{Quality-Only RL} trains with the quality reward and penalties with $\lambda_d = 0$, ablating the diversity term.
%\textbf{Diversity-Only RL} trains with the diversity reward and penalties with $\lambda_q = 0$, ablating the quality term.
%\textbf{No-Role Diversity RL} trains with the full QGDR objective but without role conditioning, testing whether an unconditioned policy can learn diverse behavior without explicit control codes.
%We include DARLING because it is the closest training-time baseline for jointly reinforcing quality and diversity. 
%When possible, we use the official implementation; otherwise, we reproduce the reward structure in the same training stack used for our method.

\subsection{Training Data}
\label{subsec:training data}
The training set contains 10K prompts, constructed from a 4:1 mixture of uniformly sampled Tulu3-SFT-Mixture prompts~\citep{lambert2024tulu3} and Infinite-Chat dataset~\citep{jiang2025artificialhivemindopenendedhomogeneity}. We use this mixture to balance quality and diversity: Tulu3-SFT-Mixture provides instruction-following prompts that help preserve response quality, while Infinite-Chat encourages open-ended exploration and promote response diversity.

\subsection{Generative Diversity Evaluation}
To evaluate generative diversity, we focus on two application domains where multiple distinct high-quality outputs are desirable: \textbf{scientific ideation} and \textbf{creative writing}. We evaluate all methods under the \textbf{diverse mode}, with the role code injected into the system prompt. Notably, none of the evaluation datasets below overlap with our training data, so strong performance here reflects genuine transfer rather than in-domain memorization. In the \textbf{science ideation} domain, we adapt HypoBench~\citep{liu2026hypobenchsystematicprincipledbenchmarking} and PreScience~\citep{Ajith2026} for our cases. In the \textbf{creative writing} domain, we evaluate on NoveltyBench~\citep{zhang2025noveltybenchevaluatinglanguagemodels} and a held-out set of Infinite-Chat~\citep{jiang2025artificialhivemindopenendedhomogeneity}. Below we detail HypoBench and PreScience, whose adaptation to our setting warrants further explanation. We measure output diversity using a comprehensive set of diversity metrics:  \textit{semantic-level} (SBERT), \textit{entropy-level} (E-Vendi) \citep{friedman2023vendiscorediversityevaluation}, and \textit{entailment-level} (Discriminator/Disc.). We train a lightweight discriminator on frozen sentence embeddings (\texttt{all-mpnet-base-v2}) to predict pairwise response diversity, using bidirectional NLI-derived labels (\texttt{microsoft/deberta-v3-large}; entailment vs. non-entailment on 200-token prefixes) from 20K response pairs generated by prompting Qwen3-30B across 5 sampling modes and 2 seeds on 2,000 Alpaca \citep{alpaca} prompts. We include more implementation details in Appendix \ref{appx:discriminator}.

\textbf{HypoBench.} We adopt HypoBench \citep{liu2026hypobenchsystematicprincipledbenchmarking}, a benchmark designed to evaluate LLMs on hypothesis generation. In each task, the model is given a dataset and asked to propose a plausible hypothesis. Although HypoBench was originally designed to evaluate a model's capacity for inductive reasoning, it also fits naturally within the broader setting of research ideation, in which theories are formed from empirical observations. Diversity is especially important in this setting because the same observation can often support multiple plausible explanations, and generating varied hypotheses increases the chance of uncovering non-obvious patterns and alternative mechanisms for further investigation.

\textbf{PreScience.} We adopt the Contribution Generation task from PreScience~\citep{Ajith2026}, where the model is given a set of prior works and asked to generate a plausible title--abstract pair for a future scientific contribution. This setup mirrors one mode of scientific collaboration: a team of scientists, each bringing expertise from their prior work, collaborates to develop a new research idea grounded in those foundations. Diversity is crucial for this task because scientific progress often depends on exploring multiple possible combinations of prior ideas.

\subsection{General Capability Retention Evaluation}

\liweiaddressed{add a short sentence to highlight why we need general capability tests.}
To assess whether our method preserves the model's reasoning capability, we evaluate each trained model in the standard single-response setting on a suite of seven widely used benchmarks under the \textbf{standard mode}. Specifically, we use GSM8K~\citep{cobbe2021gsm8k} to evaluate mathematical reasoning; MMLU~\citep{hendryckstest2021}, GPQA~\citep{rein2024gpqa}, and BoolQ~\citep{clark2019boolq} to evaluate broad knowledge and question answering; HellaSwag~\citep{zellers2019hellaswag} to evaluate commonsense inference; TruthfulQA-MC1~\citep{lin2022truthfulqameasuringmodelsmimic} to evaluate truthfulness; and IFEval~\citep{zhou2023instructionfollowingevaluationlargelanguage} to evaluate instruction following. We used a max token length of 8192 for the thinking-enabled setting, 2048 for the thinking-disabled setting. 
% \liwei{where do we include these general capability results?}

\liweiaddressed{Can we specify in what mode of the dual mode did we test the general benchmarks on?}

%\textbf{Capability retention.}
%Although diversity is the main target, the model should still behave normally when no role code is supplied. We therefore evaluate no-role performance on closed-form or standard capability benchmarks such as GPQA, GSM8K, MATH-style reasoning, and MMLU-style QA. These results are treated as retention checks rather than the main diversity claim.

%\textbf{Human evaluation.}
%For a subset of NoveltyBench and Infinite-Chat prompts, human annotators compare anonymized response sets. Annotators judge whether the set is more diverse, whether the responses remain useful, and whether any responses are off-task or repetitive. Human judgments are reported separately for diversity and usefulness to avoid rewarding low-quality variation.

%\textbf{Statistical reporting.}
%All main comparisons use prompt-level paired statistics. We report mean scores with bootstrap confidence intervals over prompts. For pairwise system comparisons, we report paired bootstrap differences. For human evaluation, we report preference rates, disagreement rates, and confidence intervals.
\section{Results}
% \begin{figure}[t!]
%     \centering
%     \includegraphics[width=\textwidth]{figures/pareto}
% \vspace{-0.5cm}
% \label{fig:intro}
% \end{figure}

\begin{figure}[t]
\centering
\begin{minipage}{0.32\textwidth}
  \centering
  \includegraphics[width=\linewidth]{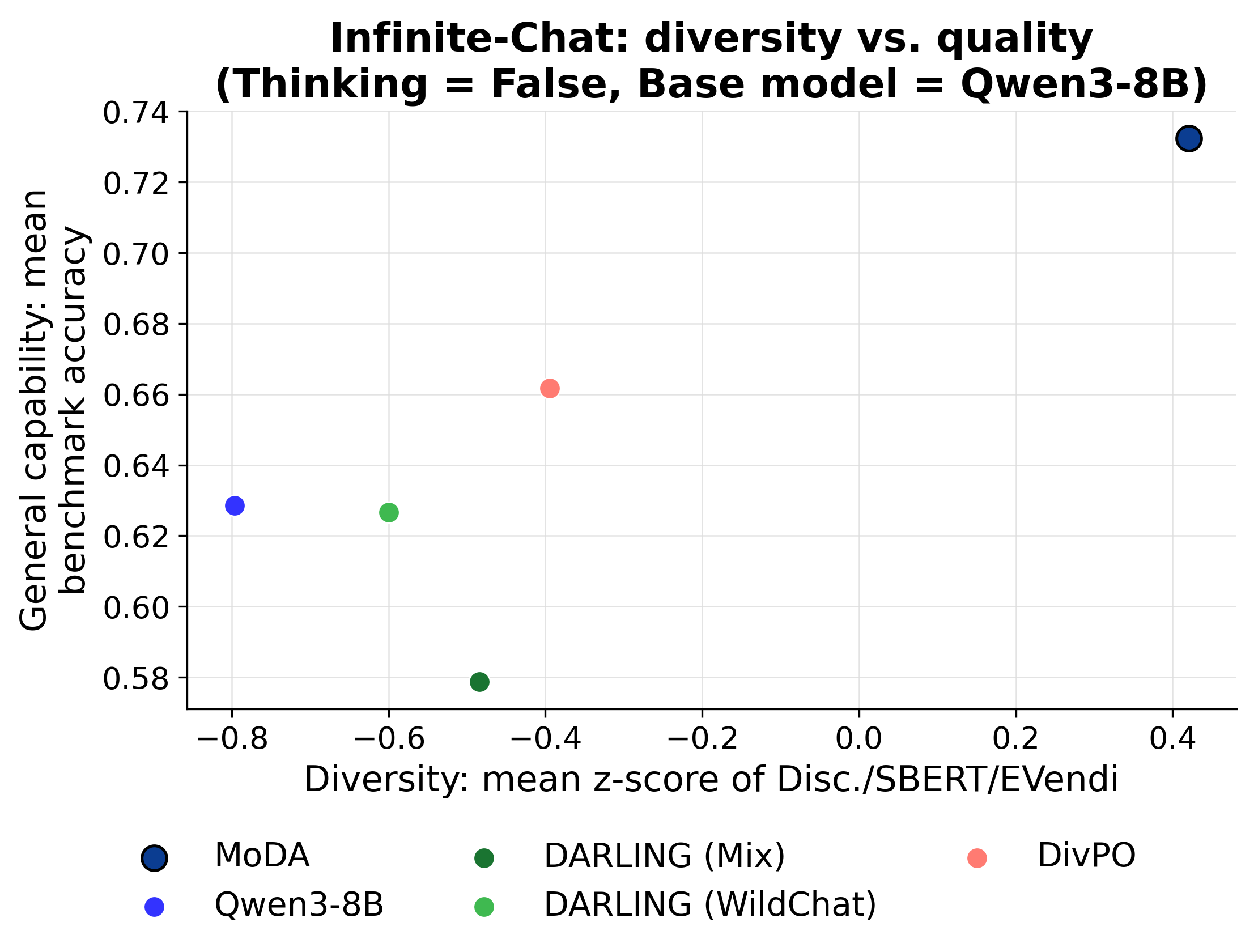}
\end{minipage}\hfill
\begin{minipage}{0.33\textwidth}
  \centering
  \includegraphics[width=\linewidth]{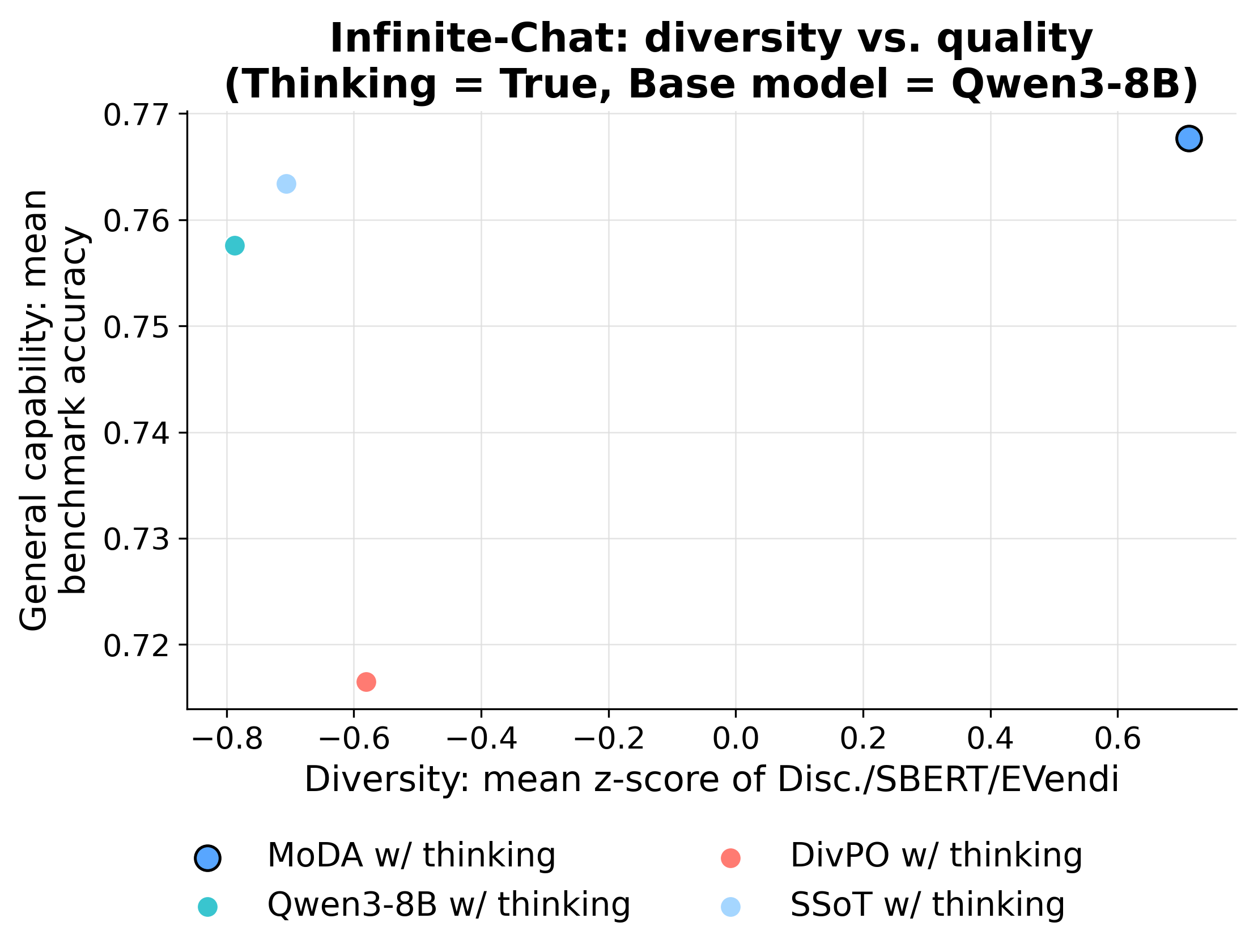}
\end{minipage}\hfill
\begin{minipage}{0.33\textwidth}
  \centering
  \includegraphics[width=\linewidth]{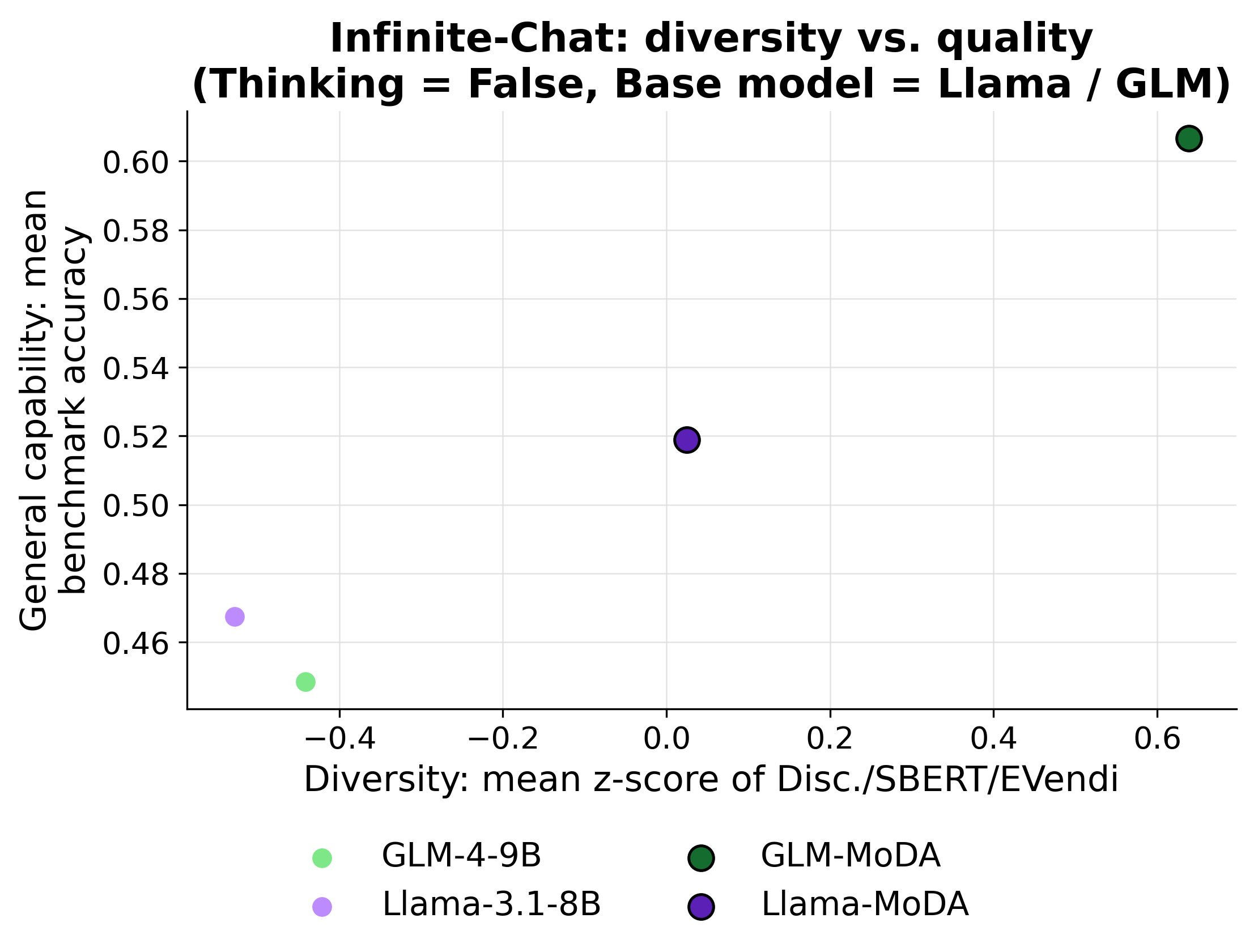}
\end{minipage}
\caption{\textbf{Diversity vs. Quality pareto figures  across base models and thinking modes.} The x-axis is Infinite-Chat response diversity (mean z-score of Disc./SBERT/EmbVendi) and the y-axis is general capability (mean benchmark accuracy). (a) Qwen3-8B base, thinking enabled. (b) Qwen3-8B base, thinking disabled. (c) Llama-3.1-8B / GLM-4-9B base, thinking disabled. Across all three settings, \method{} Pareto-dominates all the baselines, achieving better diversity and quality simultaneously.}
\label{fig: domain application pareto}
\end{figure}

\textbf{Generative Diversity Evaluation.} As shown in Figure~\ref{fig: domain application pareto}, \method{} Pareto-dominates all baselines on the held-out Infinite-Chat set across every setting (thinking enabled/disabled, different base models), achieving both higher diversity and higher quality simultaneously. We provide a breakdown by individual benchmarks in Table~\ref{tab:domain-app-diversity-main}. Across four domains and three model families, \method{} achieves the best average diversity across all evaluated baselines, while matching or improving quality on most domains except PreScience. Notably, \method{} outperforms all baselines on HypoBench, a scientific ideation task that is out of domain relative to our training set, which primarily consists of standard SFT data and open-ended questions. These results suggest that \method{} enhances general exploration capability across thinking settings and base models, even on out-of-domain topics.

\input{notes_arxiv/table/diversity_main}

\begin{figure}[t]
\centering
\begin{minipage}{0.32\textwidth}
  \centering
  \includegraphics[width=\linewidth]{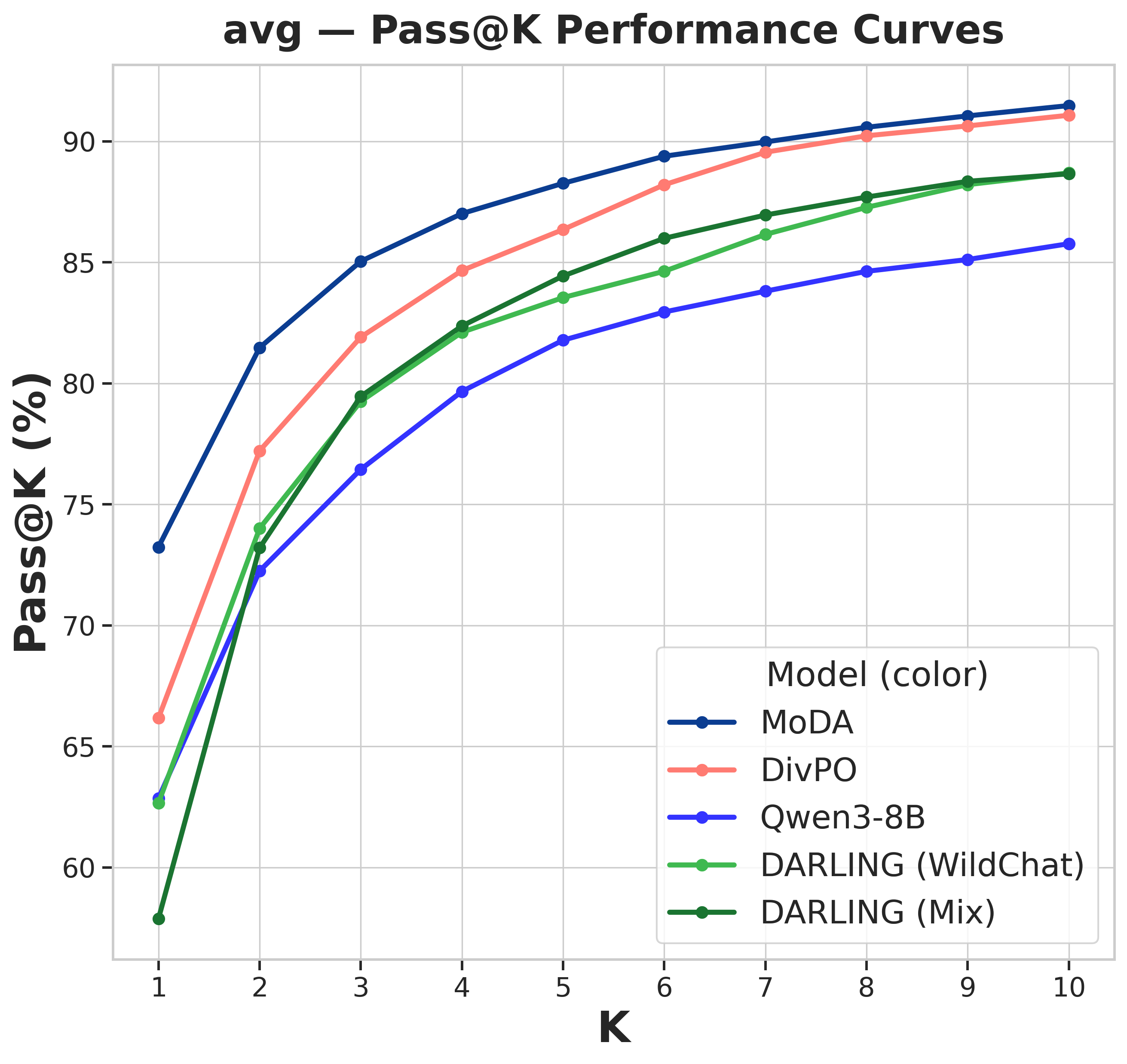}
\end{minipage}\hfill
\centering
\begin{minipage}{0.32\textwidth}
  \centering
  \includegraphics[width=\linewidth]{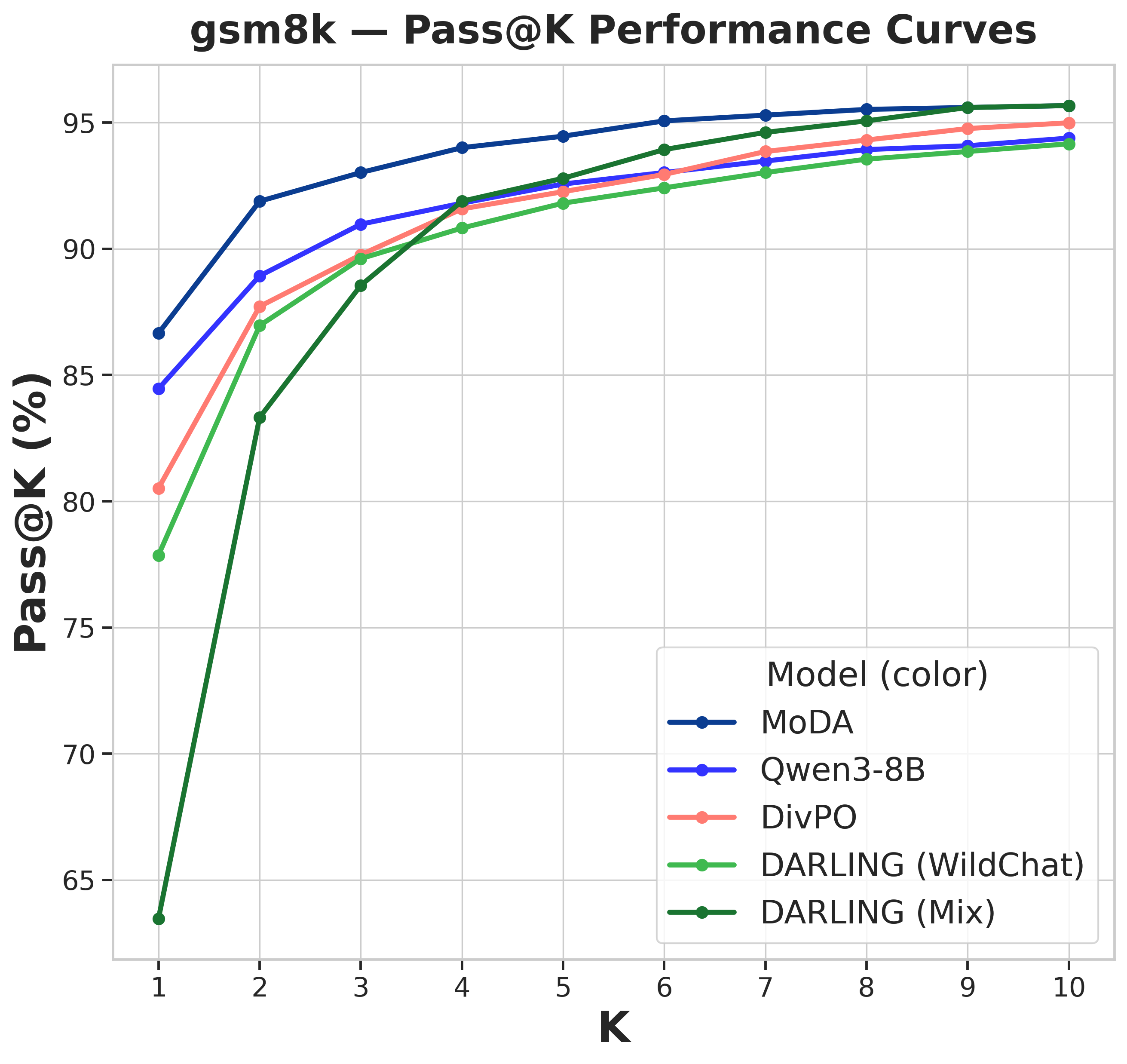}
\end{minipage}\hfill
\begin{minipage}{0.32\textwidth}
  \centering
  \includegraphics[width=\linewidth]{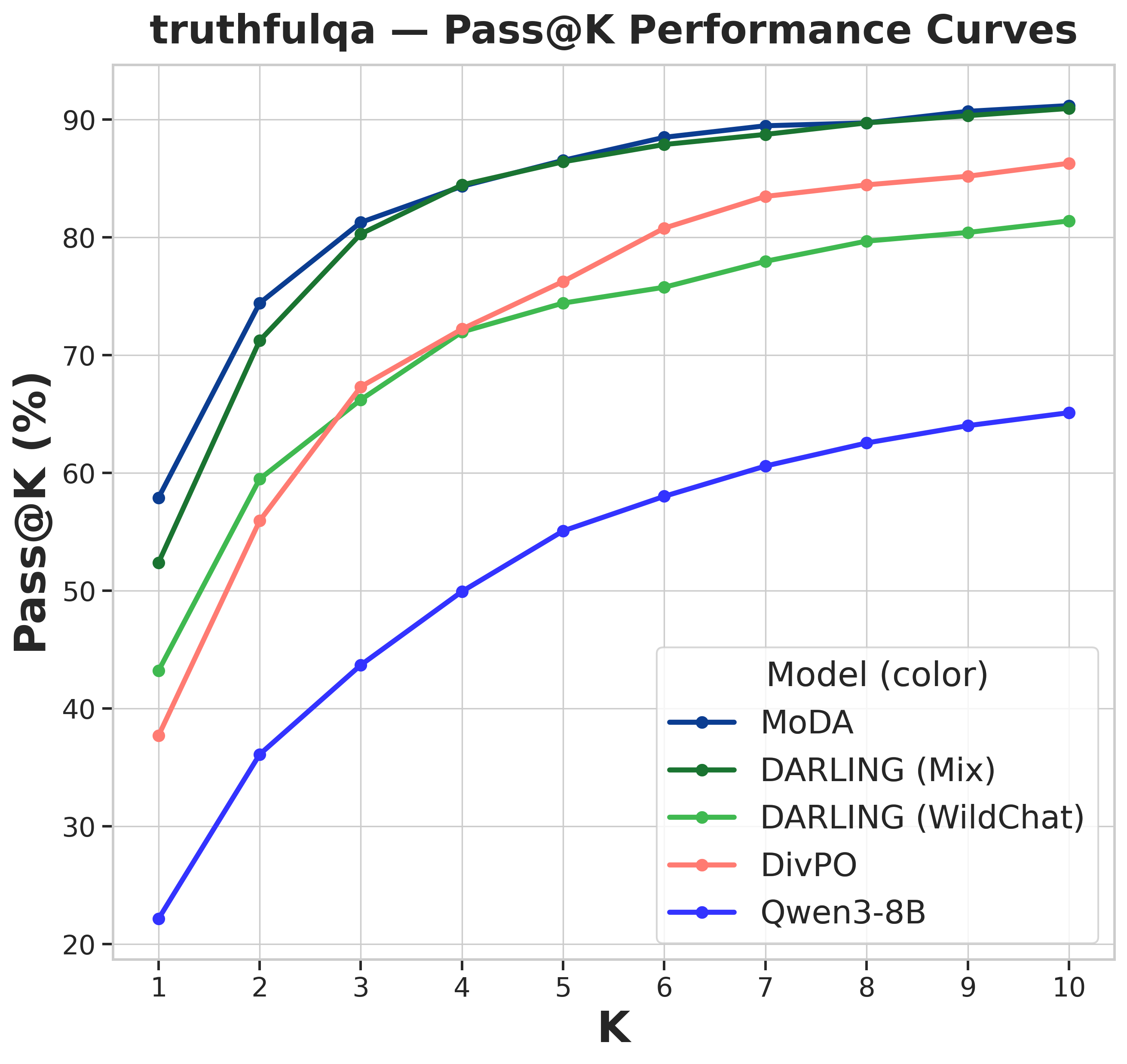}
\end{minipage}\hfill
% \begin{minipage}{0.45\textwidth}
%   \centering
%   \includegraphics[width=\linewidth]{figures/main_pass_k/pass_at_k_figures/pass_at_k_hellaswag_thinking_false_base_model_qwen3-8b.png}
% \end{minipage}
\caption{\textbf{General capability retention: pass@k accuracy.} Qwen3-8B base, thinking disabled models on the general capability suite average, GSM8K and TruthfulQA. More results can be found in Appendix \ref{appx:general capability retention} Figure \ref{fig:general capability pass@k}.}
\label{fig:general capability pass@k}
\end{figure}

\textbf{General Capability Retention.} As shown in Table \ref{tab:general-capability-pass1} and Figure \ref{fig:general capability pass@k}, our model achieves the best overall average pass@1, pass@5 and pass@10 score for general capability retention across thinking settings and base models among all baselines. With Qwen3-8B base and thinking disabled, it matches or exceeds the base model on 5 out of 7 benchmarks and achieves the best score among all baselines on 4 of them on pass@1 accuracy. Notably, with the GLM-4-9B base, \method improves average pass@1 by 15.8 percentage points, including a 3.89x relative improvement on GSM8K and a 2.37x relative improvement on MMLU. These results suggest that our method is an effective alignment approach that preserves, and in several cases significantly improves, the model's capabilities. We note that the language mixing penalty is important to maintain quality in some cases; without it, quality drops due to reward hacking from language mixing.

%We attribute this to the language-mixing penalty applied during training, which addresses the base model's native language-mixing issue. These results suggest that our method is an effective alignment approach that preserves, and in several cases significantly improves, the model's general capabilities.

\input{notes_arxiv/table/general_capability}

% \begin{table}[H]
% \centering
% \small
% \setlength{\tabcolsep}{4pt}
% \begin{tabular}{lrrrr}
% \toprule
% Model Variant & HypoBench $\uparrow$ & PreScience $\uparrow$ & NoveltyBench $\uparrow$ & Infinite-Chat $\uparrow$ \\
% \midrule
% Base (No Thinking) & \textbf{4.025 $\pm$ 0.124} & \textbf{4.150 $\pm$ 0.191} & 4.403 & 0.514 $\pm$ 0.019 \\
% Ours (No Thinking) & 3.816 $\pm$ 0.075 & 3.602 $\pm$ 0.230 & \textbf{5.826} & \textbf{0.661 $\pm$ 0.018} \\
% \midrule
% Base (Thinking) & 3.920 $\pm$ 0.159 & 1.928 $\pm$ 0.206 & 3.147 & 0.497 $\pm$ 0.019 \\
% \ssot (Thinking) & 3.825 $\pm$ 0.113 & 1.494 $\pm$ 0.147 & 2.743 & 0.614 $\pm$ 0.018 \\
% Ours (Thinking) & 3.904 $\pm$ 0.114 & 3.581 $\pm$ 0.278 & 4.082 & 0.457 $\pm$ 0.019 \\

% \bottomrule
% \end{tabular}
% \caption{Aggregate domain-application quality across HypoBench, PreScience, NoveltyBench, and Infinite-Chat. Values are means with standard errors where available; NoveltyBench reports a single official aggregate in this run. Higher is better, and bold indicates the best model in each column.}
% \label{tab:domain-app-quality-error-bars}
% \end{table}
\vspace{-0.5cm}
\subsection{Qualitative Results}
\label{sec:qualitative results}
We observe several interesting patterns in our experiments. Figure~\ref{fig:time} compares responses to the prompt \textbf{``Write a metaphor about time''} generated by the thinking-disabled \method{} model and 23 off-the-shelf models. The \method{} generations occupy a larger region of the embedding space, suggesting that mode-conditioned training expands the model's exploration space. Figure~\ref{fig:moda + 40blue} visualizes responses to the prompt \textbf{``Name a shade of blue''} from Qwen3-8B, Qwen3-8B with \ssot, and \method{} with thinking disabled. Consistent with the quantitative results, \method{} produces a larger set of valid and distinct responses.

Interestingly, although \method{} uses only abstract numbered roles rather than hand-crafted personas, the model sometimes learns role-specific behavioral patterns, suggesting that abstract mode conditioning can induce differentiated characteristics, learned via competitive multi-agent training, without manually specifying each role. We provide more qualitative generation examples in Figure \ref{fig:qualitative-example-books}.

%We include additional qualitative examples in Appendix~\ref{appx:generation examples}.

\input{notes_arxiv/appendix/generation_book_list}
\liweiaddressed{can we pick example so that we can show across dirrent roles the generations are creative and different, while for other models, resampling doesn't give sufficiently creative candidates?}

\subsection{Ablation Studies.}
% \input{notes_arxiv/table/ablation_table}
% In this section, we perform ablation studies on the design choices of \method. 

\paragraph{Ablation Setups.}
First, to understand whether the diversity improvement comes from training or solely from role injection, we inject numbered roles to the base model (\textbf{Role-conditioned prompting}). Second, to examine whether the diversity gains are driven by the explicit diversity reward or arise naturally from repeated sampling and RL optimization, we trained \method{} with \textbf{Diversity only} reward and \textbf{Quality only} reward. Third, to understand the effect of quality-gated diversity reward, we trained \method{} with additive reward of weighted sum of quality and diversity metrics, with diversity weight = 1, 10 and 100 (\textbf{Additive w=1\textbackslash10\textbackslash 100}). Finally, to understand the effect of the naive numbered role injection, we trained several variants of \method: (1) \textbf{Crafted personas}, which conditions the roles with manually crafted persona descriptions (e.g. ``You are a creative problem solver.''; (2) \textbf{Single role}, conditions with only one role; (3) \textbf{Token-based roles}\liweiaddressed{again I think we should change this name to something a bit more serious.}, conditions the role with a dummy token (e.g. ``Start your response with APPLE.''). More implementation details are in Appendix \ref{appx: ablation model implementations}. We evaluate each ablation variants on general capability suite under standard and diverse decoding, and application domain suite under diverse decoding.

\begin{figure}[t]
\centering
\begin{minipage}{0.35\textwidth}
  \centering
  \includegraphics[width=\linewidth]{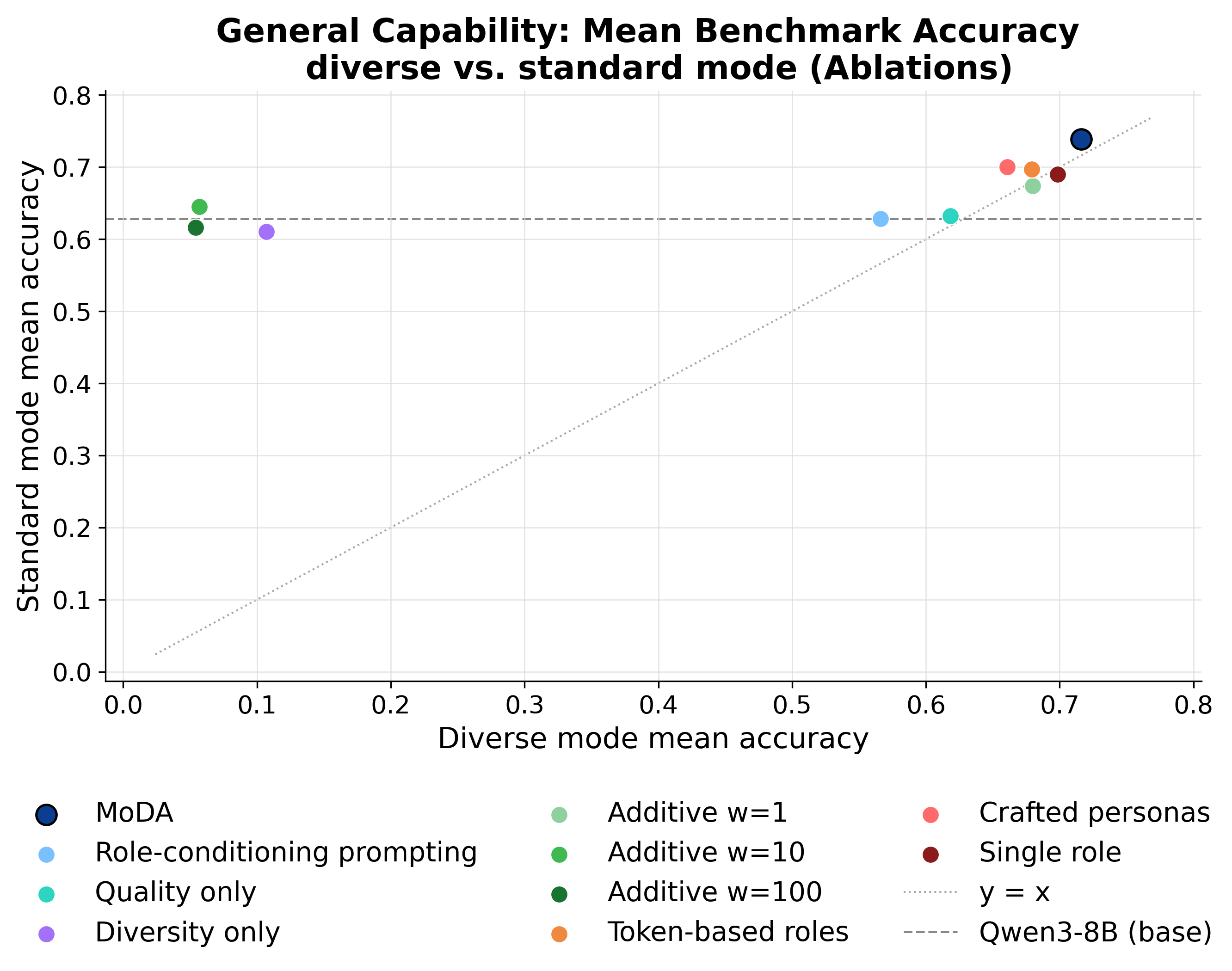}
  \small (a) Diverse vs. standard decoding mode
  \label{fig: ablation diverse standard}
\end{minipage}\hfill
\begin{minipage}{0.35\textwidth}
  \centering
  \includegraphics[width=\linewidth]{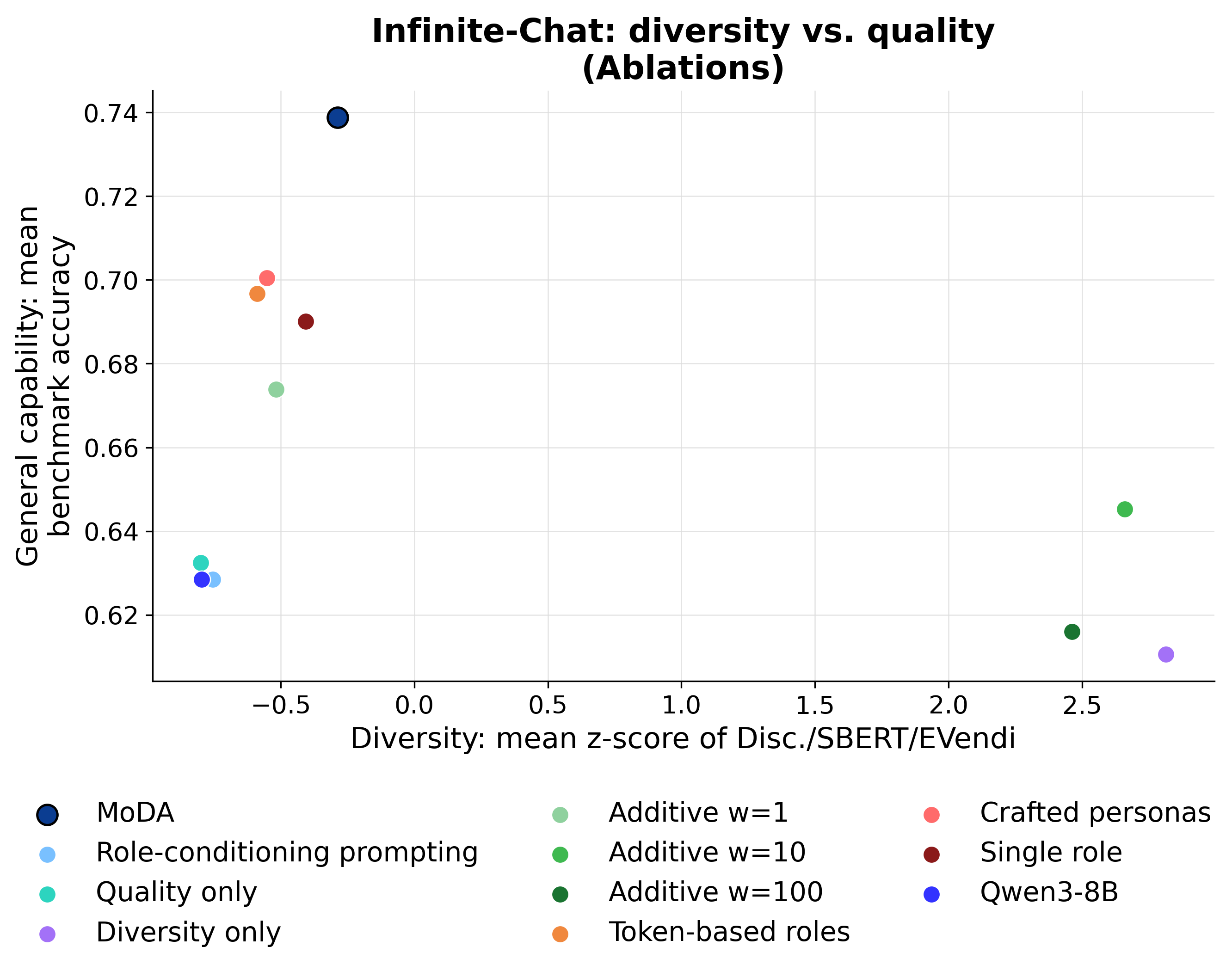}
  \small (b) Diversity-quality pareto front
  \label{fig: ablation diversity quality pareto}
\end{minipage}\hfill
\begin{minipage}{0.29\textwidth}
  \centering
  \includegraphics[width=\linewidth]{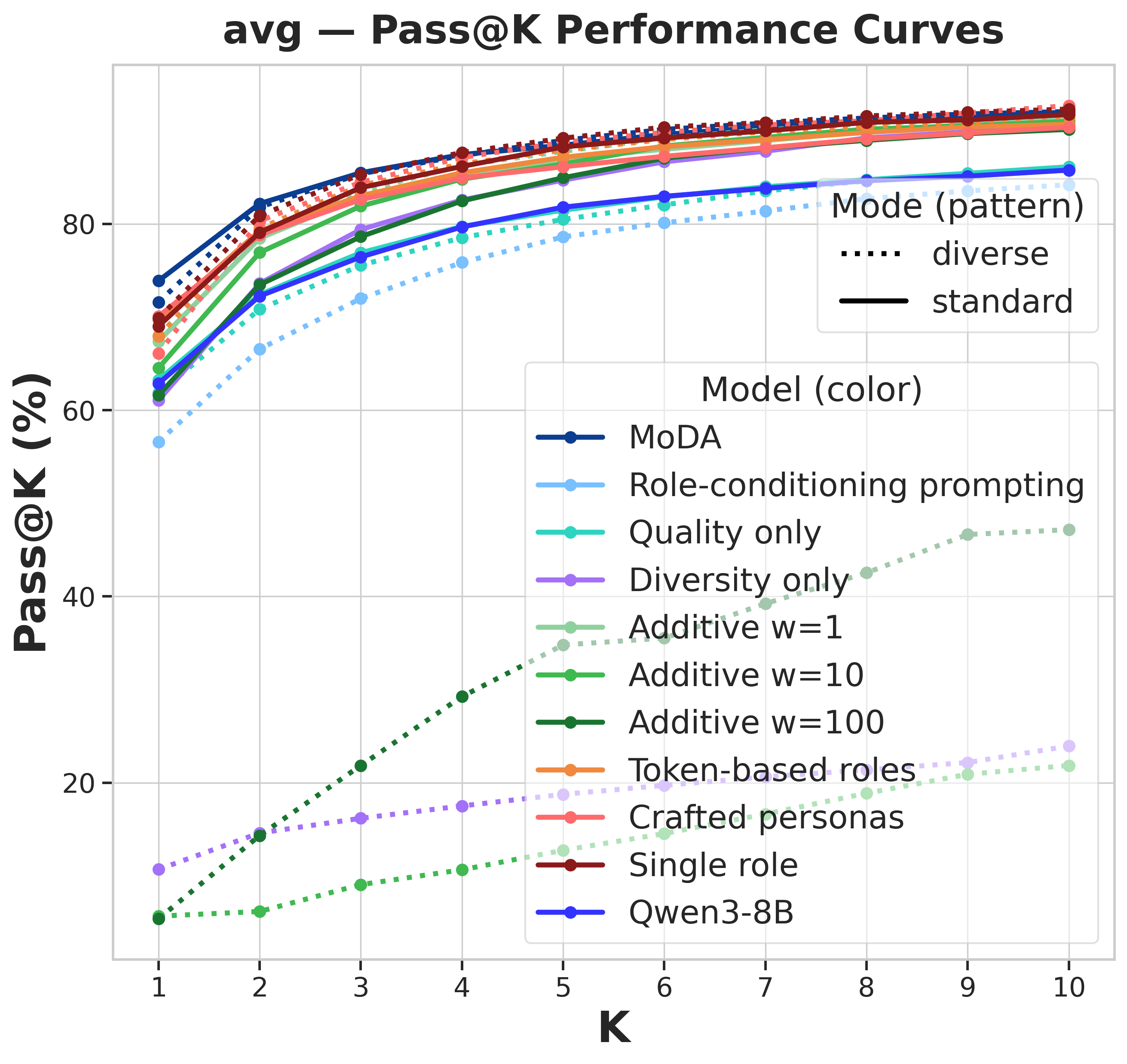}
  \small (b) Pass@k, Avg.
  \label{fig: ablation pass@k avg}
\end{minipage}\hfill
\caption{\textbf{Ablation study: diversity-quality trade-off under diverse and standard mode, and general capability retention pass@k accuracy.} 
\textbf{(a)} Diverse vs standard mode avg. pass@1 accuracy.
\textbf{(b)} Infinite-Chat diversity vs.\ quality (general capability pass@1 accuracy) across ablation variants. \textbf{(c)} Pass@k accuracy curves for the ablation variants on the general capability suite average, showing how capability under standard (solid) and diverse (dotted) decoding scales with $k$. We provide a breakdown by individual benchmarks in Appendix \ref{appx: ablation results} Figure \ref{fig:general capability pass@k ablation by benchmark}.}
\label{fig:ablation-pareto-pass-k}
\end{figure}

\liweiaddressed{add a paragraph title to this, like the follwing two paragraphs.}
\textbf{Quality-Diversity Trade-off.} As shown in Figure~\ref{fig:ablation-pareto-pass-k}a and \ref{fig:ablation-pareto-pass-k}b, \method{} achieves the best accuracy on the general capability suite among all variants. \method{} also achieves 77\% higher SBERT diversity and 41.6\% higher E-Vendi diversity than role-conditioning prompting alone, indicating that role-conditioning prompting by itself is insufficient to induce diverse generation. The quality-only variant regresses in diversity relative to role-conditioning prompting, suggesting that an explicit diversity objective is necessary for diversity gains. The diversity-only variant achieves the best diversity metrics overall, but with the lowest accuracy under standard decoding, and its performance collapses entirely under diverse decoding, indicating that explicit quality gating is critical to preserving response quality.

\textbf{Additive Aggregation.}\label{ablation: additive aggregation} Many existing works on quality-diversity optimization aggregate quality and diversity objectives using a weighted sum~\citep{mouret2015illuminatingsearchspacesmapping, Chen2025PosttrainingLL}. We compare our method with variants trained using an additive reward objective. We find that additive aggregation is highly sensitive to the diversity weight $\lambda_d$. at $\lambda_d=1$, the additive variant shows modest diversity improvement over the baseline, but substantially less than \method{}; at $\lambda_d=10$ and $\lambda_d=100$, diversity improves significantly, but at the cost of general capability performance under diverse decoding. In contrast, \method{} separates quality control from diversity optimization through the quality gate, making the reward less sensitive to coefficient scaling and enables faster iteration.

\textbf{Role Injection.}
\label{ablation: role injection}
Prior work has shown that manually designed prompts can induce different model
behaviors \citep{park2024generative, argyle2023out}. We therefore craft six personas and inject them into the system prompt instead of the abstract numbered roles, testing whether hand-designed personas provide a stronger diversity prior than the abstract numbered role used in the main setting. To isolate the effect of role multiplicity from role content, we train a single-role variant in which every response is generated under the same \textit{``You are Role 1.''} system prompt. To study the effect of the semantic role identity, we train a token-based conditioning variant in which the system prompt instructs a token-based distinguishing signal (e.g. \textit{``Start your response with APPLE/BANANA/ORANGE.''}).
As shown in Table~\ref{tab:ablation}, all variants exhibit a slight diversity improvement over the base model, but their gains are smaller than \method{}'s, indicating that multiple abstract numbered identity-based role provides a more effective exploration space than all the ablated variants, without sacrificing generalization.

%% file: notes_arxiv/table/diversity_main.tex
\begin{table*}[t]
\centering
\setlength{\tabcolsep}{3pt}
\resizebox{\textwidth}{!}{%
\begin{tabular}{lccccccccccccccccccc}
\toprule
Model & \multicolumn{4}{c}{\textit{Infinite-Chat}} & \multicolumn{4}{c}{\textit{NoveltyBench}} & \multicolumn{4}{c}{\textit{HypoBench}} & \multicolumn{4}{c}{\textit{PreScience}} & \multicolumn{3}{c}{Avg.} \\
\cmidrule(lr){2-5} \cmidrule(lr){6-9} \cmidrule(lr){10-13} \cmidrule(lr){14-17} \cmidrule(lr){18-20}
 & Disc. & SBERT & E-V & Qual.(\%) & Disc. & SBERT & E-V & Qual & Disc. & SBERT & E-V & Qual. & Disc. & SBERT & E-V & Qual. & Disc. & SBERT & E-V \\

\midrule
\multicolumn{9}{l}{\textsc{\underline{Qwen3-8B base, thinking disabled}}} \\
Qwen3-8B & 0.400 & 0.132 & 1.83 & 62.9  & 0.493 & 0.224 & 2.36 & 4.58 & 0.384 & 0.144 & 1.91 & \textbf{4.03} & 0.431 & 0.155 & 1.94 & \textbf{4.25} & 0.427 & 0.164 & 2.01 \\
DARLING (Mix) & \underline{0.430} & 0.203 & 2.33 & 57.9  & 0.513 & 0.281 & 2.96 & \underline{5.24} & 0.406 & \underline{0.222} & \underline{2.48} & 3.44 & \textbf{0.559} & \textbf{0.541} & \textbf{4.73} & 2.33 & \underline{0.477} & \underline{0.312} & \underline{3.12} \\
DARLING (WildChat) & 0.416 & 0.185 & 2.15 & 62.7  & 0.480 & 0.335 & 3.11 & 4.00 & 0.415 & 0.205 & 2.34 & 3.73 & 0.446 & 0.192 & 2.21 & \underline{4.03} & 0.439 & 0.229 & 2.45 \\
DivPO & 0.410 & \underline{0.274} & \underline{2.86} & \underline{66.2}  & \underline{0.519} & \underline{0.380} & \underline{3.58} & 4.87 & \underline{0.427} & 0.198 & 2.30 & \underline{3.89} & \underline{0.455} & \underline{0.233} & \underline{2.41} & 3.96 & 0.453 & 0.271 & 2.79 \\
\method & \textbf{0.472} & \textbf{0.482} & \textbf{4.40} & \textbf{73.2}  & \textbf{0.537} & \textbf{0.470} & \textbf{4.19} & \textbf{5.47} & \textbf{0.480} & \textbf{0.547} & \textbf{4.91} & 3.80 & 0.440 & 0.176 & 2.10 & 3.96 & \textbf{0.482} & \textbf{0.419} & \textbf{3.90} \\
\midrule
\multicolumn{9}{l}{\textsc{\underline{Qwen3-8B base, thinking enabled}}} \\
Qwen3-8B & 0.400 & 0.136 & 1.85 & 75.8  & 0.510 & 0.258 & 2.67 & \underline{4.90} & 0.399 & 0.164 & 2.05 & \textbf{3.93} & 0.441 & 0.136 & 1.83 & \underline{3.79} & 0.438 & 0.174 & 2.10 \\
SSoT & 0.397 & 0.178 & 2.09 & \underline{76.3}  & 0.438 & 0.239 & 2.57 & 0.46 & \underline{0.419} & \underline{0.220} & \underline{2.33} & 3.75 & \textbf{0.461} & \textbf{0.281} & \textbf{2.39} & 3.30 & 0.429 & \underline{0.230} & 2.34 \\
DivPO & \underline{0.414} & \underline{0.200} & \underline{2.20} & 71.7  & \underline{0.521} & \underline{0.327} & \underline{3.12} & \textbf{5.06} & 0.409 & 0.162 & 2.04 & \underline{3.89} & \underline{0.448} & 0.185 & 2.04 & \textbf{3.87} & \underline{0.448} & 0.218 & \underline{2.35} \\
\method & \textbf{0.600} & \textbf{0.909} & \textbf{9.04} & \textbf{76.8}  & \textbf{0.589} & \textbf{0.800} & \textbf{7.63} & 3.39 & \textbf{0.579} & \textbf{0.875} & \textbf{8.06} & 3.71 & 0.439 & \underline{0.194} & \underline{2.10} & 3.68 & \textbf{0.552} & \textbf{0.695} & \textbf{6.71} \\

\multicolumn{20}{l}{\textsc{\underline{{Llama-3.1-8B base, thinking disabled}}}} \\
Llama-3.1-8B & 0.414 & 0.217 & 2.40 & 46.7  & 0.512 & 0.312 & 3.04 & 4.80 & \textbf{0.469} & 0.207 & 2.38 & 3.60 & 0.461 & \textbf{0.514} & 3.67 & \textbf{2.62} & \textbf{0.464} & 0.312 & 2.87 \\
\method{} (Llama-3.1-8B) & \textbf{0.441} & \textbf{0.373} & \textbf{3.77} & \textbf{51.9}  & \textbf{0.532} & \textbf{0.392} & \textbf{3.75} & \textbf{6.17} & 0.410 & \textbf{0.324} & \textbf{3.27} & \textbf{3.65} & \textbf{0.472} & 0.470 & \textbf{3.89} & 2.54 & \textbf{0.464} & \textbf{0.390} & \textbf{3.67} \\
\midrule
\multicolumn{20}{l}{\textsc{\underline{GLM-4-9B base, thinking disabled}}} \\
GLM-4-9B & 0.421 & 0.234 & 2.60 & 44.9  & \textbf{0.544} & 0.408 & 4.08 & \textbf{4.14} & 0.402 & 0.225 & 2.52 & \textbf{3.75} & 0.461 & 0.438 & 4.15 & 2.57 & 0.457 & 0.326 & 3.34 \\
\method{}(GLM-4-9b) & \textbf{0.481} & \textbf{0.528} & \textbf{5.14} & \textbf{60.7}  & 0.502 & \textbf{0.521} & \textbf{5.01} & 2.70 & \textbf{0.452} & \textbf{0.485} & \textbf{4.64} & 3.68 & \textbf{0.464} & \textbf{0.448} & \textbf{4.28} & \textbf{2.58} & \textbf{0.475} & \textbf{0.495} & \textbf{4.77} \\
\bottomrule
\end{tabular}
}
\caption{Per-domain diversity (Disc. = discriminator diversity, SBERT = SBERT embedding pairwise distance, E-V = E-Vendi score) and quality (Qual.), averaged over $n{=}3$ random seeds. Infinite-Chat quality is general capability accuracy (\%); all other domains use the native quality score. Bold marks the best value within each model group. Table with error bar can be found in Appx \ref{appx:generative diversity evaluation}.}
\label{tab:domain-app-diversity-main}
\end{table*}

%% file: notes_arxiv/table/general_capability.tex
\begin{table}[H]
\small
\setlength{\tabcolsep}{2.5pt}
\centering
\begin{tabular}{l*{7}{>{\centering\arraybackslash}p{0.95cm}}|>{\centering\arraybackslash}p{0.95cm}}
\toprule
Model & GSM8K & MMLU & GPQA & BoolQ & HS & TQA & IFEval & Avg. \\
\midrule
\multicolumn{9}{l}{\textsc{\underline{Qwen3-8B base, thinking disabled}}} \\
Qwen3-8B & \underline{84.5} & \textbf{82.0} & \underline{41.4} & 83.1 & 48.4 & 22.2 & \textbf{78.4} & 62.9 \\
DARLING (Mixture) & 63.5 & 63.0 & 34.3 & 76.7 & 62.2 & \underline{52.4} & 53.0 & 57.9 \\
DARLING (WildChat) & 77.9 & 55.0 & 34.3 & \textbf{86.4} & \underline{68.8} & 43.2 & 73.0 & 62.7 \\
DivPO & 80.5 & \underline{81.0} & 40.4 & 83.5 & 62.8 & 37.7 & \underline{77.3} & \underline{66.2} \\
\method & \textbf{86.7} & \underline{81.0} & \textbf{52.0} & \underline{84.8} & \textbf{73.4} & \textbf{57.9} & 76.9 & \textbf{73.2} \\
\midrule
\multicolumn{9}{l}{\textsc{\underline{Qwen3-8B base, thinking enabled}}} \\
Qwen3-8B & \textbf{90.3} & \textbf{94.0} & 43.4 & \textbf{87.2} & 70.9 & 64.9 & \textbf{79.7} & 75.8 \\
SSoT & 84.3 & \underline{93.0} & \textbf{55.1} & \textbf{87.2} & \textbf{80.1} & \textbf{72.8} & 61.9 & \underline{76.3} \\
DivPO & \underline{84.9} & 91.0 & 44.4 & \underline{83.2} & 65.0 & 55.3 & 77.6 & 71.7 \\
\method & 83.9 & \underline{93.0} & \underline{50.5} & \textbf{87.2} & \underline{76.2} & \underline{67.2} & \underline{79.5} & \textbf{76.8} \\
\midrule
\multicolumn{9}{l}{\textsc{\underline{Llama-3.1-8B base, thinking disabled}}} \\
Llama-3.1-8B & 71.3 & 16.0 & 6.1 & \textbf{81.6} & 30.2 & 53.4 & 68.8 & 46.7 \\
\method(Llama-3.1-8B) & \textbf{74.9} & \textbf{30.0} & \textbf{18.7} & 75.6 & \textbf{38.5} & \textbf{54.7} & \textbf{70.8} & \textbf{51.9}\\
\midrule
\multicolumn{9}{l}{\textsc{\underline{GLM-4-9B base, thinking disabled}}} \\
GLM-4-9B & 21.8 & 35.0 & 19.7 & 74.5 & \textbf{68.0} & \textbf{44.8} & 50.3 & 44.9  \\
\method(GLM-4-9B) & \textbf{84.9} & \textbf{83.0} & \textbf{33.3} & \textbf{81.2} & 52.3 & 34.5 & \textbf{55.5} & \textbf{60.7} \\
\bottomrule
\end{tabular}
\vspace{4pt}
\caption{General capability retention (Pass@1 accuracy \%), averaged over $n=3$ random seeds. Bold marks the best value within each model group; underline marks the second-best value within the Qwen3-8B groups. Table with error bar can be found in Appendix \ref{appx:general capability retention} Table \ref{tab:general-capability-pass1-error-bar}.}
\label{tab:general-capability-pass1}
\end{table}

%% file: notes_arxiv/appendix/generation_book_list.tex
\label{expr:generation-example-book-list}
% Prompt box
\newtcolorbox{promptbox}{
  colback=gray!8,
  colframe=gray!50!black,
  boxrule=0.5pt,
  arc=2pt,
  left=6pt,
  right=6pt,
  top=4pt,
  bottom=4pt,
  fonttitle=\bfseries,
  title=Prompt
}

% Response box, parameterized by mode/role label
\newtcolorbox{respbox}[1]{
  colback=blue!1,
  colframe=blue!80!white,
  boxrule=0.5pt,
  arc=2pt,
  left=5pt,
  right=5pt,
  top=3pt,
  bottom=3pt,
  fonttitle=\bfseries\tiny,
  fontupper=\tiny,
  title=#1,
  breakable
}

\newtcolorbox{modarespbox}[1]{
  colback=blue!4,
  colframe=blue!60!black,
  boxrule=0.5pt,
  arc=2pt,
  left=5pt,
  right=5pt,
  top=3pt,
  bottom=3pt,
  fonttitle=\bfseries\tiny,
  fontupper=\tiny,
  title=#1,
  breakable
}

\begin{figure}[t]
\centering

\begin{promptbox}
\textit{Name one well received book that often appears in top book lists.}
\end{promptbox}
\begin{minipage}[t]{0.32\linewidth}
\begin{respbox}{Sample 1}
One well-received book that often appears in top book lists is \textbf{To Kill a Mockingbird} by Harper Lee. It is a classic of American literature and has consistently ranked high in various lists, including the Modern Library's list of the 100 Best Novels and the New York Times' Best Seller list.
\end{respbox}
\end{minipage}\hfill%
\begin{minipage}[t]{0.32\linewidth}
\begin{respbox}{Sample 2}
One well-received book that often appears in top book lists is \textbf{To Kill a Mockingbird} by Harper Lee. It is widely celebrated for its powerful themes of racial injustice, morality, and empathy, and has been a staple in literature curricula and bestseller lists for decades.
\end{respbox}
\end{minipage}\hfill%
\begin{minipage}[t]{0.32\linewidth}
\begin{respbox}{Sample 3}
One well-received book that often appears in top book lists is \textbf{To Kill a Mockingbird} by Harper Lee. It is widely praised for its powerful themes of racial injustice, moral growth, and empathy. The novel has been a staple in literature curricula and frequently appears on lists such as the Modern Library's Top 100 Novels and the BBC's ``The Big Read.''
\end{respbox}
\end{minipage}

\vspace{6pt}

\begin{minipage}[t]{0.32\linewidth}
\begin{respbox}{Sample 4}
\textbf{To Kill a Mockingbird} by Harper Lee is a well-received book that frequently appears in top book lists. It has been widely praised for its powerful themes of racial injustice, moral growth, and empathy, and it has been a staple of literature curricula and bestseller lists for decades.
\end{respbox}
\end{minipage}\hfill%
\begin{minipage}[t]{0.32\linewidth}
\begin{respbox}{Sample 5}
One well-received book that often appears in top book lists is \textbf{To Kill a Mockingbird} by Harper Lee. It is widely celebrated for its powerful themes of racial injustice, moral growth, and empathy. The novel has been frequently included in lists such as the Modern Library's 100 Best Novels, the BBC's The Big Read, and other notable literary rankings.
\end{respbox}
\end{minipage}\hfill%
\begin{minipage}[t]{0.32\linewidth}
\begin{respbox}{Sample 6}
One well-received book that often appears in top book lists is \textbf{To Kill a Mockingbird} by Harper Lee. It is a classic of American literature and has consistently been featured in lists such as the Modern Library's 100 Best Novels and the Pulitzer Prize winners list.
\end{respbox}
\end{minipage}

\vspace{6pt}

\begin{minipage}[t]{0.32\linewidth}
\begin{modarespbox}{Role 1}
\textbf{``The Night Fire''} by Michael Connelly

\textbf{Reasons for its popularity and recognition:}

\textbf{Award-winning author}: Michael Connelly is a best-selling and critically acclaimed author, known for his intelligent, fast-paced crime fiction.
\end{modarespbox}
\end{minipage}\hfill%
\begin{minipage}[t]{0.32\linewidth}
\begin{modarespbox}{Role 2}
\textbf{``The Night Circus''} by Erin Morgenstern

\textbf{Why it's well-received}: Known for its richly imaginative world-building, intricate plot, and lyrical prose, the novel blends fantasy, magic, and romance in a unique and immersive way.
\end{modarespbox}
\end{minipage}\hfill%
\begin{minipage}[t]{0.32\linewidth}
\begin{modarespbox}{Role 3}
\textbf{``To Kill a Mockingbird''} by Harper Lee is a frequently cited example. It is often found on top book lists for various reasons, including: \textbf{Literary Significance}: A cornerstone of American literature, it addresses themes of racial injustice, moral growth, and empathy. 
\end{modarespbox}
\end{minipage}

\vspace{6pt}

\begin{minipage}[t]{0.32\linewidth}
\begin{modarespbox}{Role 4}
Certainly! One well-received book that frequently appears in top book lists is \textbf{``The Night Tiger''} by Yangsze Choo. This historical fiction novel is set in colonial Malaysia during the 1930s and weaves a compelling narrative around a young boy, a mysterious tiger, and a forbidden love story.
\end{modarespbox}
\end{minipage}\hfill%
\begin{minipage}[t]{0.32\linewidth}
\begin{modarespbox}{Role 5}
\textbf{``The Midnight Library''} by Matt Haig

\textbf{Why it's well received:}

\textbf{Themes:} Explores existential questions, regret, and the idea of parallel lives, resonating with readers on a personal and philosophical level. Frequently appears on bestseller lists and in literary award discussions.
\end{modarespbox}
\end{minipage}\hfill%
\begin{minipage}[t]{0.32\linewidth}
\begin{modarespbox}{Role 6}
\textbf{``The Great Gatsby''} by F. Scott Fitzgerald

\textbf{Why It's Well Received and Often Appeared on Top Lists:}

\textbf{Cultural Impact:} A cornerstone of American literature, it explores themes of idealism, resistance to change, and the American Dream.
\end{modarespbox}
\end{minipage}

\vspace{6pt}
\caption{\textbf{Qualitative example of diverse generation.} Upper: Responses sampled from Qwen3-8B. Bottom: Responses sampled from \method{} under distinct numbered roles for the same prompt, illustrating conceptual diversity across generations.}

\label{fig:qualitative-example-books}
\vspace{-0.4cm}
\end{figure}

%% file: notes_arxiv/5_discussions.tex
\section{Discussions}
\label{sec:discussions}

In this work, we introduced \method, an online-MARL-inspired alignment method that encourages diverse generation while preserving response quality. \method~serves as a drop-in replacement for existing post-training alignment pipelines, requiring no architectural modifications to the underlying model. Compared to previous methods, its dual-mode design enables seamless switching between producing a single high-confidence answer and generating a diverse set of plausible responses, offering greater flexibility depending on the downstream needs. The ability to surface conceptually distinct yet high-quality responses has broad implications for high-stakes domains such as medical diagnosis, legal reasoning, and scientific ideation.

\textbf{Limitations and Future Work.} Several limitations point to promising directions for future research. First, our quality reward model (SkyReward-V2) tends to favor verbose responses, even for prompts that should be answered briefly, and our uniform length penalty insufficiently addresses this across prompts of varying complexity. For instance, when training GLM-4-9B with \method, we observed suboptimal native quality score on NoveltyBench compared to the base model because its reward model doesn't penalize verbosity therefore it reward hack by adding unnecessary preambles such as "As a model.../To answer these questions as a model..." Removing these preambles  recovers the performance from 2.739 to 3.302 (+20\%). Future work should explore adaptive length penalties or alternative reward models that better reflect conciseness. Second, because \method~introduces no external data during training, generative diversity is bounded by the initial policy's capacity, and incorporating retrieval augmentation or external data sources could meaningfully expand the exploration space. Finally, there is a risk of producing more varied but misleading, unsafe, or unsupported outputs, especially in high-stakes domains such as medical diagnosis and legal reasoning. Although \method~mitigates this through its dual-mode design, whose standard mode retains the model's general capability, future work can further combine the diverse mode with verification, calibration, retrieval, or domain-specific safeguards.

%% file: notes_arxiv/z_acknowledgement.tex
\section*{Acknowledgment}
We thank our colleagues at the SocialRL Lab at the University of Washington for their valuable feedback and support. The work of Natasha Jaques was supported by the UW-Amazon Science Gift Hub, UW-Tsukuba Amazon NVIDIA Cross Pacific AI Initiative (XPAI), Sony Research Award, Character.AI, DoorDash, Open Philanthropy, Toyota Research Institute, and the Schmidt AI2050 Fellows program. This work was supported by DARPA under the ITM program (FA8650-23-C-7316). The views expressed are those of the author and do not reflect the official policy or position of the Department of Defense or the U.S.~Government.

% This work was also supported in part by the Defense Advanced Research Projects Agency's (DARPA) SciFy program (Agreement No. HR00112520300). 

%% file: notes_arxiv/appendix.tex
\onecolumn
\newpage
\appendix

\begin{appendices}

\startcontents[sections]
\printcontents[sections]{l}{1}{\setcounter{tocdepth}{2}}

\clearpage
\newpage

\input{notes_arxiv/appendix/algorithm}

\newpage
\clearpage

% implementation details
% dataset creation
% say something about open
% baseline
% 
% benchmark description
% qualitative result by domain

% compute
% different alternative for qgdr

\input{notes_arxiv/appendix/implementation}

\newpage
\clearpage

\input{notes_arxiv/appendix/additional_results}

\newpage
\clearpage

\input{notes_arxiv/appendix/benchmarks}

\newpage
\clearpage

\input{notes_arxiv/appendix/generation_examples}

\end{appendices}

%% file: notes_arxiv/appendix/algorithm.tex
\begin{breakablealgorithm}
\caption{\textsc{\method}: Mode-conditioned Diversity Alignment}
\label{alg:moda}
\begin{algorithmic}[1]
\Require Prompt dataset $\mathcal{D}=\{x_n\}_{n=1}^N$; initial policy $\pi_{\theta}$; reference/base policy $\pi_{\mathrm{ref}}$; quality reward model $S_{\phi}$; set of mode instructions $\mathcal{M}=\{m_1,\ldots,m_K\}$; diversity metric $\delta(\cdot,\cdot)$; quality and diversity weights $\lambda_q,\lambda_d$; KL weight $\beta$
\Require Number of reference samples $M$; number of training responses per prompt $K$; small constants $\epsilon,\eta>0$
\Statex

\State \textbf{Stage 1: Precompute prompt-adaptive reference quality statistics}
\ForAll{$x \in \mathcal{D}$}
    \State Initialize reference score set $\mathcal{S}_{\mathrm{ref}}(x) \gets \emptyset$
    \For{$j=1,\ldots,M$}
        \State Sample reference response $\tilde{y}_j \sim \pi_{\mathrm{ref}}(\cdot \mid x)$
        \State Compute quality score $\tilde{s}_j \gets S_{\phi}(x,\tilde{y}_j)$
        \State Add $\tilde{s}_j$ to $\mathcal{S}_{\mathrm{ref}}(x)$
    \EndFor
    \State Compute
    \[s_{\min}(x) \gets \min \mathcal{S}_{\mathrm{ref}}(x),
    s_{\max}(x) \gets \max \mathcal{S}_{\mathrm{ref}}(x),
    \mu_{\mathrm{ref}}(x) \gets \frac{1}{M}\sum_{\tilde{s}\in \mathcal{S}_{\mathrm{ref}}(x)} \tilde{s}\].
    \State Store $\bigl(s_{\min}(x),s_{\max}(x),\mu_{\mathrm{ref}}(x)\bigr)$ with prompt $x$ in $\mathcal{D}$
\EndFor
\Statex

%VI debugging compile error
\State \textbf{Stage 2: Mode-conditioned reinforcement learning}
\For{training iteration $t=1,\ldots,T$}
    \State Sample a minibatch $\mathcal{B}\subset \mathcal{D}$
    \ForAll{$x \in \mathcal{B}$}
        \State Retrieve stored statistics $\bigl(s_{\min}(x),s_{\max}(x),\mu_{\mathrm{ref}}(x)\bigr)$
        \State Compute prompt-adaptive quality threshold and reward magnitude factor:
        \[
        q_{\mathrm{tar}}(x)
        \gets
        s_{\min}(x)+\alpha\bigl(s_{\max}(x)-\mu_{\mathrm{ref}}(x)\bigr),
        \]
        \[
        \tau_q(x)
        \gets
        \gamma\bigl(\mu_{\mathrm{ref}}(x)-s_{\min}(x)+\epsilon\bigr).
        \]
        \State Initialize response set $\mathcal{Y}(x)\gets \emptyset$
        \For{$i=1,\ldots,K$}
            \State Select mode instruction $m_i \in \mathcal{M}$
            \State Sample response
            $y_i \sim \pi_{\theta}(\cdot \mid x,m_i)$
            \State Add $y_i$ to $\mathcal{Y}(x)$
        \EndFor
        \Statex

        \For{$i=1,\ldots,K$}
            \State Compute quality score
            \[
            q_i \gets S_{\phi}(x,y_i)
            \]
            \State Compute normalized quality margin
            \[
            z_i \gets \frac{q_i-q_{\mathrm{tar}}(x)}{\tau_q(x)}
            \]
            \State Compute bounded quality reward
            \[
            R_{\mathrm{qual}}(x,y_i)
            \gets
            \mu\tanh(z_i)
            \]
            \State Compute quality gate
            \[
            G_{\mathrm{qual}}(x,y_i)
            \gets
            \mathbbm{1}\{q_i \ge q_{\mathrm{tar}}(x)\}
            \]
            \State Compute diversity reward
        \[
            R_{\mathrm{div}}(x,y_i)
            \gets
            \min_{j\neq i}\delta(y_i,y_j)
            \]
            \State Compute reward-hacking penalty
            \[
            R_{\mathrm{pen}}(x,y_i)
            \gets
            R_{\mathrm{len}}(y_i)+R_{\mathrm{lang}}(y_i)
            \]
            \State Compute \textbf{Quality Gated Diversity Reward}
            \[
            R_i
            \gets
            \lambda_q R_{\mathrm{qual}}(x,y_i)
            +
            \lambda_d G_{\mathrm{qual}}(x,y_i)R_{\mathrm{div}}(x,y_i)
            +
            R_{\mathrm{pen}}(x,y_i).
            \]
        \EndFor
        \Statex

        \State Compute group-relative advantages:
        \[
        A_i
        \gets
        \frac{R_i-\mathrm{mean}(R_{1:K})}
        {\mathrm{std}(R_{1:K})+\eta},
        \qquad i=1,\ldots,K.
        \]
    \EndFor
    \State Update $\pi_{\theta}$ using GRPO with advantages $\{A_i\}$ and KL regularization to $\pi_{\mathrm{ref}}$:
    \[
    \theta
    \gets
    \arg\max_{\theta}
    \mathbb{E}
    \left[
    \frac{1}{K}\sum_{i=1}^{K}
    \min\left(
    \rho_i A_i,
    \mathrm{clip}(\rho_i,1-\epsilon_{\mathrm{clip}},1+\epsilon_{\mathrm{clip}})A_i
    \right)
    -
    \beta D_{\mathrm{KL}}\bigl(\pi_{\theta}(\cdot\mid x,m_i)\,\|\,\pi_{\mathrm{ref}}(\cdot\mid x,m_i)\bigr)
    \right],
    \]
    where
    \[
    \rho_i
    =
    \frac{\pi_{\theta}(y_i\mid x,m_i)}
    {\pi_{\mathrm{old}}(y_i\mid x,m_i)}.
    \]
\EndFor
\State \Return trained policy $\pi_{\theta}$
\end{algorithmic}
\end{breakablealgorithm}

%% file: notes_arxiv/appendix/implementation.tex
\section{Implementation}
\label{appx:implementation}

\subsection{Overview}

\paragraph{Training Stage} For rollout generation, we use stochastic sampling with temperature $1.0$ and top-$p=1.0$, which encourages broad exploration of the model's response distribution during RL training. This is important for \method, since the diversity reward can only provide useful learning signal when the sampled response group contains meaningful variation. We sample $k=6$ mode-conditioned responses per prompt, forming the group over which diversity rewards and GRPO advantages are computed. We use a maximum prompt length of $1024$ tokens and a maximum response length of $2048$ tokens. For the Quality Gated Diversity Reward, we used quality weight of $\lambda_q=1$, diversity reward weight of $\lambda_d=100$, length penalty weight of $0.01$, and language-mismatch penalty of weight  $0.1$. For GLM-4-9B base, we observed a severe language-mismatch issue in its response, and therefore increased the weight of the language-mismatch penalty to $100$.

For Qwen3-8B base, we trained \method{} and all baselines for 4 epochs. For Llama-3.1-8B, GLM-4-9B base and ablation models, we trained for 2 epochs due to compute resource constraints.

Our training are done on NVIDIA A100, H100 or H200 GPUs, depending on availability. A single training for 2 epochs run takes approximately 24 hours on 4 H200 GPUs.
We provide the rest of the training parameters in Table \ref{tab:hyperparameter}.

\begin{table}[t]
\centering
\caption{
Training hyperparameters used for \method{}.
}
\label{tab:training-hyperparams}
\small
\begin{tabular}{ll}
\toprule
\textbf{Hyperparameter} & \textbf{Value} \\
\midrule
\multicolumn{2}{l}{\textit{Model and data}} \\
Base model & Qwen3-8B \\
Training batch size & 32 \\
Validation batch size & 32 \\
Maximum prompt length & 1024 \\
Maximum response length & 2048 \\
Number of training epochs & 2 \\
\midrule
\multicolumn{2}{l}{\textit{Rollout generation}} \\
Rollout engine & vLLM \\
Number of responses per prompt, $N$ & 6 \\
Sampling & Enabled \\
Temperature & 1.0 \\
Top-$p$ & 1.0 \\
Top-$k$ & $-1$ \\
Maximum model length & 4096 \\
\midrule
\multicolumn{2}{l}{\textit{Policy optimization}} \\
RL algorithm & GRPO \\
Actor learning rate & $1\times 10^{-6}$ \\
Optimizer & AdamW \\
AdamW betas & $(0.9, 0.999)$ \\
Weight decay & 0.01 \\
Gradient clipping & 1.0 \\
PPO epochs & 1 \\
PPO mini-batch size & 32 \\
PPO micro-batch size per GPU & 4 \\
PPO clip ratio & 0.2 \\
Advantage normalization & Enabled \\
Loss aggregation & Token mean \\
Entropy coefficient & 0 \\
\midrule
\multicolumn{2}{l}{\textit{KL regularization}} \\
Use KL loss & Enabled \\
KL loss coefficient & 0.01 \\
KL loss type & Low-variance KL \\
KL control coefficient & 0.001 \\
Target KL & 0.1 \\
\midrule
\multicolumn{2}{l}{\textit{Reward function}} \\
Quality reward weight & 1.0 \\
Distinctiveness reward weight & 100.0 \\
Thresholding base reward & 0.5 \\
Length penalty weight & 0.01 \\
Maximum length before penalty & 512 tokens \\
Language mixing penalty weight (Qwen/Llama) & 0.1 \\
Language mixing penalty weight (GLM) & 10 \\
\bottomrule
\end{tabular}
\label{tab:hyperparameter}
\end{table}

\subsection{Ablation Model Implementations}
\label{appx: ablation model implementations}
\paragraph{Token-based roles}
As an ablation, we replace the abstract numbered role instructions with dummy roles. Specifically, for each prompt, we prepend one of the
following system prompts. We stripped the dummy tokens (e.g., \textit{APPLE}/\textit{ORANGE}/\textit{BANANA}) from the responses before computing the diversity and quality metrics, both during the training and during the evaluation.

\begin{quote}
\small
\begin{enumerate}
    \item ``Start your response with APPLE.''
    \item ``Start your response with ORANGE.''
    \item ``Start your response with BANANA.''
    \item ``Start your response with CHERRY.''
    \item ``Start your response with DATE.''
    \item ``Start your response with ELDERBERRY.''
\end{enumerate}
\end{quote}

\paragraph{Crafted roles}
As an ablation, we replace the abstract numbered role instructions with manually
crafted role descriptions. Specifically, for each prompt, we prepend one of the
following system prompts:
\begin{quote}
\small
\begin{enumerate}
    \item ``You are a helpful assistant.''
    \item ``You are a creative problem solver.''
    \item ``You are a patient educator.''
    \item ``You are a rigorous mathematician.''
    \item ``You are a skeptical analyst.''
    \item ``You are an optimistic encourager.''
\end{enumerate}
\end{quote}

\paragraph{Single role}
As an ablation, we use only one role for mode condition. We prepend ``You are Role 1." on each system prompt.

\subsection{Dataset}
\label{appx:dataset}
Our training dataset comprises a mixture of general-purpose SFT prompts, which anchor baseline capability, and open-ended prompts, which encourage exploration and response diversity. Specifically, we sample 8k prompts from \texttt{allenai/tulu-3-sft-mixture} \citep{lambert2024tulu3} as general-purpose examples and 2k prompts from Infinite-Chat as open-ended queries. We held out a part of the Infinite-Chat to used in the evaluation.

\subsection{Discriminator}
\label{appx:discriminator}
We train a lightweight pairwise binary discriminator to predict whether two responses to the same prompt are conceptually distinct. The training data is constructed from \(2{,}000\) Alpaca prompts. For each
prompt, we employ Qwen3-30B to generate responses under \(5\) modes and \(2\) random seeds,
yielding \(20{,}000\) response pairs. Pseudo-labels are assigned using bidirectional \texttt{microsoft/deberta-v3-large} on the first \(200\) tokens of each response pair. A pair is labeled non-diverse if both directions predict entailment, and diverse otherwise. We discard low-confidence pairs
whose maximum softmax probability is below \(0.7\), balance the two classes, and split the resulting \(5{,}816\) pairs by prompt into train/validation/test sets of \(4{,}064/890/862\) pairs.

\textbf{Architecture.} The discriminator uses frozen \texttt{all-mpnet-base-v2} sentence embeddings to construct its feature vector.
Given a pairs of sentence embeddings \(u,v\in\mathbb{R}^{768}\), we construct the feature vector
\[
z =
\left[
u \,\|\, v \,\|\, |u-v| \,\|\, u\odot v \,\|\, \phi(y_i,y_j)
\right]
\in \mathbb{R}^{3075},
\]
where \(\phi(y_i,y_j)\) denotes auxiliary scalar pair features.

The classifier is a three-layer MLP with $3075 \rightarrow 256 \rightarrow 64 \rightarrow 1$ nodes, ReLU activations, dropout \(0.3\), and a sigmoid output. We train with focal loss \((\alpha=0.25,\gamma=2)\), AdamW with learning rate \(3\times10^{-4}\), batch size \(64\), and early stopping on validation AUC. The model has approximately \(804\)K trainable parameters.

\subsection{Training Curve}
\begin{figure}[t]
    \centering

    % Row 1: No role, thinking disabled
    \begin{subfigure}{\textwidth}
        \centering
        \includegraphics[width=0.9\linewidth]{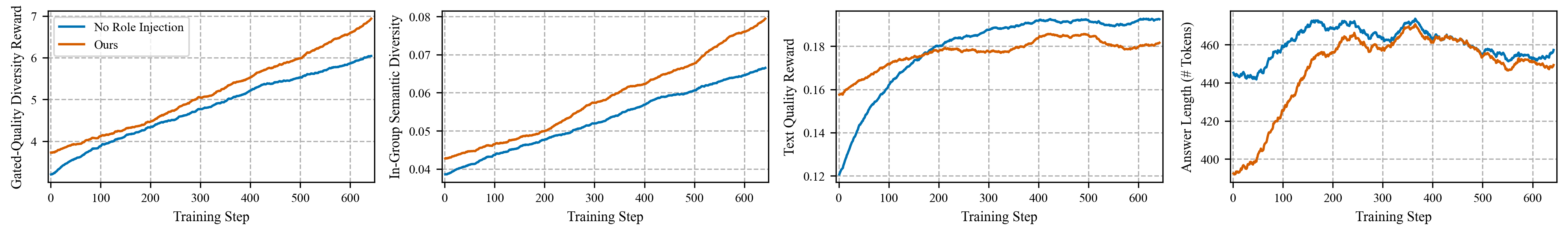}
    \end{subfigure}
    \caption*{\textbf{No role injection, thinking disabled.}}

    \begin{subfigure}{\textwidth}
        \centering
        \includegraphics[width=0.9\linewidth]{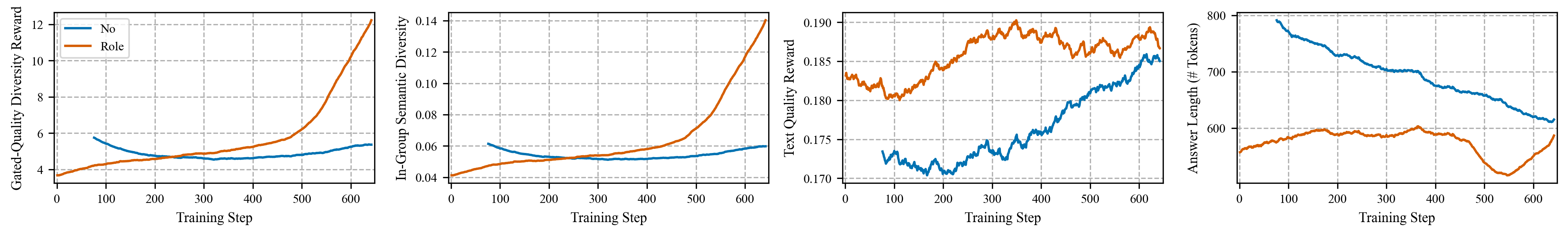}
    \end{subfigure}
    \caption*{\textbf{No role injection, thinking enabled.}}

    \vspace{0.7em}
    % Row 3: Quality-only, thinking disabled
    \begin{subfigure}{\textwidth}
        \centering
        \includegraphics[width=0.9\linewidth]{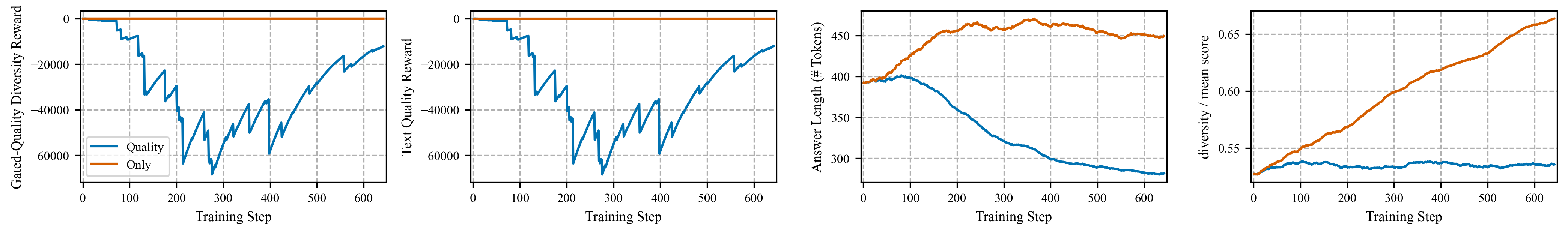}
    \end{subfigure}
    \caption*{\textbf{Quality-only reward, thinking disabled.}}

    \vspace{0.7em}

    % Row 4: Quality-only, thinking enabled
    \begin{subfigure}{\textwidth}
        \centering
        \includegraphics[width=0.9\linewidth]{figures/quality_only_wo.png}
    \end{subfigure}
    \caption*{\textbf{Quality-only reward, thinking enabled.}}

     \vspace{0.7em}
    % Row 3: Diversity-only, thinking disabled
    \begin{subfigure}{\textwidth}
        \centering
        \includegraphics[width=0.9\linewidth]{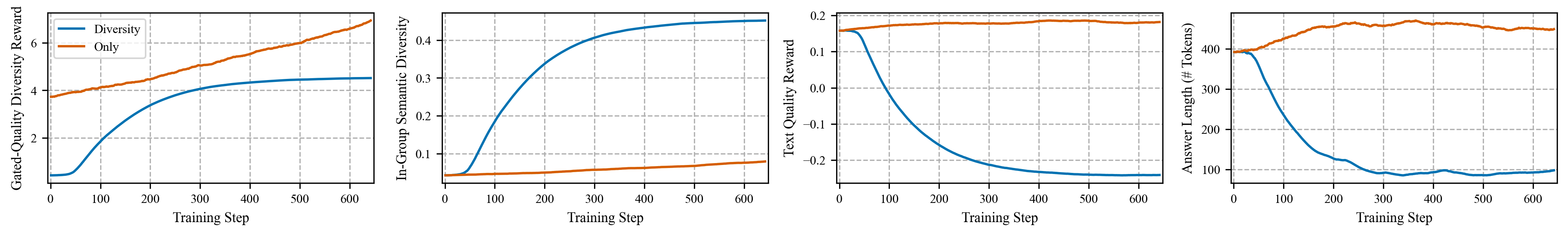}
    \end{subfigure}
    \caption*{\textbf{Diversity-only reward, thinking disabled.}}

    \vspace{0.7em}

    % Row 4: Diversity-only, thinking enabled
    \begin{subfigure}{\textwidth}
        \centering
        \includegraphics[width=0.9\linewidth]{figures/diversity_only_wo.png}
    \end{subfigure}

    \caption*{\textbf{Diversity-only reward, thinking enabled.}}

    % Row 3: Crafted role, thinking disabled
    \begin{subfigure}{\textwidth}
        \centering
        \includegraphics[width=0.9\linewidth]{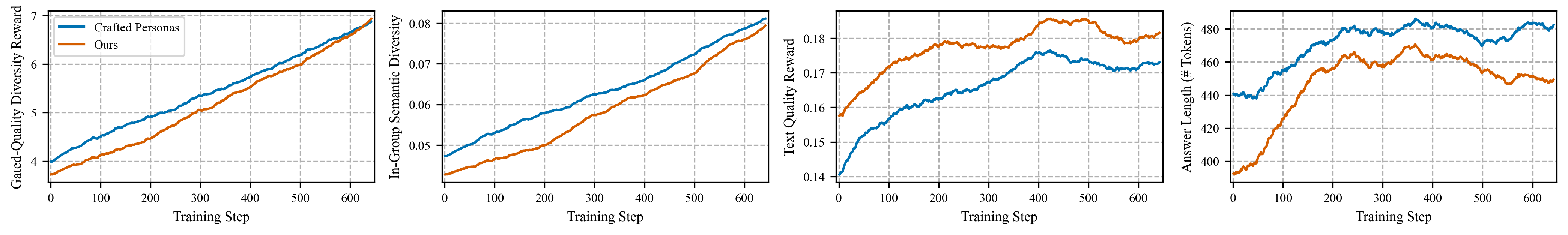}
    \end{subfigure}
    \caption*{\textbf{Crafted personas, thinking disabled.}}

    \vspace{0.7em}

    % Row 4: Crafted role, thinking enabled
    \begin{subfigure}{\textwidth}
        \centering
        \includegraphics[width=0.9\linewidth]{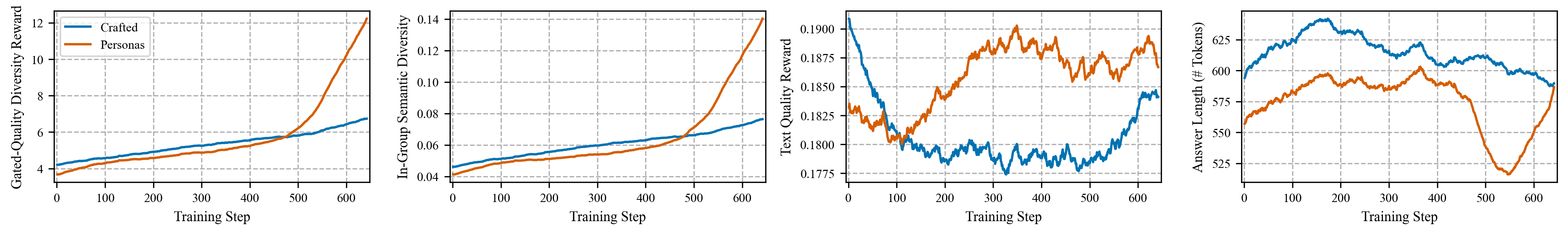}
    \end{subfigure}
    \caption*{\textbf{Crafted personas, thinking enabled.}}

    \caption{
    Training curves for ablation studies. Each row corresponds to one ablation setting, and each row reports four training diagnostics: total reward, diversity reward, quality reward, and answer token length. Results are shown under both thinking-disabled and thinking-enabled settings.}
    \label{fig:ablation-training-curves}
\end{figure}

We share the training curve for ablation studies in Figure \ref{fig:ablation-training-curves}.

%% file: notes_arxiv/appendix/additional_results.tex
\section{Additional Experiment Results}
\label{appx:additional results}

\subsection{Generative Diversity Evaluation.}
\label{appx:generative diversity evaluation}
We show the results with error bar for generative diversity across all domain application tasks.

\input{notes_arxiv/appendix/diversity_main_error_bar}

\subsection{General Capability Retention}
\label{appx:general capability retention}

\subsubsection{Pass@1}
We show general capability retention with error bar for pass@1 across baselines and our model.
\input{notes_arxiv/appendix/general_capability_error_bar}

\subsubsection{Pass@5}
We show general capability retention for pass@5 across baselines and our model.
\begin{table}[H]
\centering
\small
\setlength{\tabcolsep}{4pt}
\resizebox{\textwidth}{!}{%
\begin{tabular}{lcccccccc}
\toprule
Model & GSM8K & MMLU & GPQA & BoolQ & HellaSwag & TruthfulQA & IFEval & Avg. \\
\midrule
\multicolumn{9}{l}{\textsc{\underline{Qwen3-8B base, thinking disabled}}} \\
Qwen3-8B & 92.6 & 92.0 & 73.7 & 89.7 & 86.0 & 55.1 & 83.4 & 81.8 \\
DARLING (Mixture) & 92.8 & \textbf{95.0} & \textbf{75.3} & 93.1 & 91.2 & 86.4 & 57.3 & 84.4 \\
DARLING (WildChat) & 91.8 & 86.0 & 69.2 & 91.3 & 91.3 & 74.4 & 80.8 & 83.5 \\
DivPO & 92.3 & 92.0 & 74.7 & 93.2 & 90.5 & 76.3 & \textbf{85.6} & 86.4 \\
\method & \textbf{94.5} & 93.0 & 74.7 & \textbf{93.9} & \textbf{91.3} & \textbf{86.5} & 83.9 & \textbf{88.3} \\
\midrule
\multicolumn{9}{l}{\textsc{\underline{Qwen3-8B base, thinking enabled}}} \\
Qwen3-8B & 96.1 & 96.0 & 66.2 & 92.2 & 89.1 & 82.5 & 85.4 & 86.8 \\
SSoT & 90.6 & 96.0 & \textbf{74.7} & 92.0 & 88.9 & \textbf{83.8} & 73.6 & 85.7 \\
DivPO & 94.0 & \textbf{97.0} & 72.2 & \textbf{93.0} & \textbf{89.8} & \textbf{83.8} & 85.6 & \textbf{87.9} \\
\method & \textbf{96.3} & 96.0 & 68.7 & 92.0 & 89.6 & 82.5 & \textbf{85.8} & 87.3 \\
\midrule
\multicolumn{9}{l}{\textsc{\underline{Llama-3.1-8B base, thinking disabled}}} \\
Llama-3.1-8B & 90.1 & 57.0 & 28.3 & \textbf{93.2} & 70.2 & 72.9 & \textbf{82.3} & 70.6 \\
\method(Llama-3.1-8B) & \textbf{93.5} & \textbf{76.0} & \textbf{56.6} & 92.3 & \textbf{79.6} & \textbf{80.4} & 80.2 & \textbf{79.8} \\
\midrule
\multicolumn{9}{l}{\textsc{\underline{GLM-4-9b base, thinking disabled}}} \\
GLM-4-9B & 60.0 & 69.0 & 59.6 & \textbf{97.1} & \textbf{91.3} & \textbf{82.6} & 57.5 & 73.9 \\
\method(GLM-4-9B) & \textbf{95.2} & \textbf{94.0} & \textbf{73.2} & 95.3 & 89.1 & 77.5 & \textbf{74.7} & \textbf{85.6} \\
\bottomrule
\end{tabular}
}
\vspace{4pt}
\caption{General capability retention pass@5. For IFEval, strict instruction-following accuracy is reported. Avg.\ is the row mean across benchmarks. Bold marks the best average within each thinking setting.}
\label{tab:general-capability-pass5}
\end{table}

We show pass@k figures for the rest of the general capability suite tasks below.
\begin{figure}[H]
\centering
\begin{minipage}{0.24\textwidth}
  \centering
  \includegraphics[width=\linewidth]{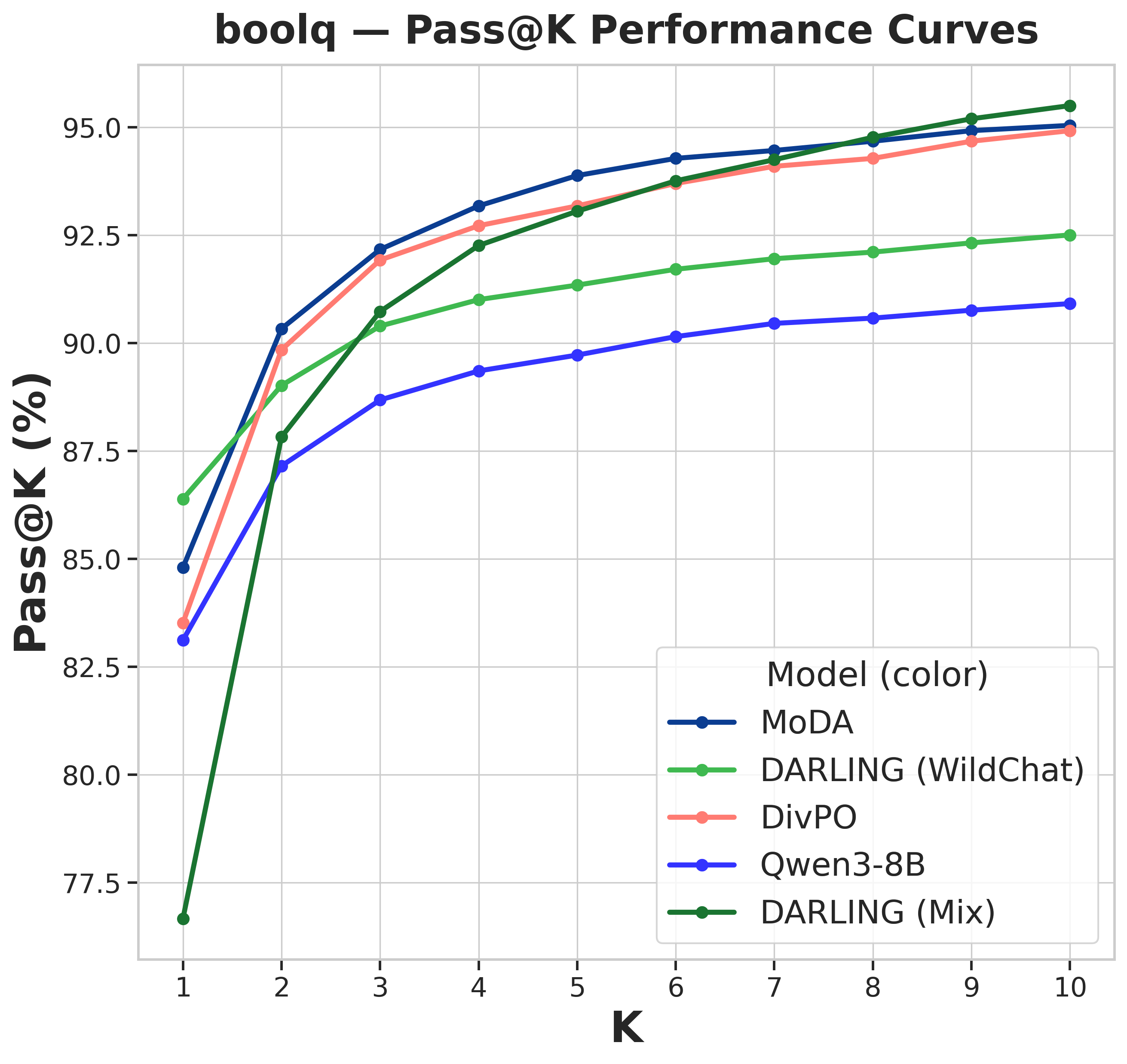}
\end{minipage}\hfill
\centering
\begin{minipage}{0.24\textwidth}
  \centering
  \includegraphics[width=\linewidth]{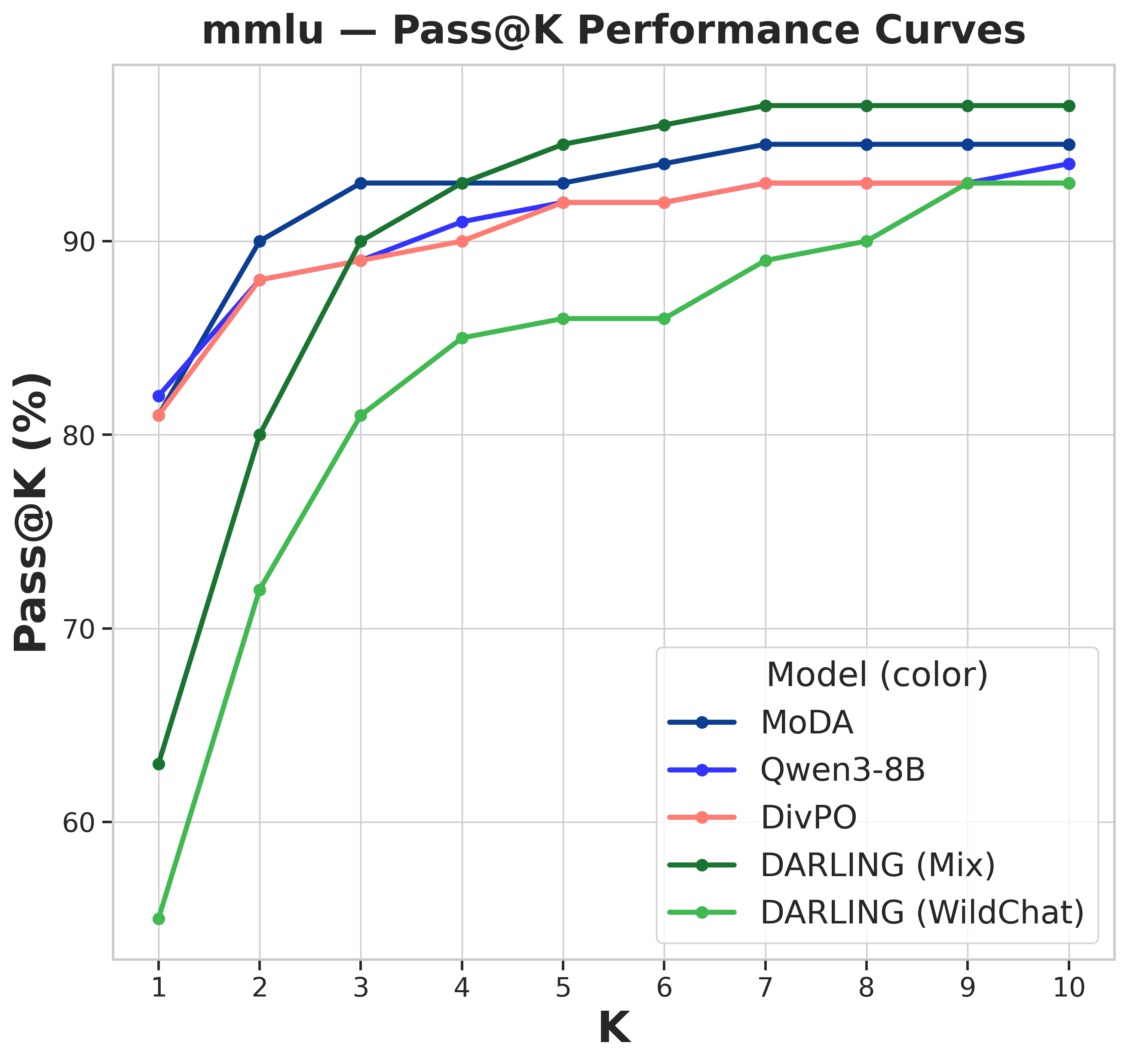}
\end{minipage}\hfill
\begin{minipage}{0.24\textwidth}
  \centering
  \includegraphics[width=\linewidth]{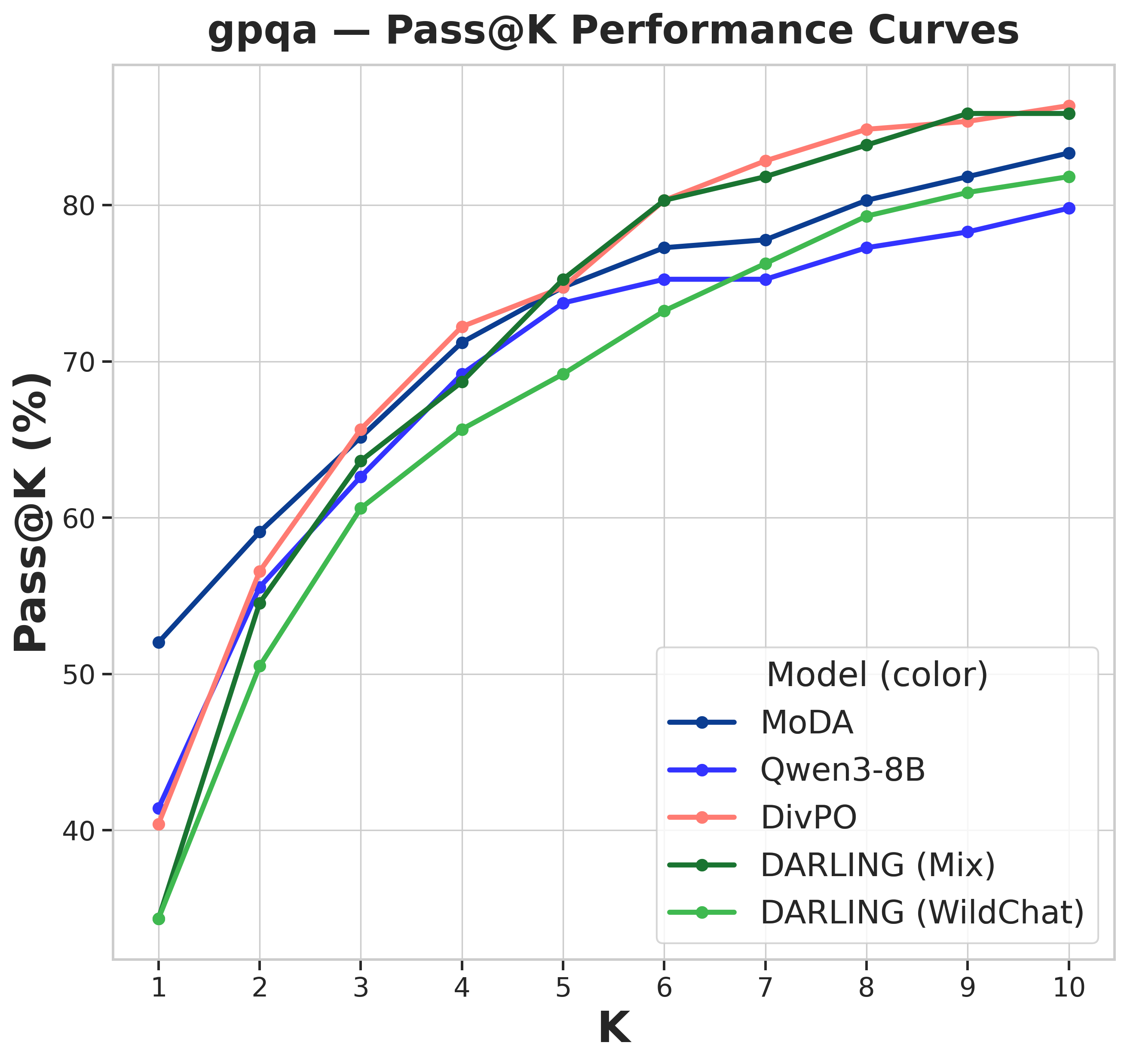}
\end{minipage}\hfill
\begin{minipage}{0.24\textwidth}
  \centering
  \includegraphics[width=\linewidth]{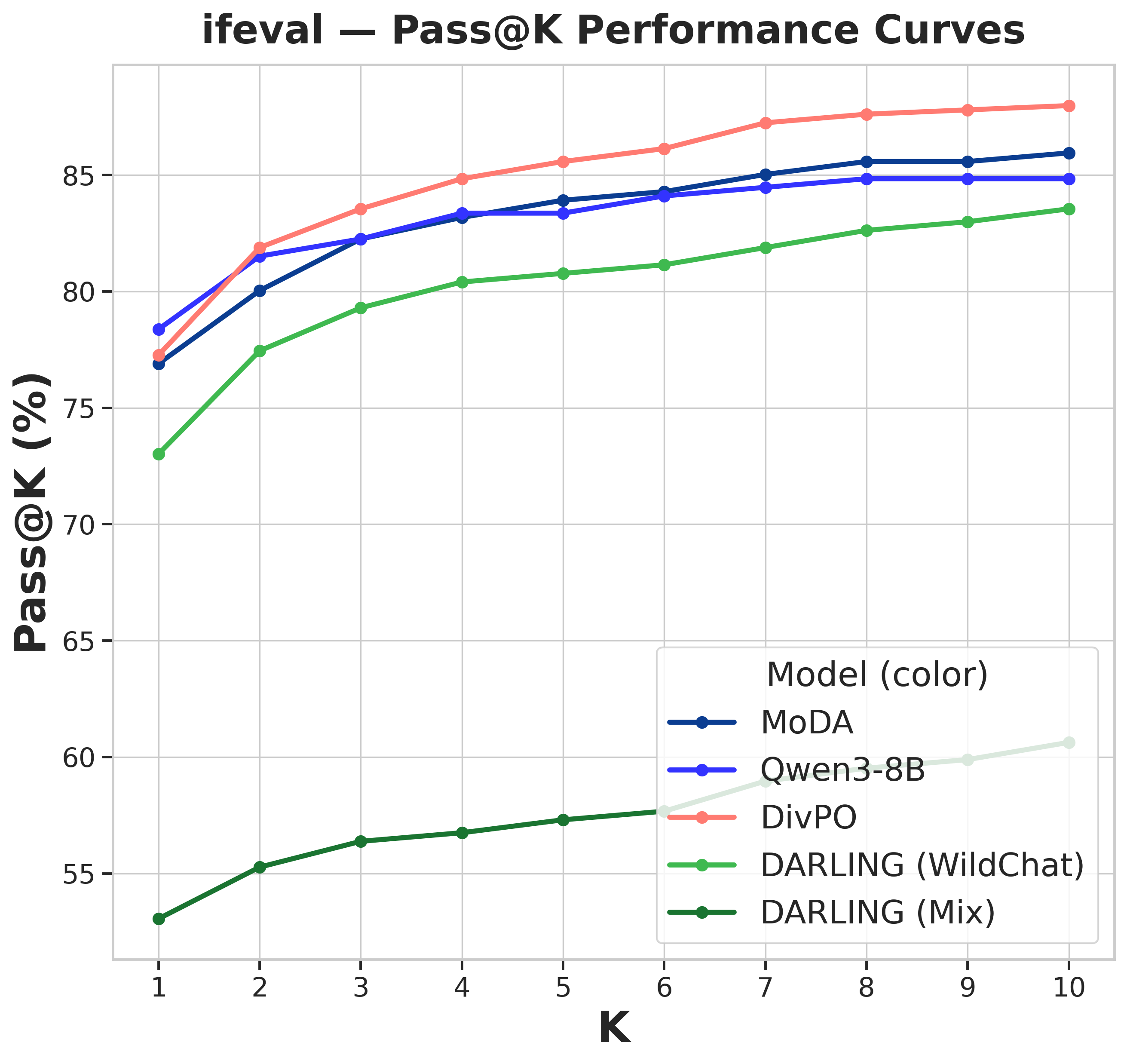}
\end{minipage}
\caption{\textbf{General capability retention: pass@k accuracy.} Qwen3-8B base and thinking disabled models on the general capability suite tasks: BoolQ, MMLU, GPQA, and IFEval.}
\label{fig:general capability pass@k thinking disabled}
\end{figure}

\begin{figure}[H]
\begin{minipage}{0.24\textwidth}
  \centering
  \includegraphics[width=\linewidth]{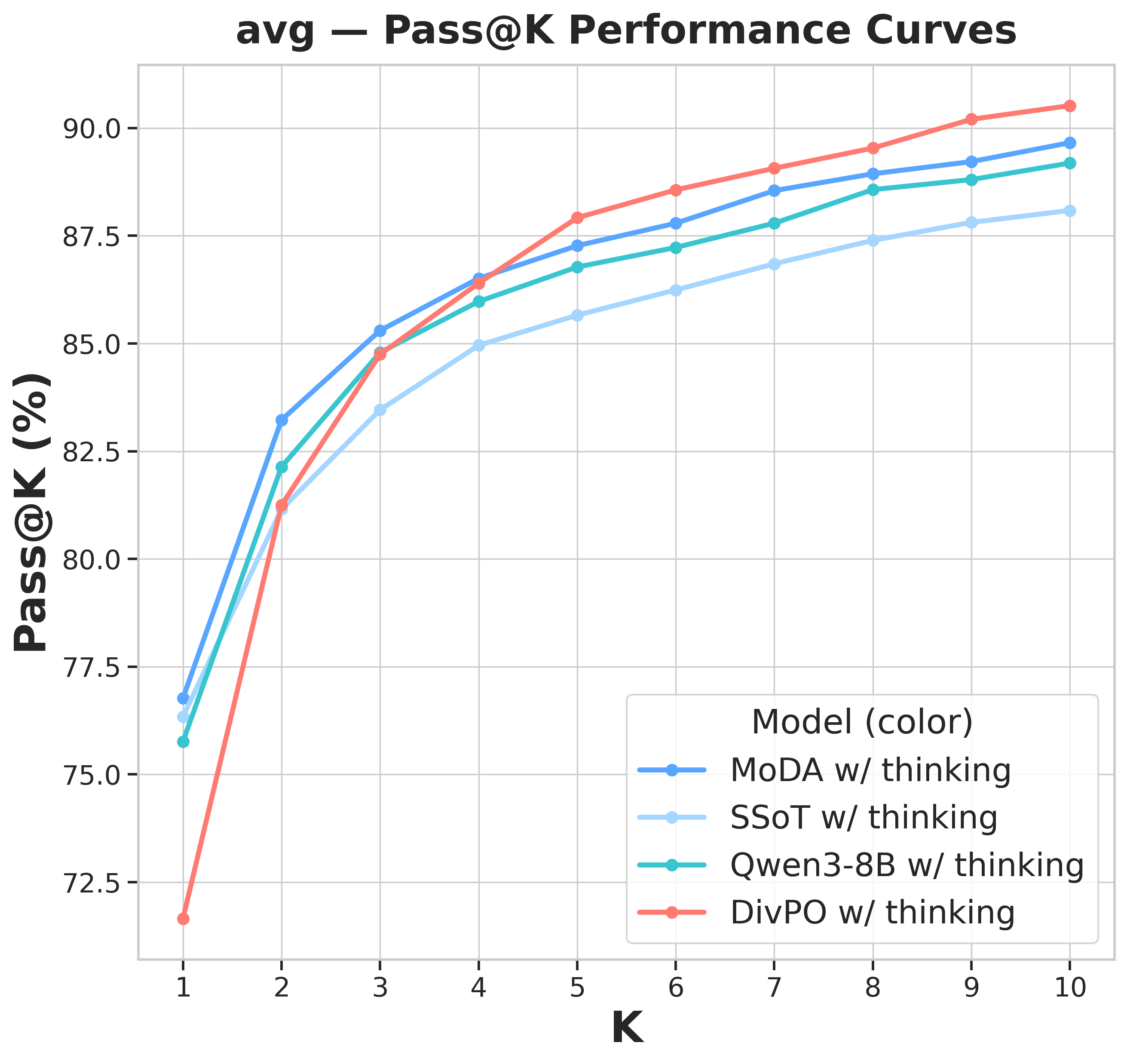}
\end{minipage}\hfill
\centering
\begin{minipage}{0.24\textwidth}
  \centering
  \includegraphics[width=\linewidth]{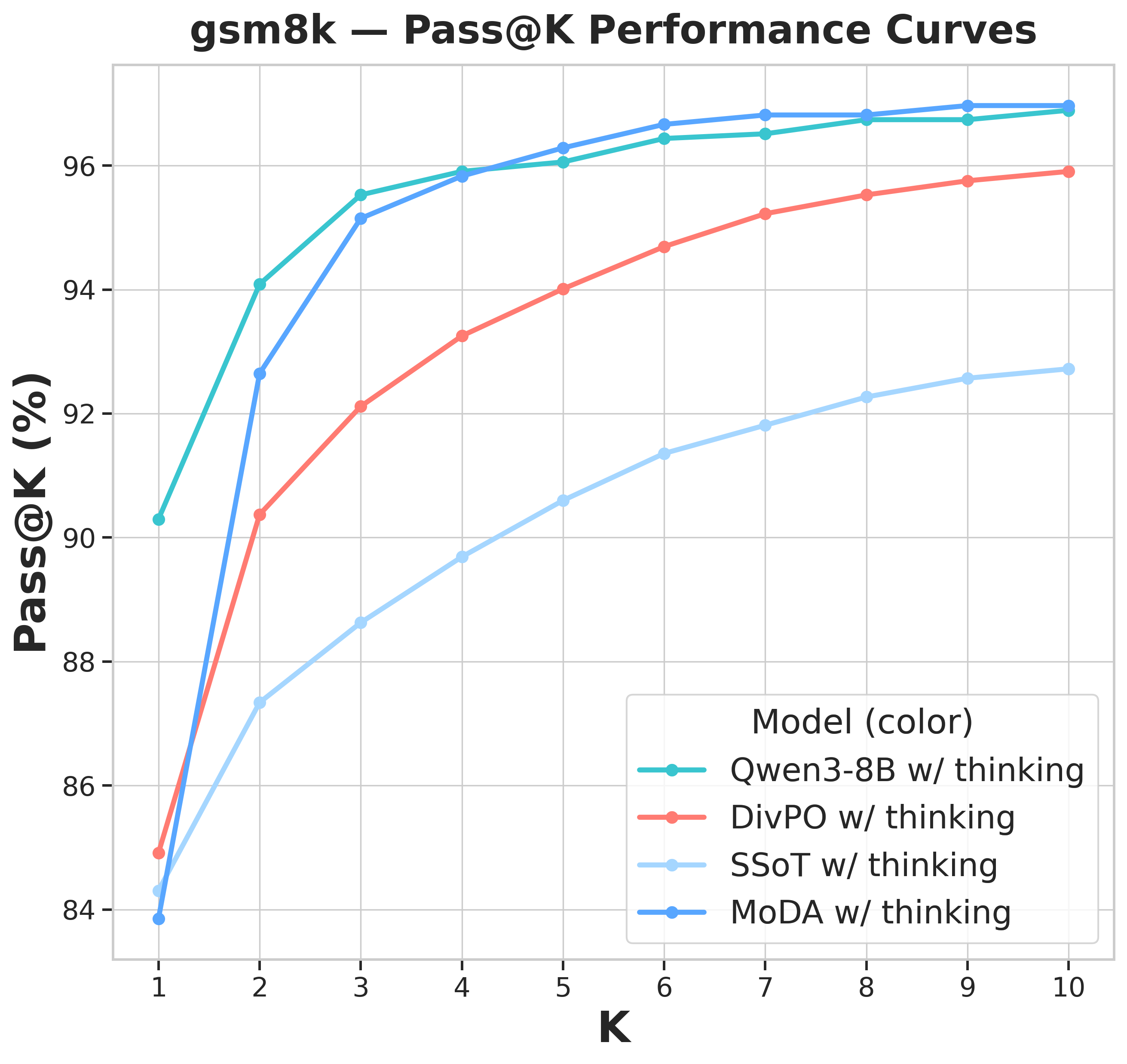}
\end{minipage}\hfill
\begin{minipage}{0.24\textwidth}
  \centering
  \includegraphics[width=\linewidth]{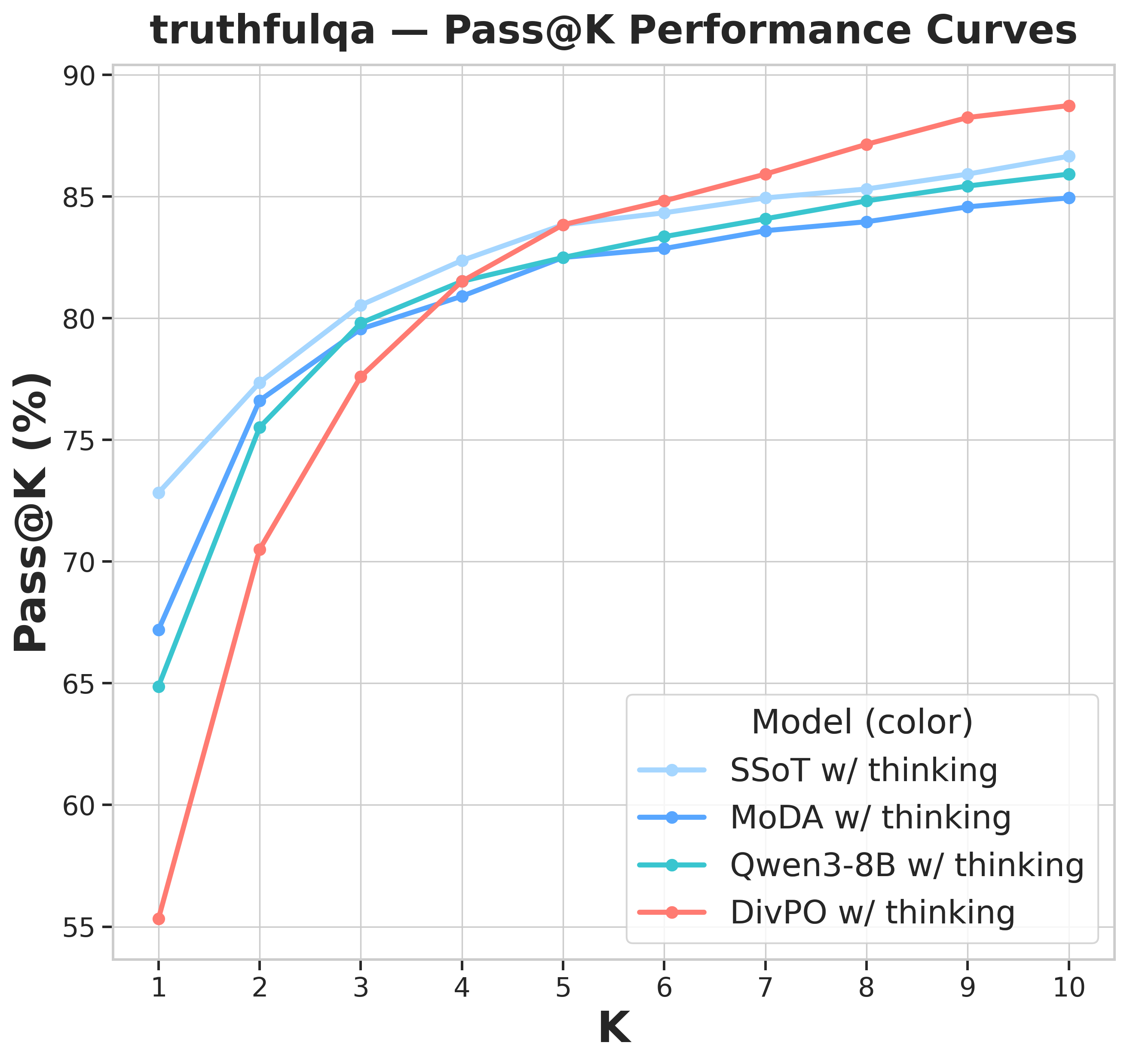}
\end{minipage}\hfill
\begin{minipage}{0.24\textwidth}
  \centering
  \includegraphics[width=\linewidth]{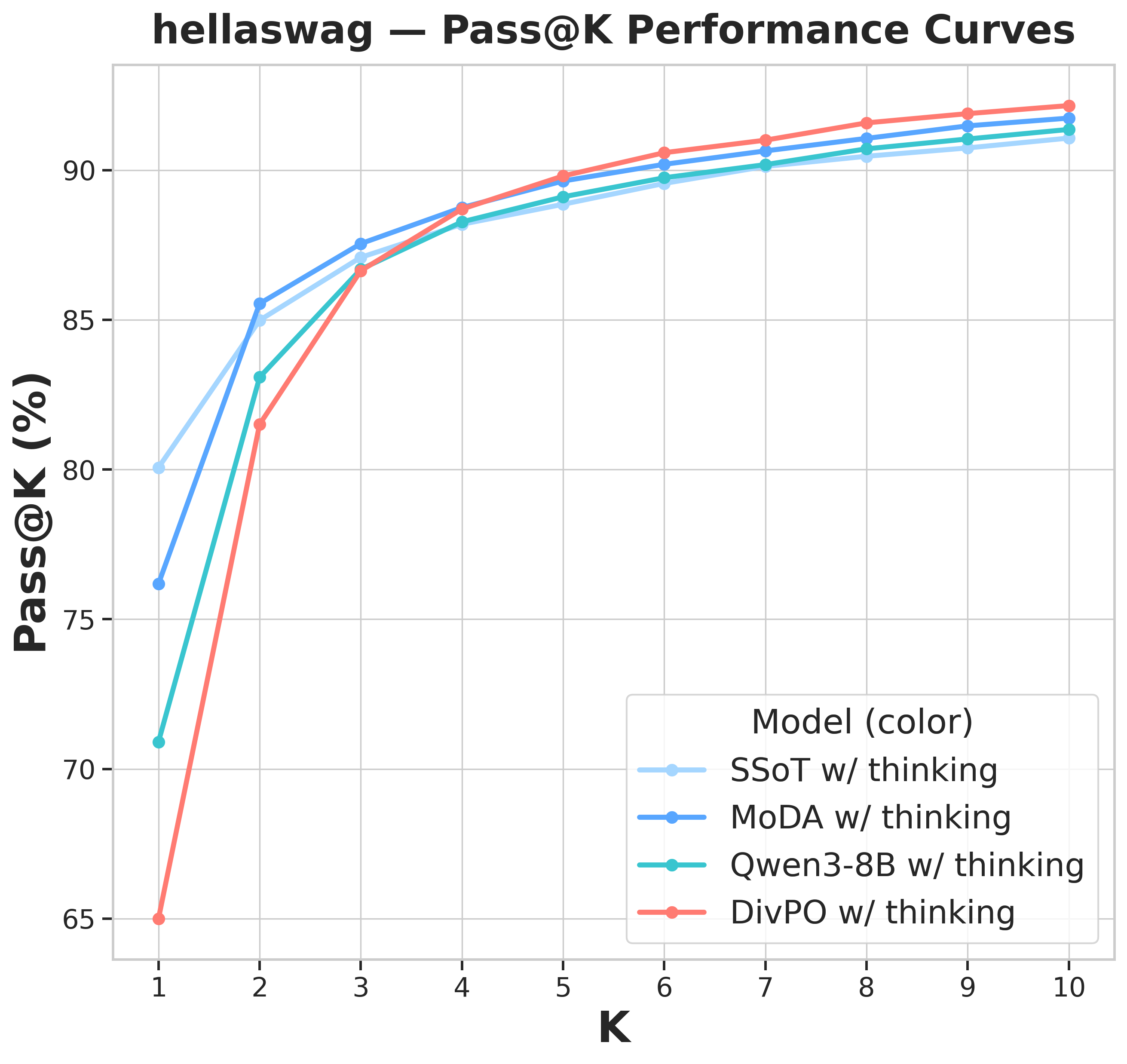}
\end{minipage}
\centering
\begin{minipage}{0.24\textwidth}
  \centering
  \includegraphics[width=\linewidth]{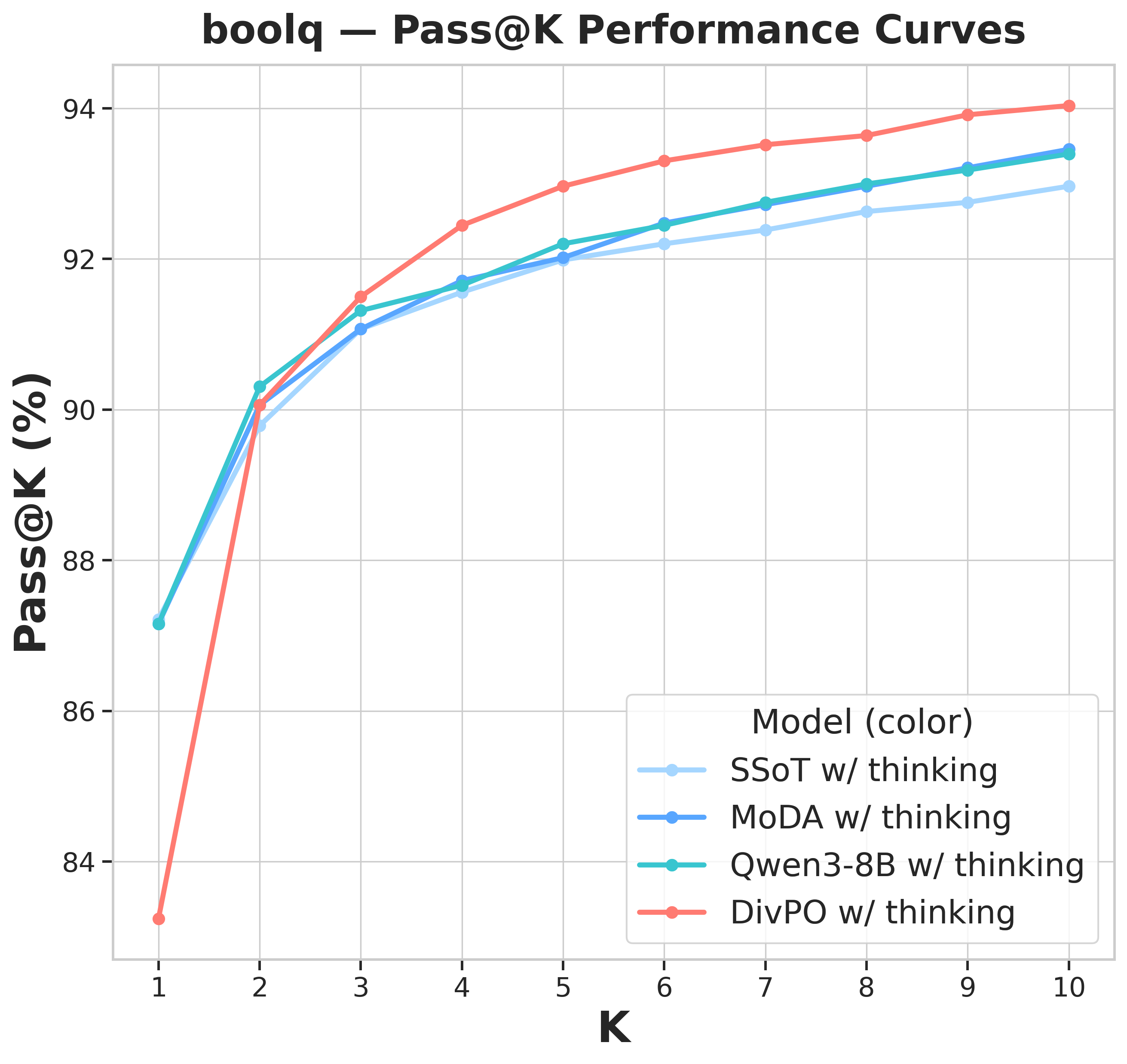}
\end{minipage}\hfill
\centering
\begin{minipage}{0.24\textwidth}
  \centering
  \includegraphics[width=\linewidth]{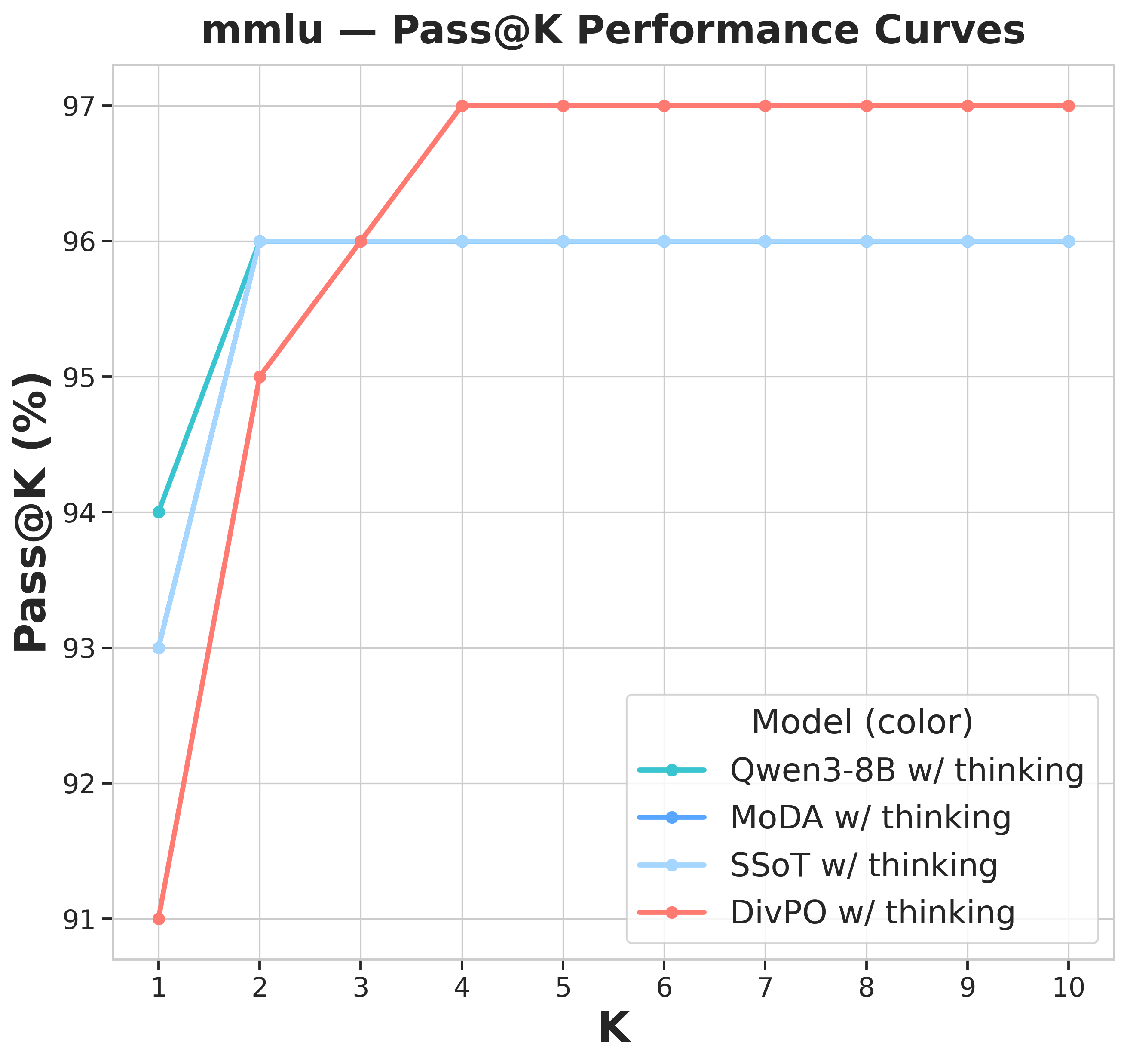}
\end{minipage}\hfill
\begin{minipage}{0.24\textwidth}
  \centering
  \includegraphics[width=\linewidth]{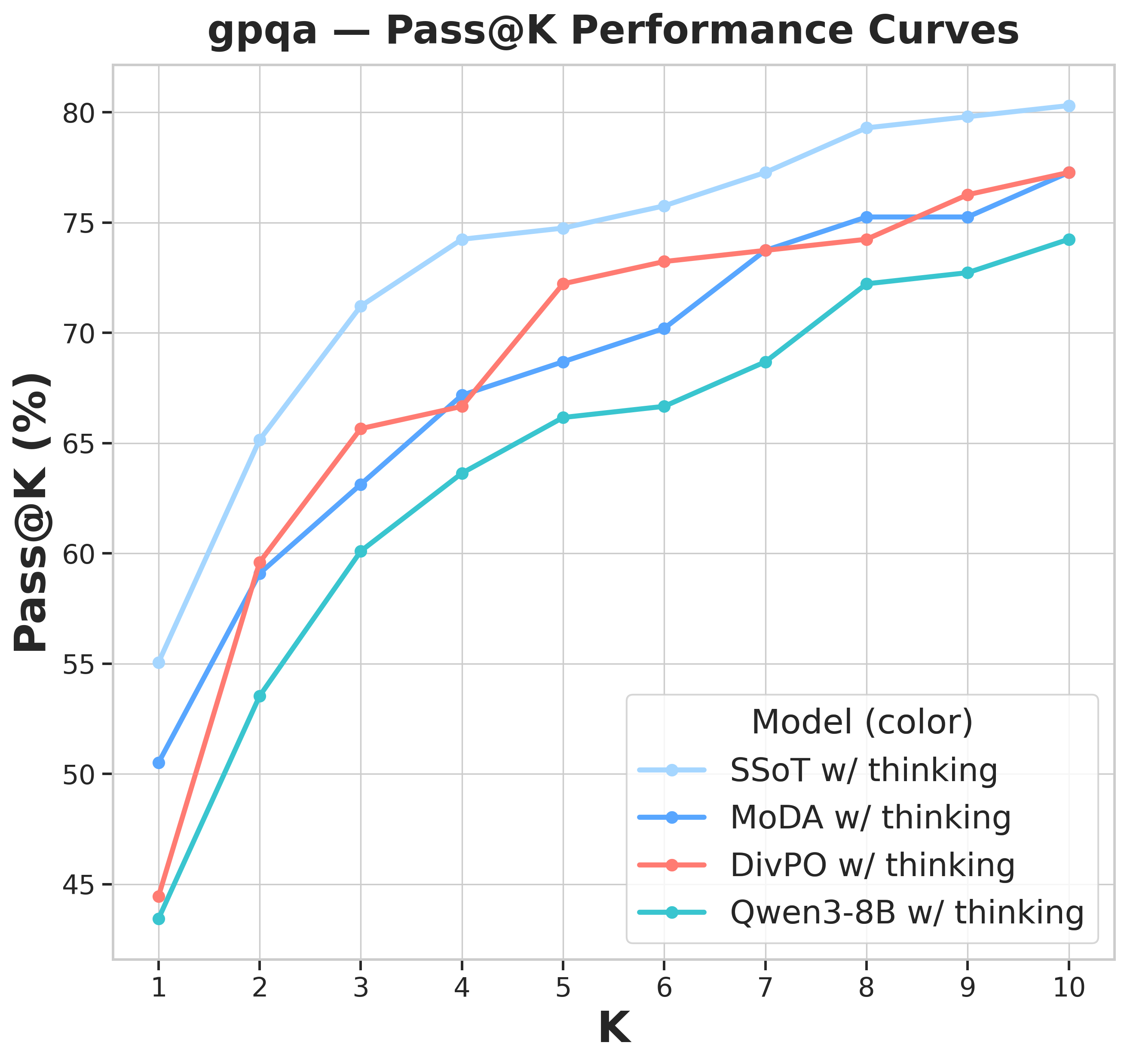}
\end{minipage}\hfill
\begin{minipage}{0.24\textwidth}
  \centering
  \includegraphics[width=\linewidth]{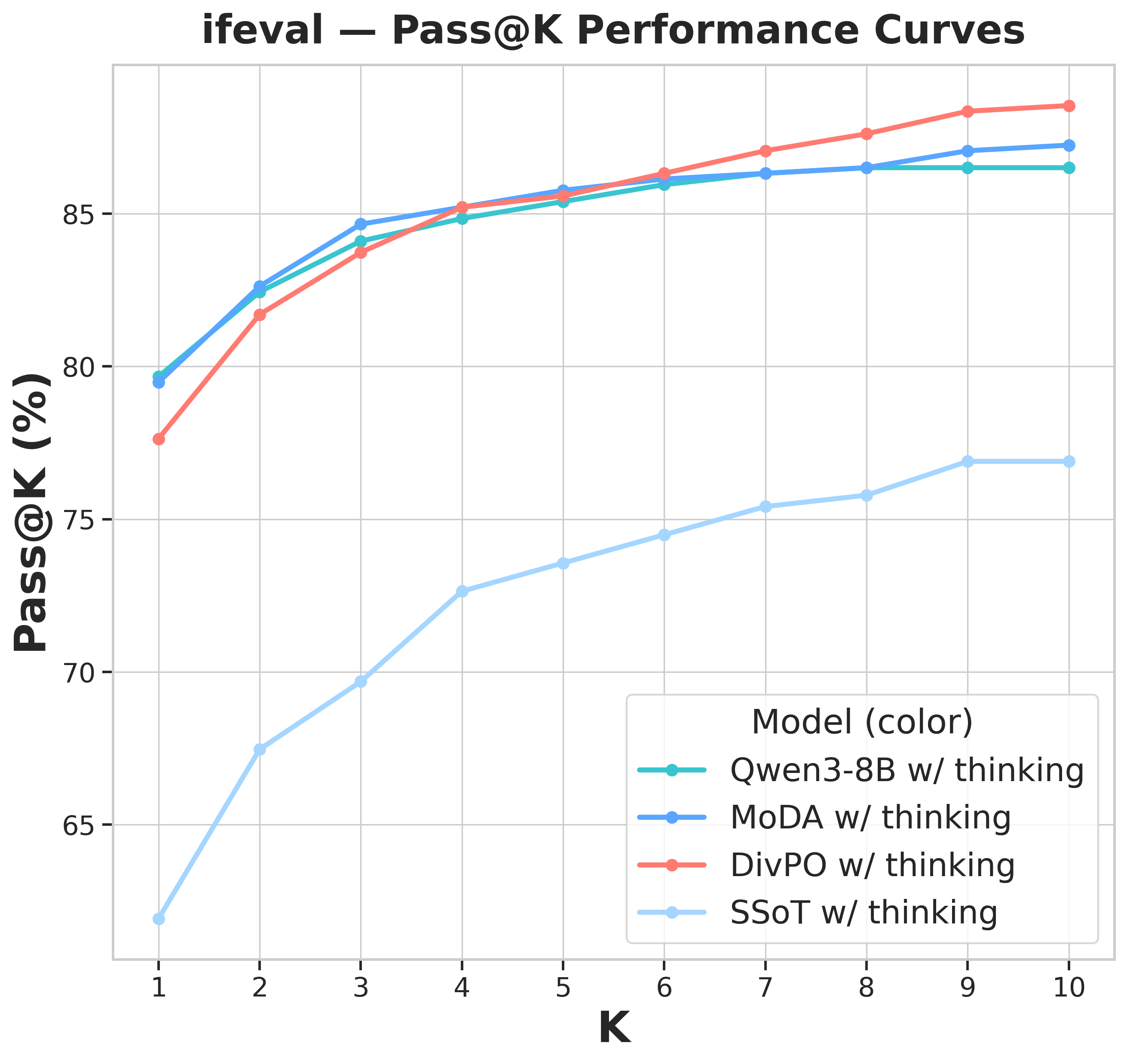}
\end{minipage}
\caption{\textbf{General capability retention: pass@k accuracy.} Qwen3-8B base and thinking enabled models.}
\label{fig:general capability pass@k thinking enabled}
\end{figure}

\begin{figure}[H]
\begin{minipage}{0.24\textwidth}
  \centering
  \includegraphics[width=\linewidth]{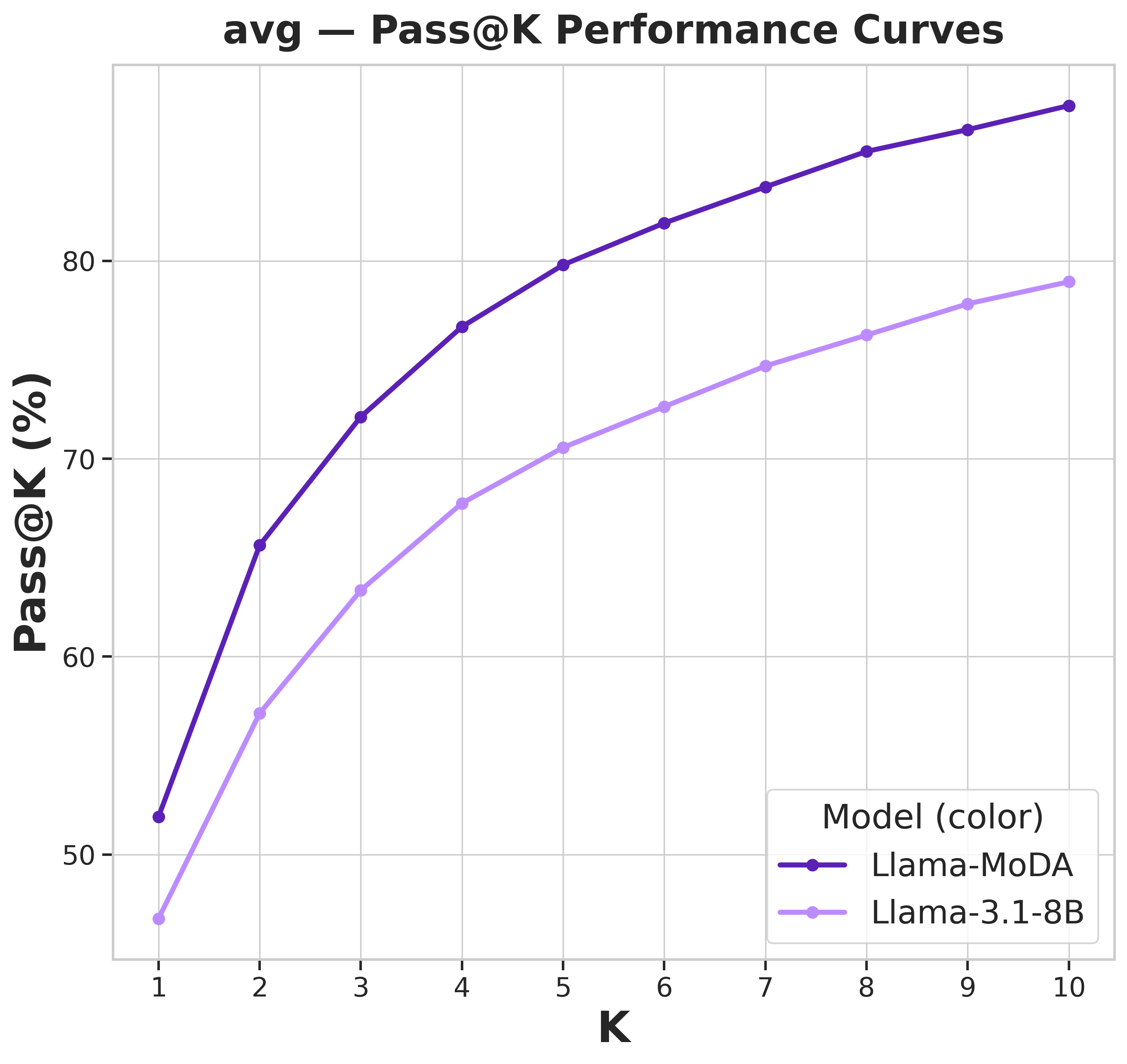}
\end{minipage}\hfill
\centering
\begin{minipage}{0.24\textwidth}
  \centering
  \includegraphics[width=\linewidth]{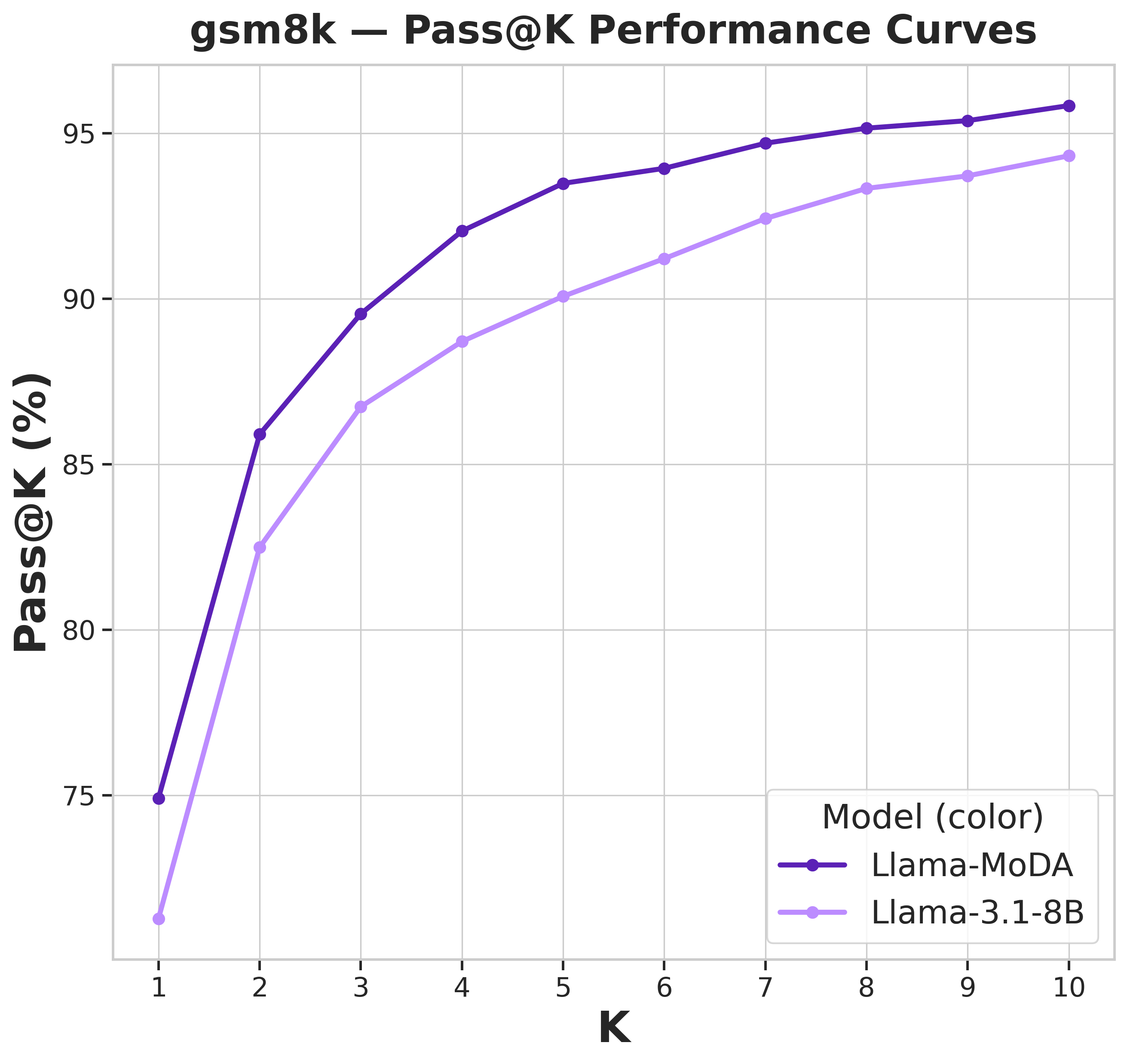}
\end{minipage}\hfill
\begin{minipage}{0.24\textwidth}
  \centering
  \includegraphics[width=\linewidth]{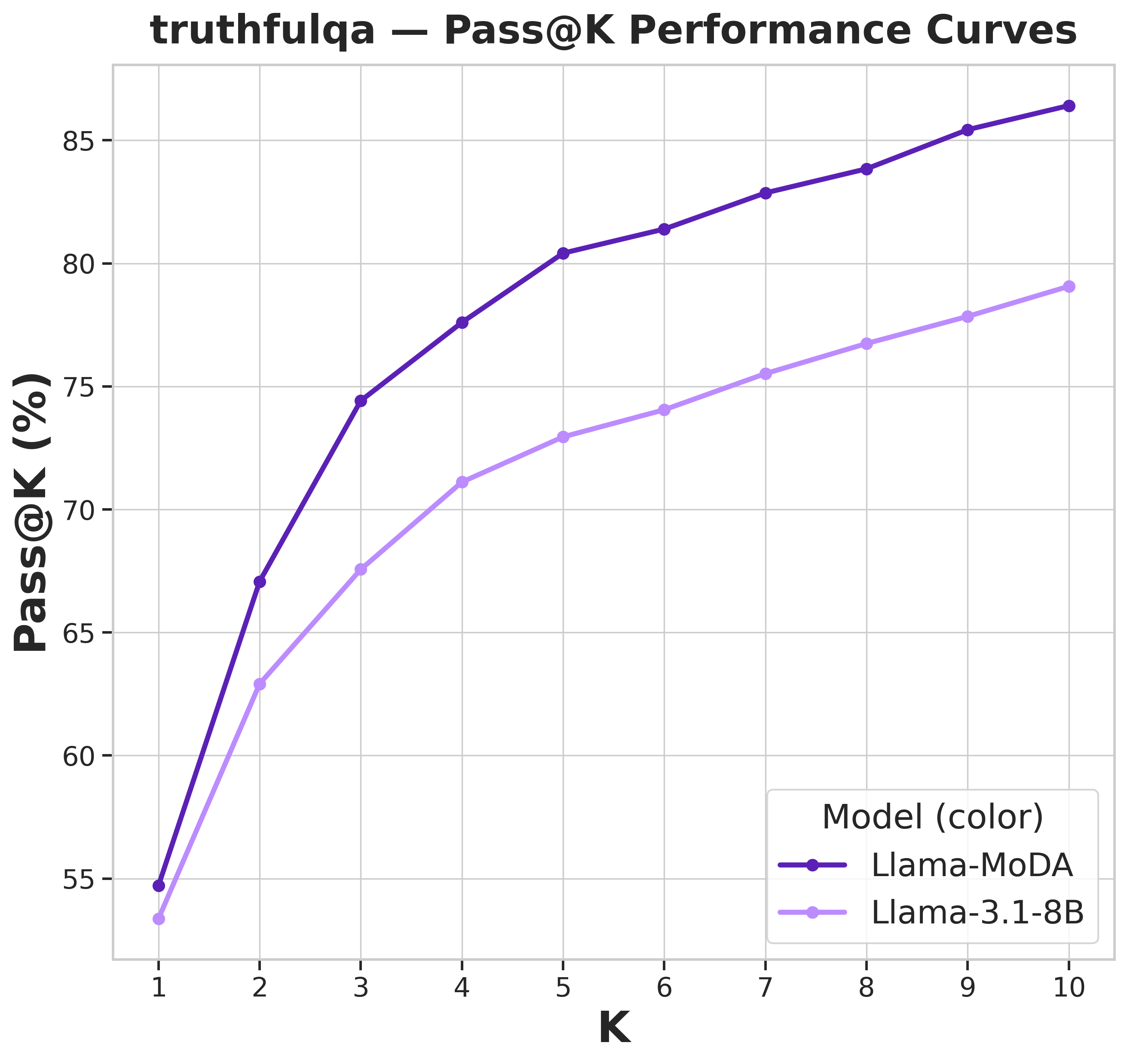}
\end{minipage}\hfill
\begin{minipage}{0.24\textwidth}
  \centering
  \includegraphics[width=\linewidth]{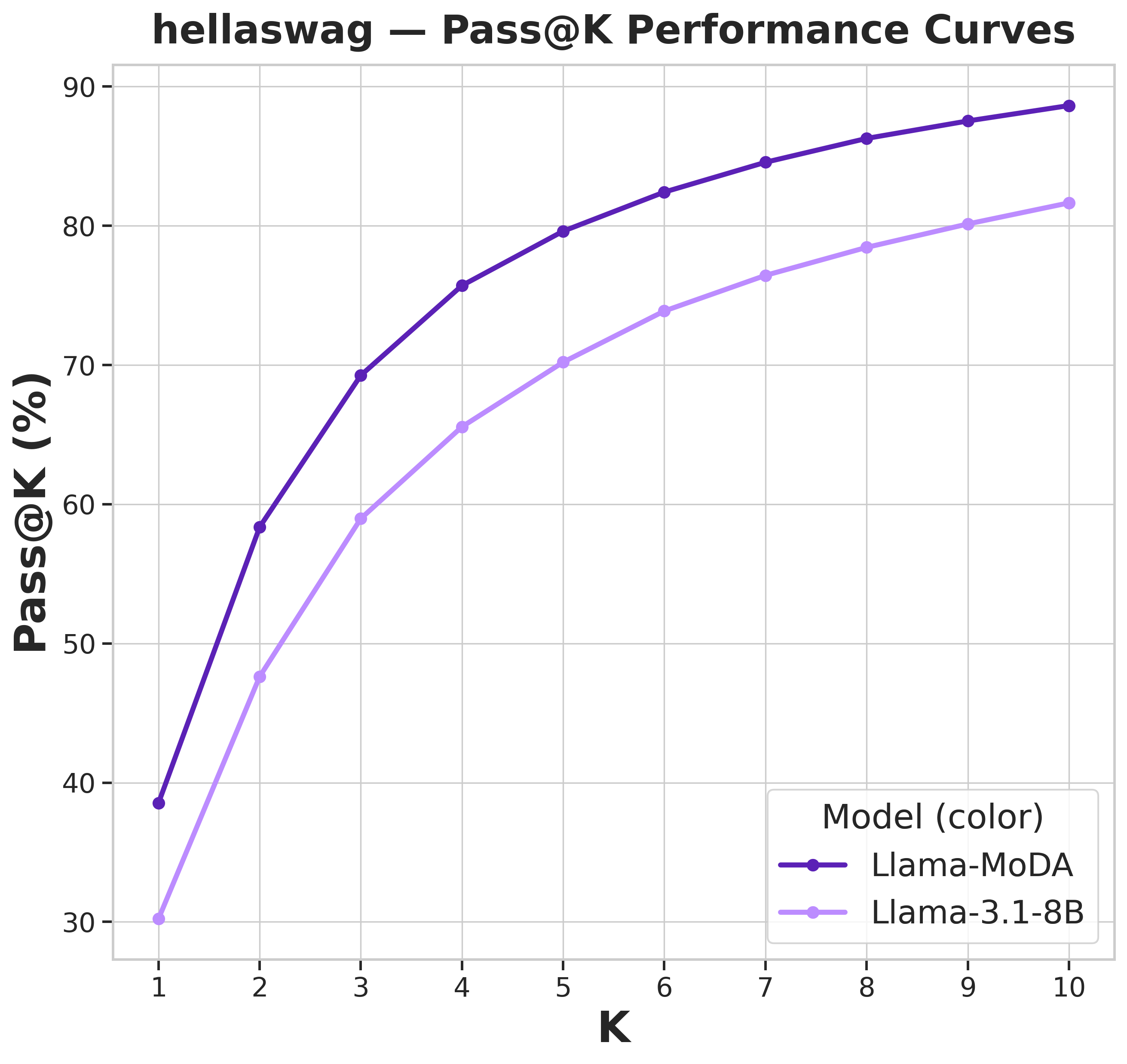}
\end{minipage}
\centering
\begin{minipage}{0.24\textwidth}
  \centering
  \includegraphics[width=\linewidth]{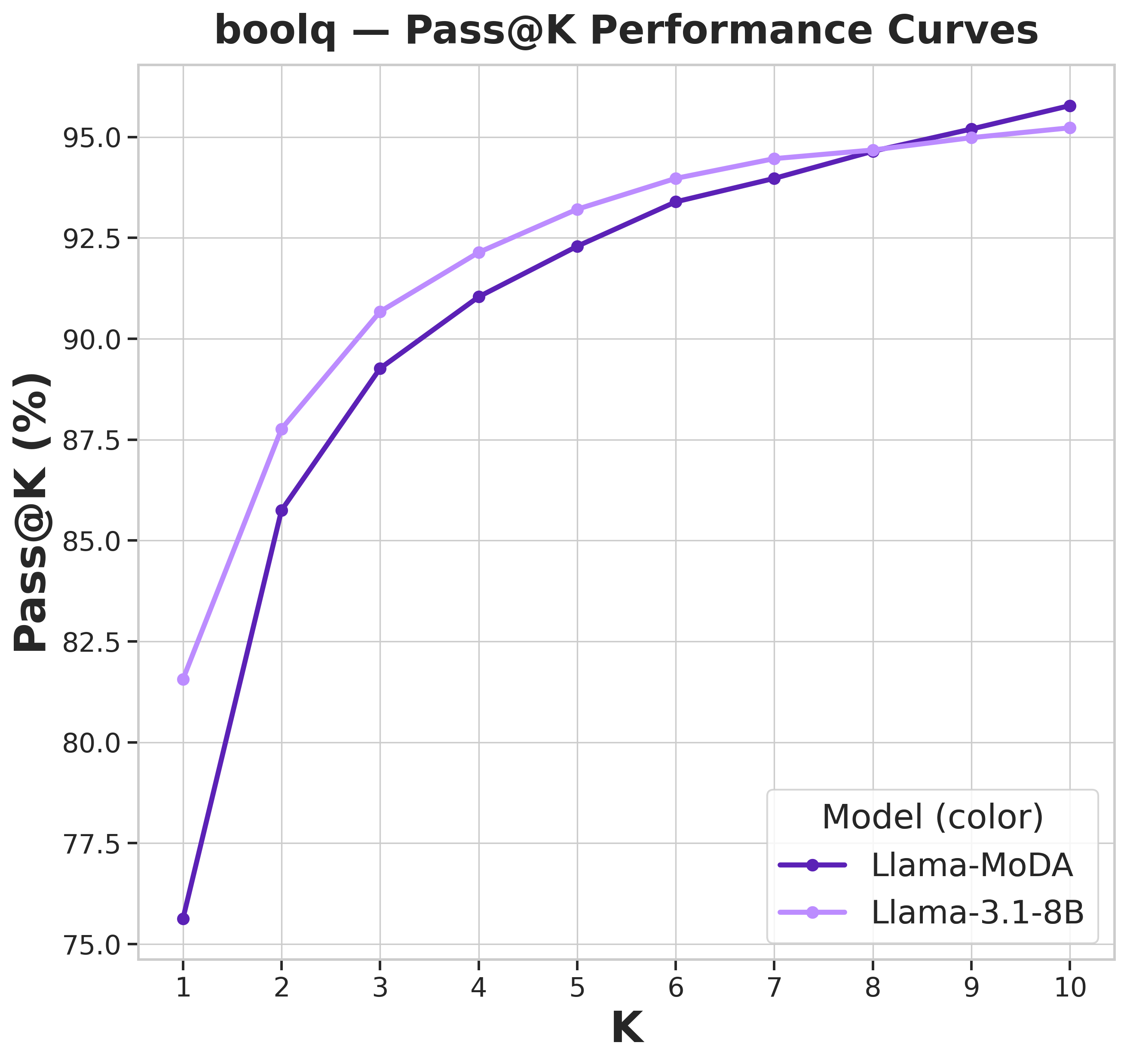}
\end{minipage}\hfill
\centering
\begin{minipage}{0.24\textwidth}
  \centering
  \includegraphics[width=\linewidth]{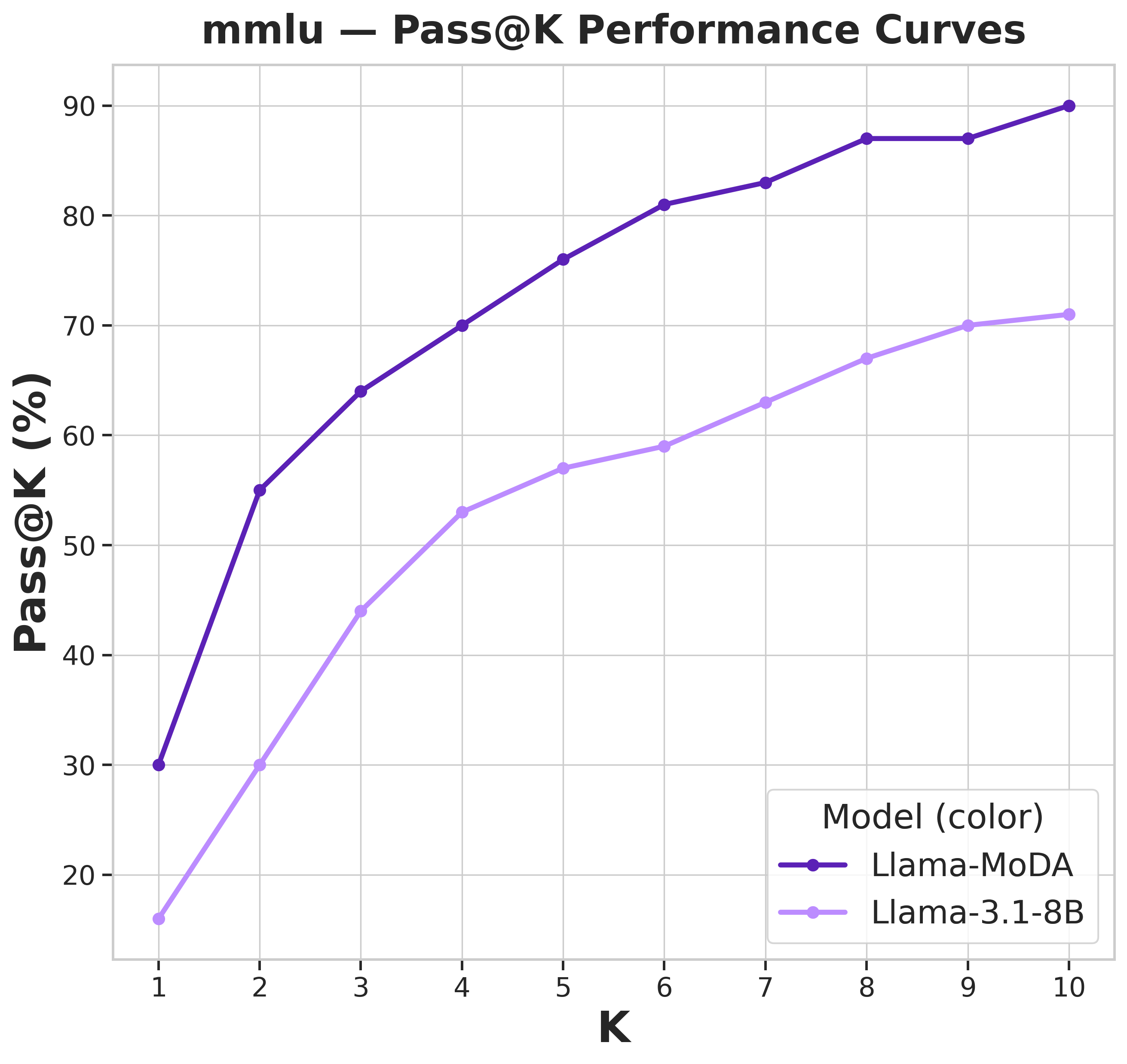}
\end{minipage}\hfill
\begin{minipage}{0.24\textwidth}
  \centering
  \includegraphics[width=\linewidth]{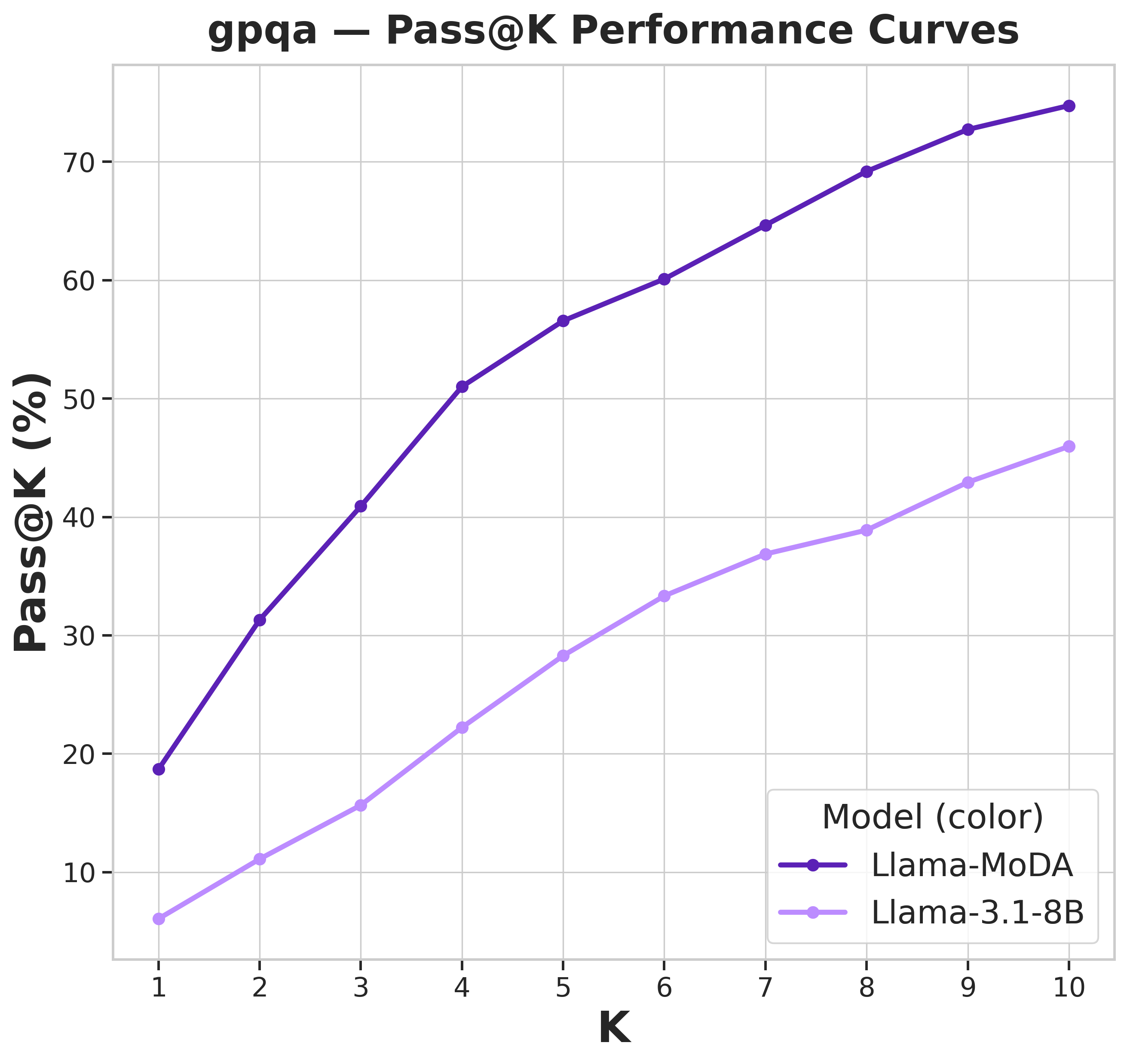}
\end{minipage}\hfill
\begin{minipage}{0.24\textwidth}
  \centering
  \includegraphics[width=\linewidth]{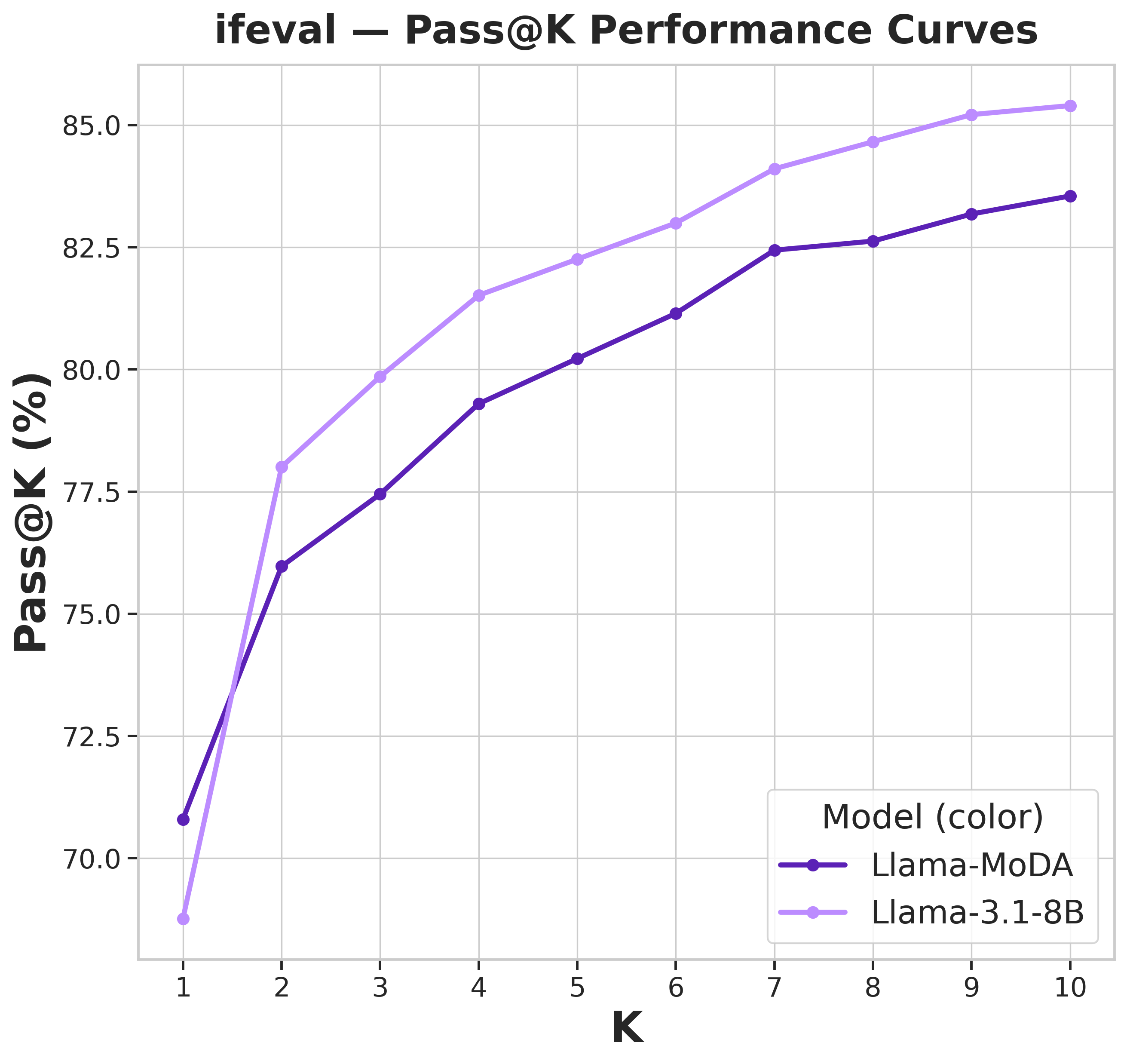}
\end{minipage}
\caption{\textbf{General capability retention: pass@k accuracy.} Llama-3.1-8B base and thinking disabled models.}
\label{fig:general capability pass@k llama}
\end{figure}

\begin{figure}[H]
\begin{minipage}{0.24\textwidth}
  \centering
  \includegraphics[width=\linewidth]{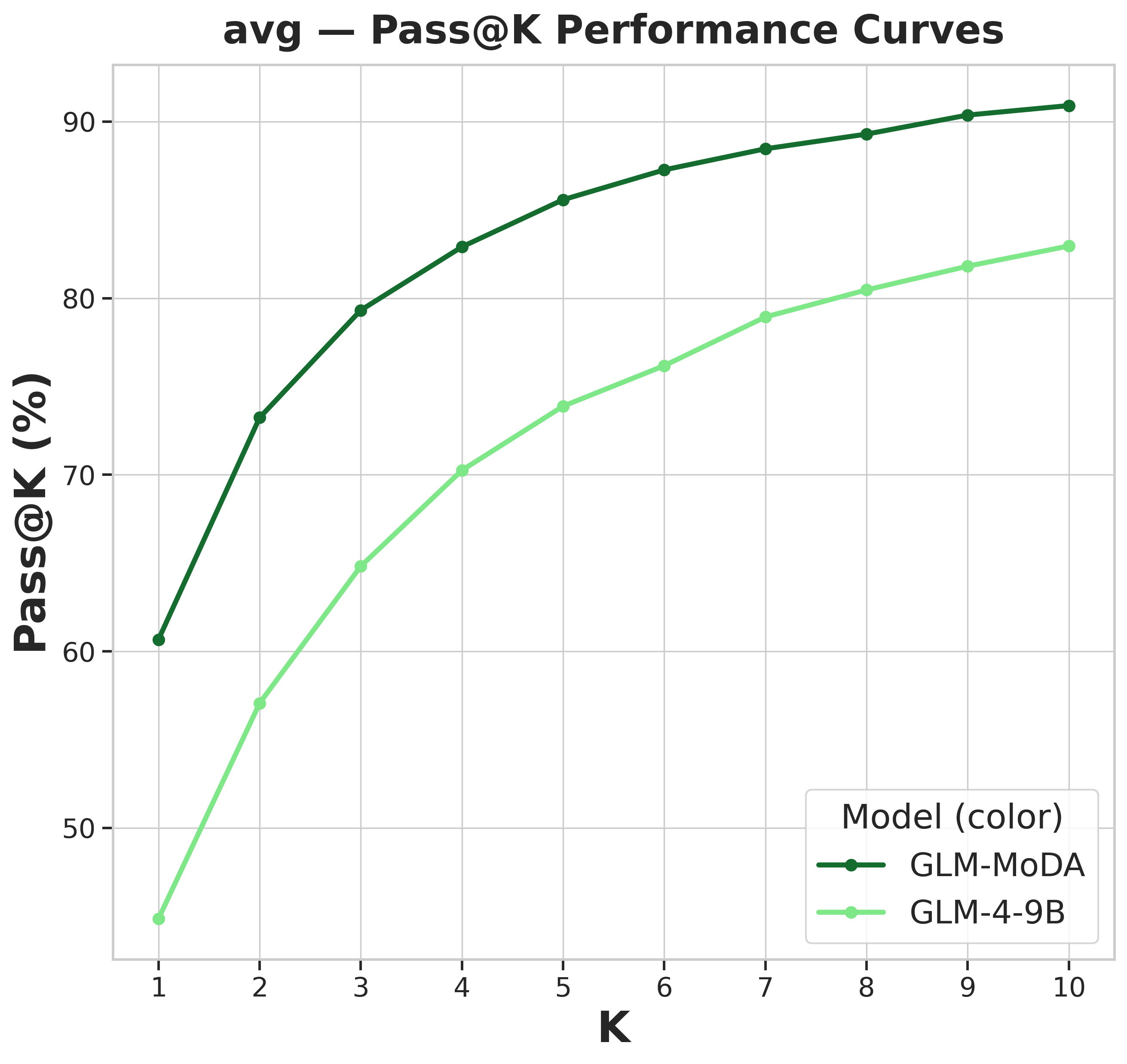}
\end{minipage}\hfill
\centering
\begin{minipage}{0.24\textwidth}
  \centering
  \includegraphics[width=\linewidth]{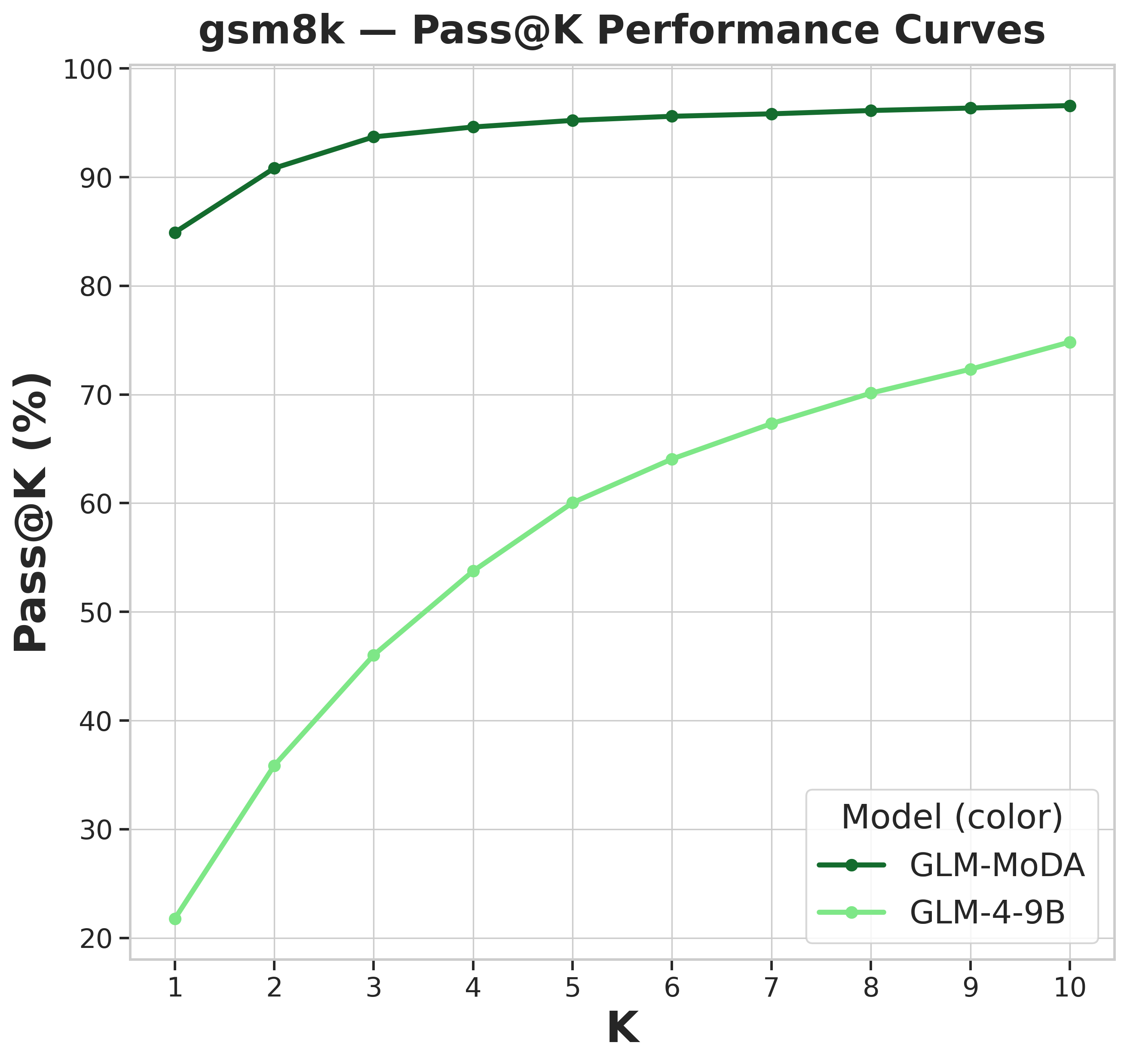}
\end{minipage}\hfill
\begin{minipage}{0.24\textwidth}
  \centering
  \includegraphics[width=\linewidth]{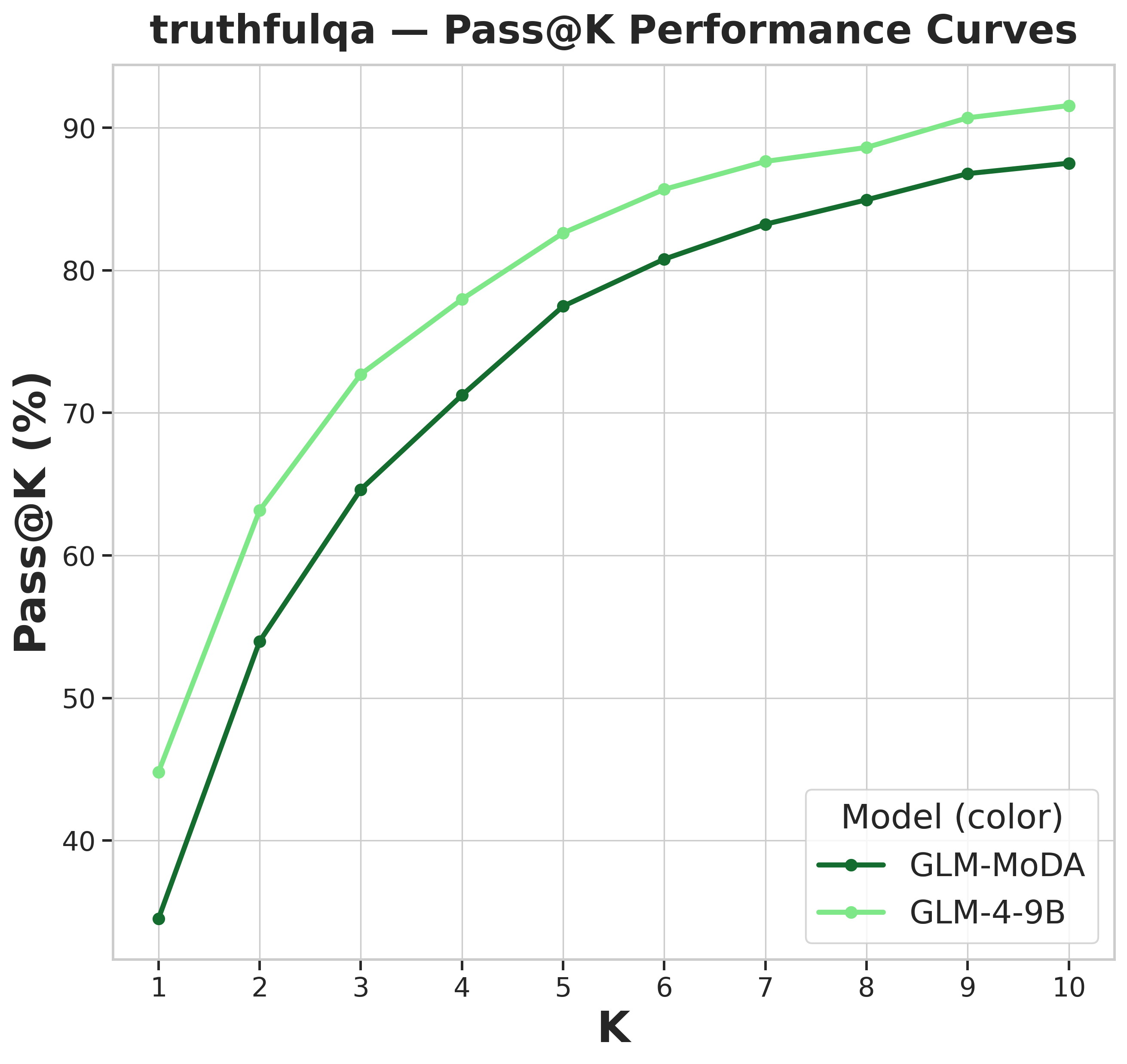}
\end{minipage}\hfill
\begin{minipage}{0.24\textwidth}
  \centering
  \includegraphics[width=\linewidth]{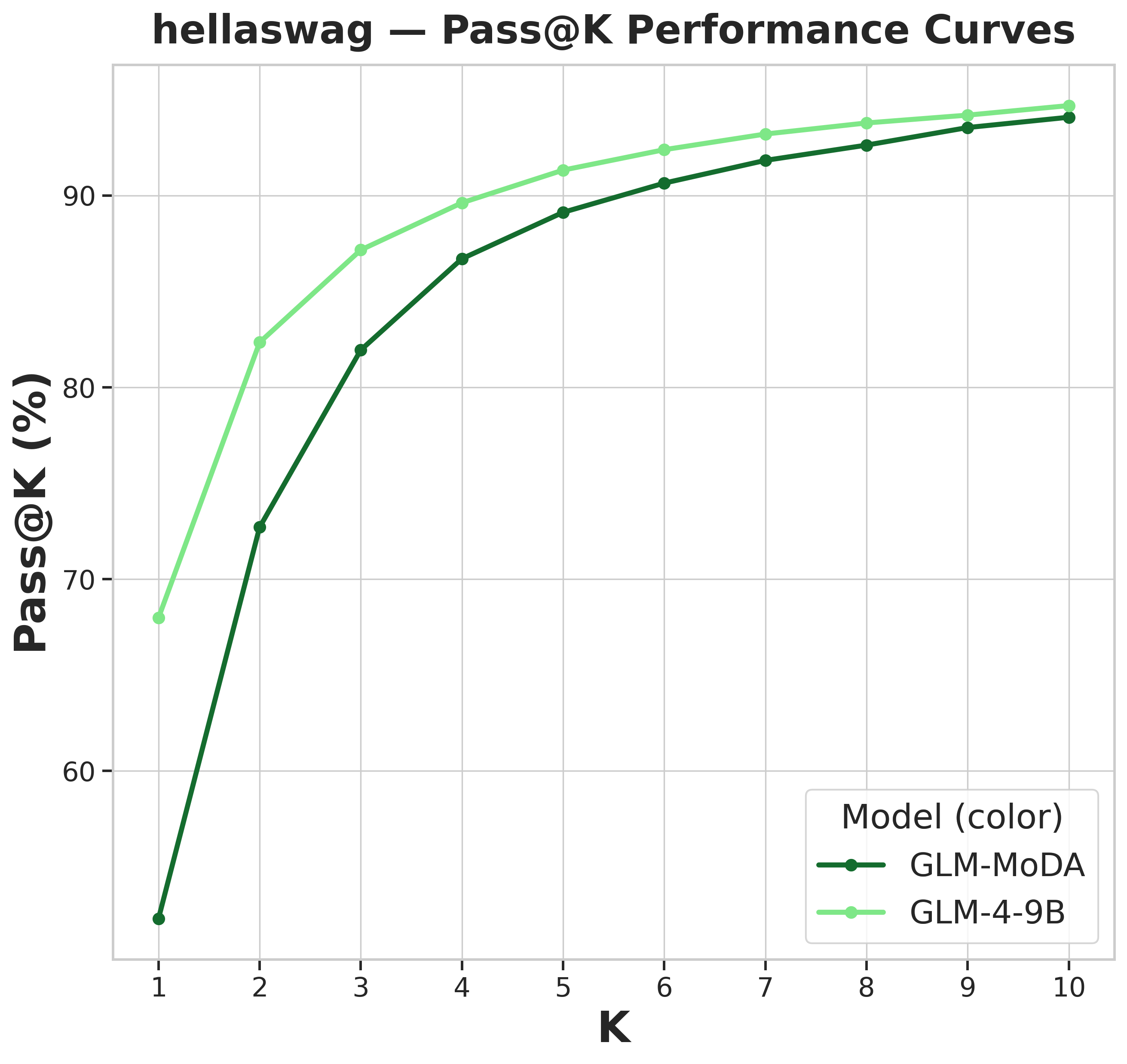}
\end{minipage}
\centering
\begin{minipage}{0.24\textwidth}
  \centering
  \includegraphics[width=\linewidth]{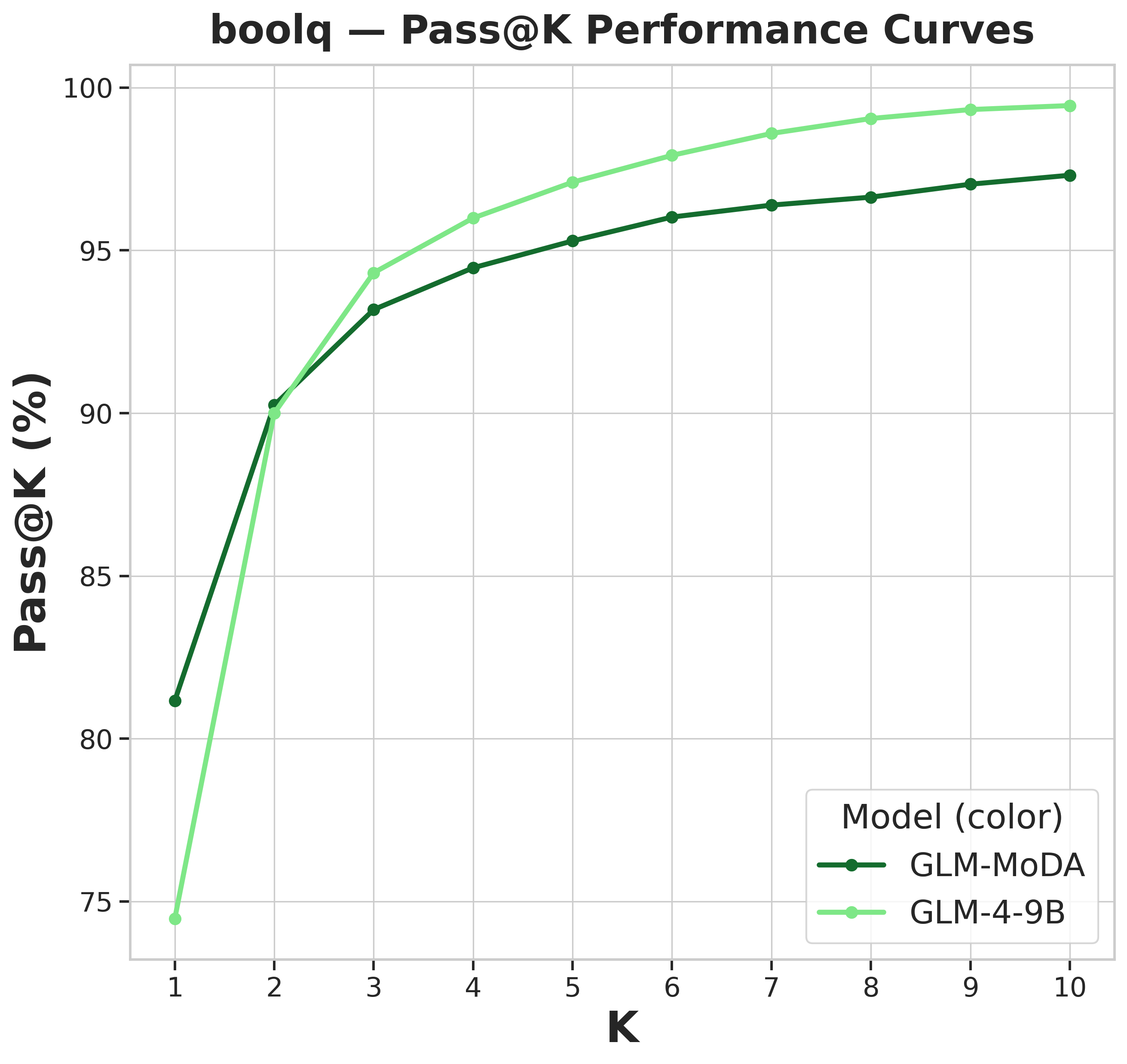}
\end{minipage}\hfill
\centering
\begin{minipage}{0.24\textwidth}
  \centering
  \includegraphics[width=\linewidth]{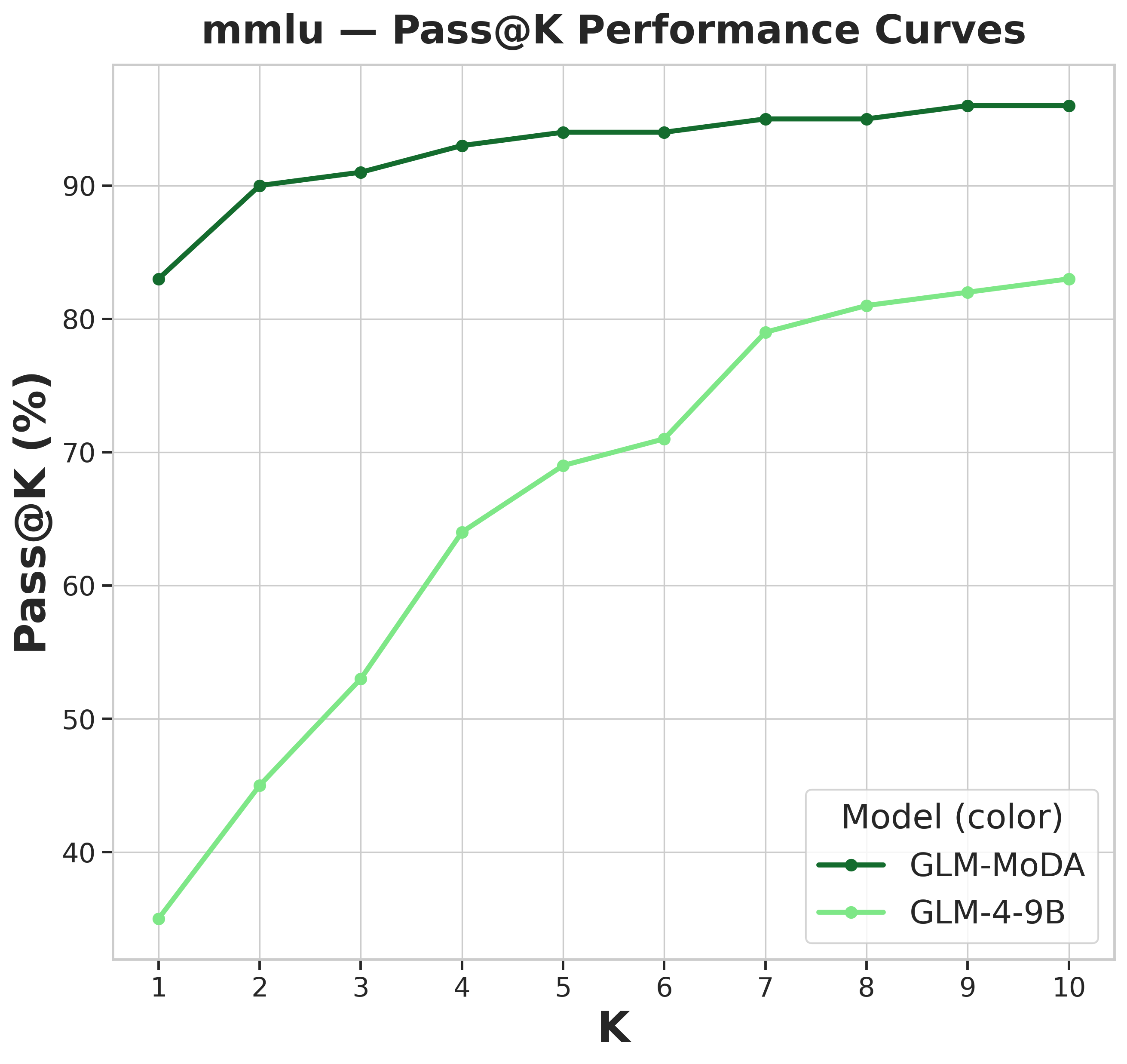}
\end{minipage}\hfill
\begin{minipage}{0.24\textwidth}
  \centering
  \includegraphics[width=\linewidth]{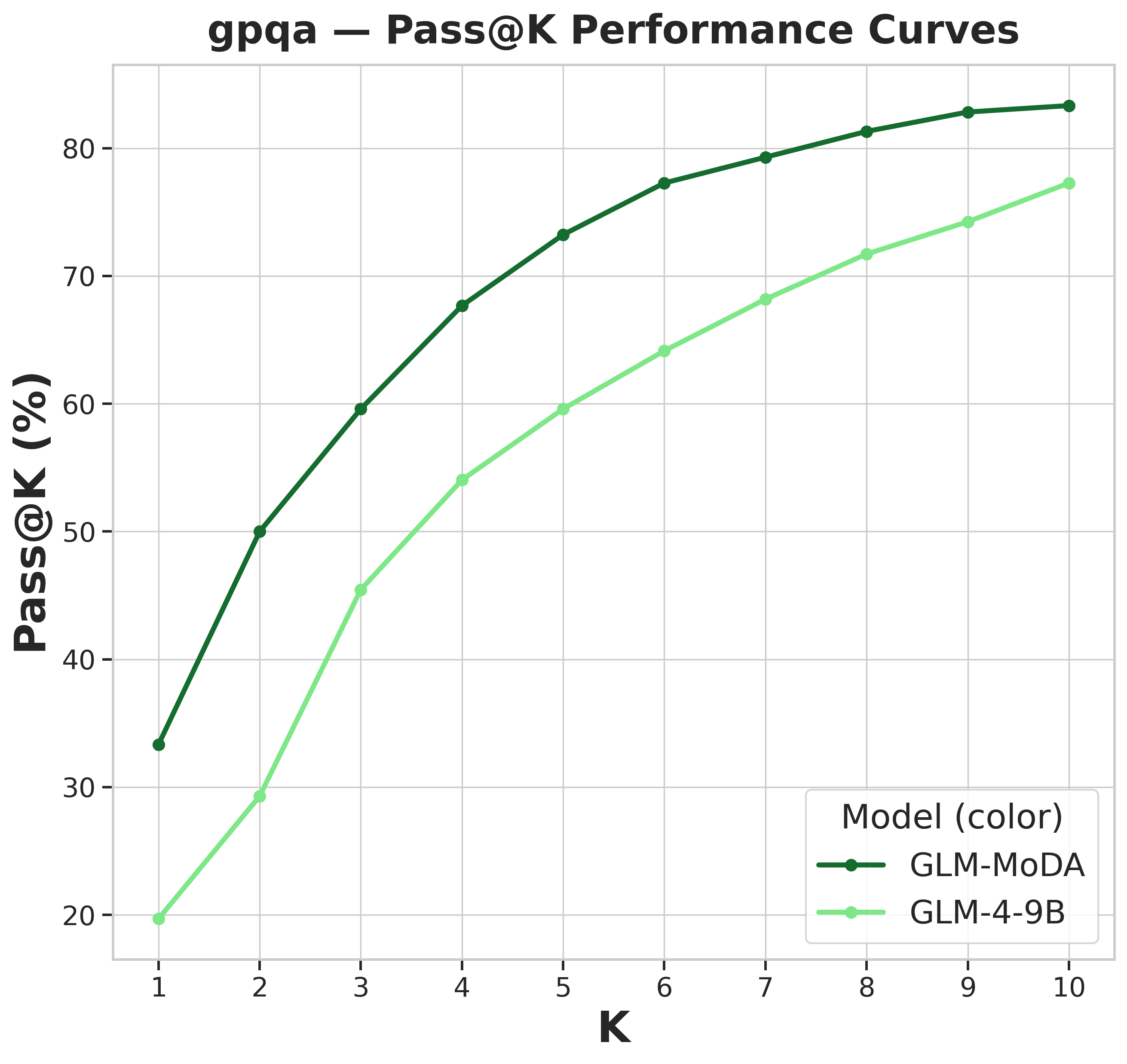}
\end{minipage}\hfill
\begin{minipage}{0.24\textwidth}
  \centering
  \includegraphics[width=\linewidth]{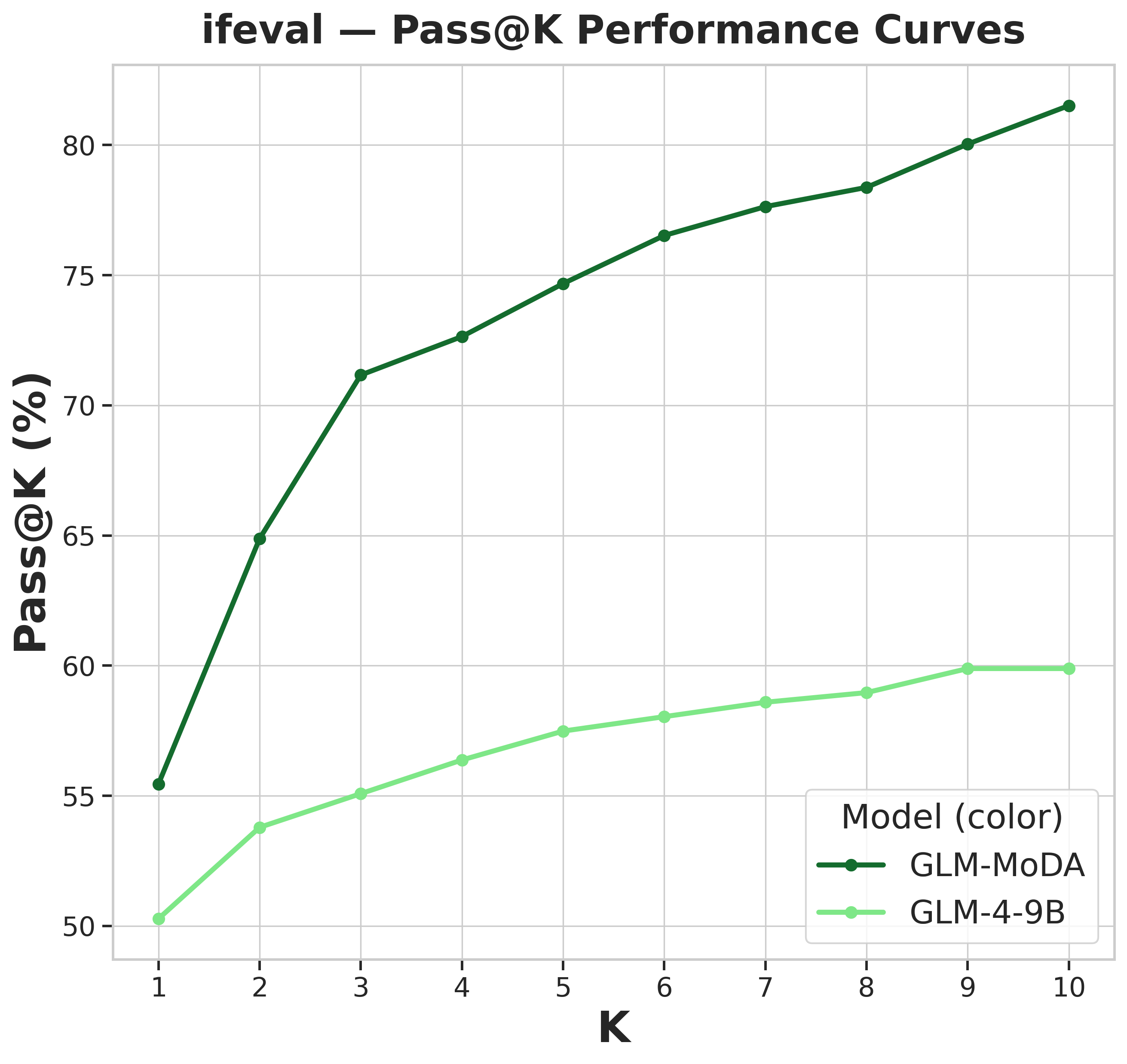}
\end{minipage}
\caption{\textbf{General capability retention: pass@k accuracy.} GLM-4-9B base and thinking disabled models.}
\label{fig:general capability pass@k glm}
\end{figure}

\subsection{Ablation Results}
\label{appx: ablation results}
We show the mean benchmark accuracy, under both standard and diverse decoding mode, on the general capability suite, and Infinite-Chat response diversity on the held out test set for the ablation study models.
\input{notes_arxiv/table/ablation_table}

\begin{figure}
\begin{minipage}{0.32\textwidth}
  \centering
  \includegraphics[width=\linewidth]{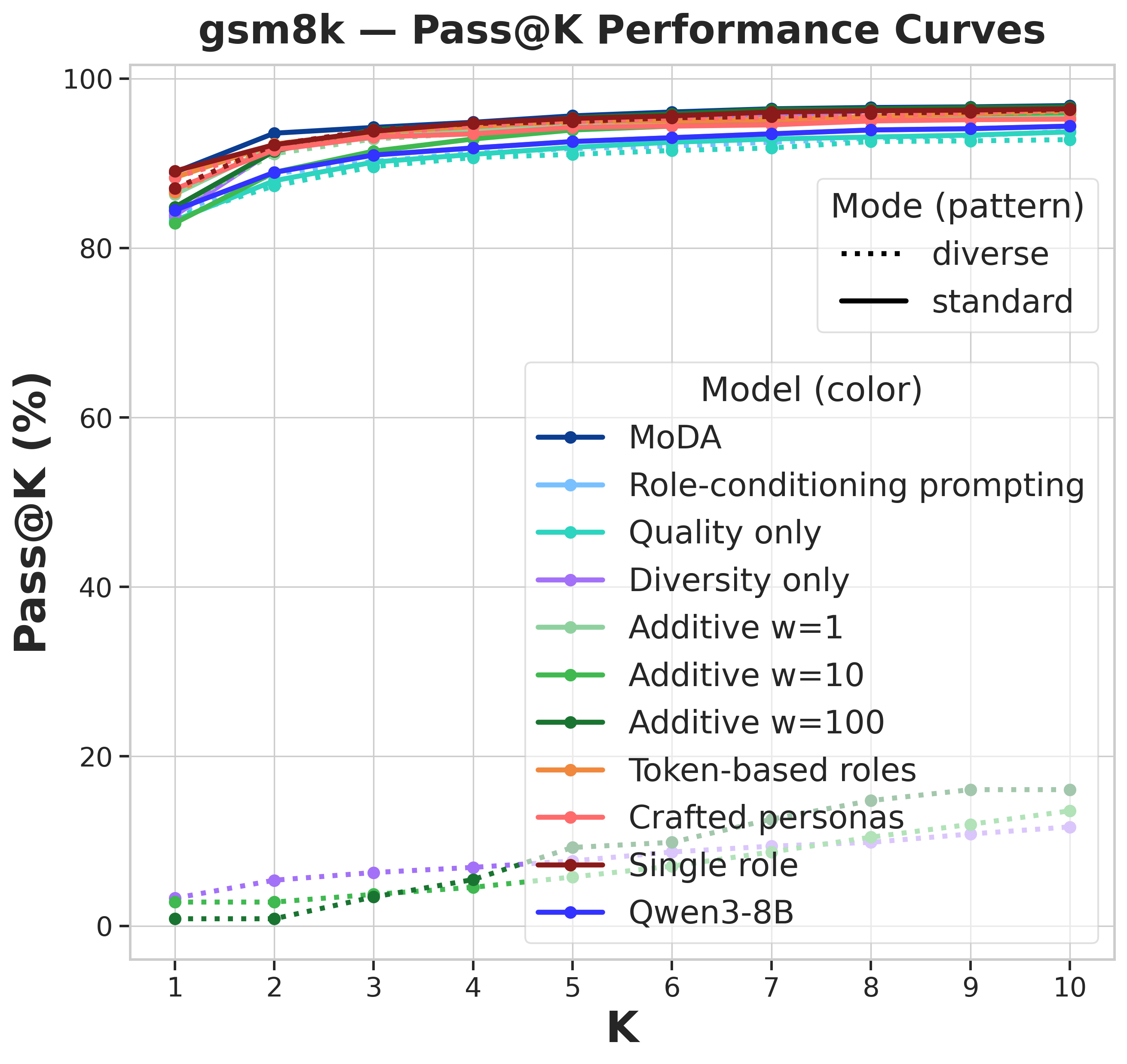}
  \small Pass@k, GSM8K
  \label{fig: ablation pass@k GSM8K}
\end{minipage}\hfill
\centering
\begin{minipage}{0.32\textwidth}
  \centering
  \includegraphics[width=\linewidth]{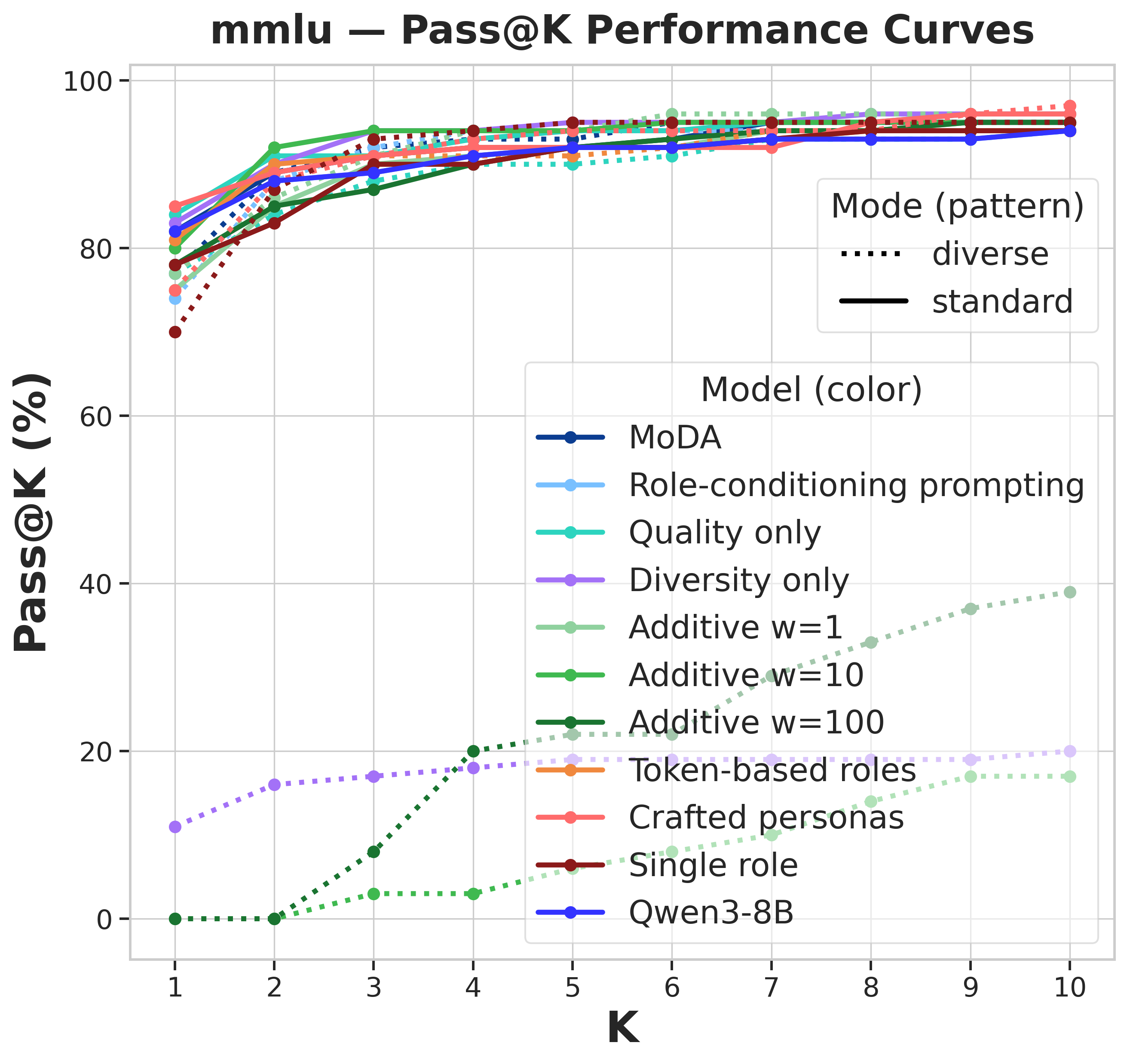}
  \small Pass@k, MMLU
  \label{fig: ablation pass@k MMLU}
\end{minipage}\hfill
\centering
\begin{minipage}{0.32\textwidth}
  \centering
  \includegraphics[width=\linewidth]{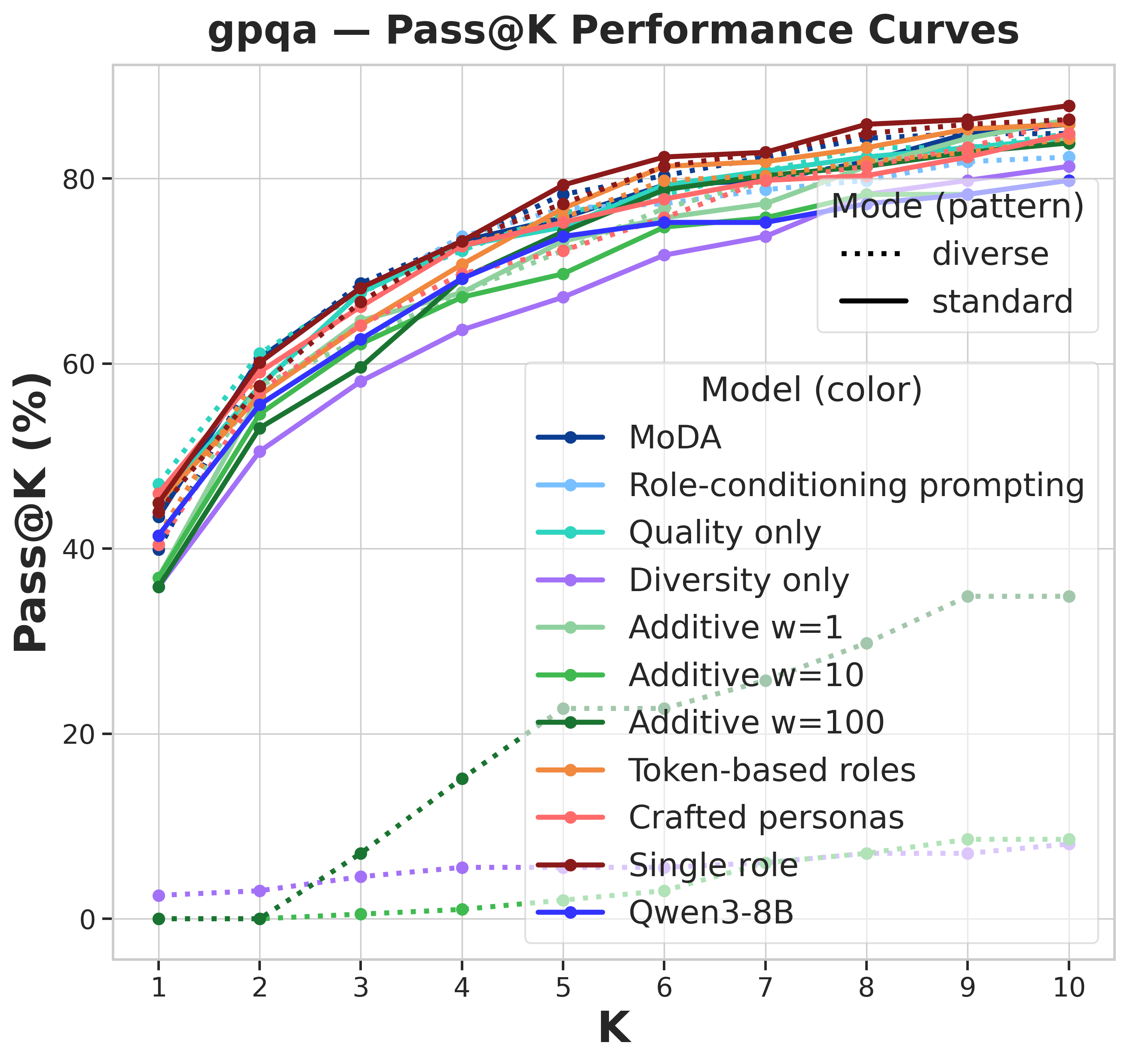}
  \small Pass@k, GPQA
  \label{fig: ablation pass@k GPQA}
\end{minipage}\hfill
\centering
\begin{minipage}{0.32\textwidth}
  \centering
  \includegraphics[width=\linewidth]{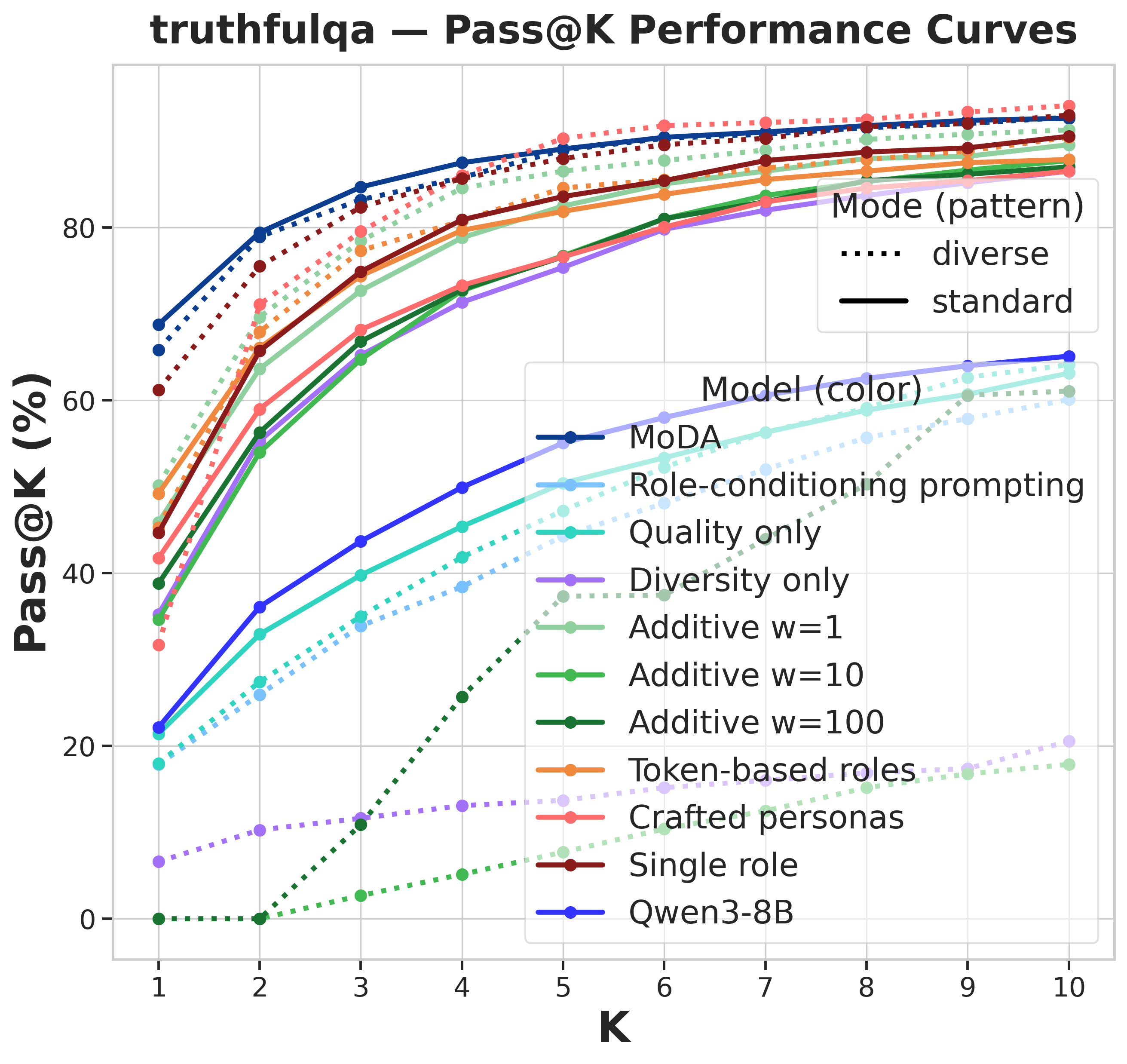}
  \small Pass@k, TruthfulQA
  \label{fig: ablation pass@k TruthfulQA}
\end{minipage}\hfill
\centering
\begin{minipage}{0.32\textwidth}
  \centering
  \includegraphics[width=\linewidth]{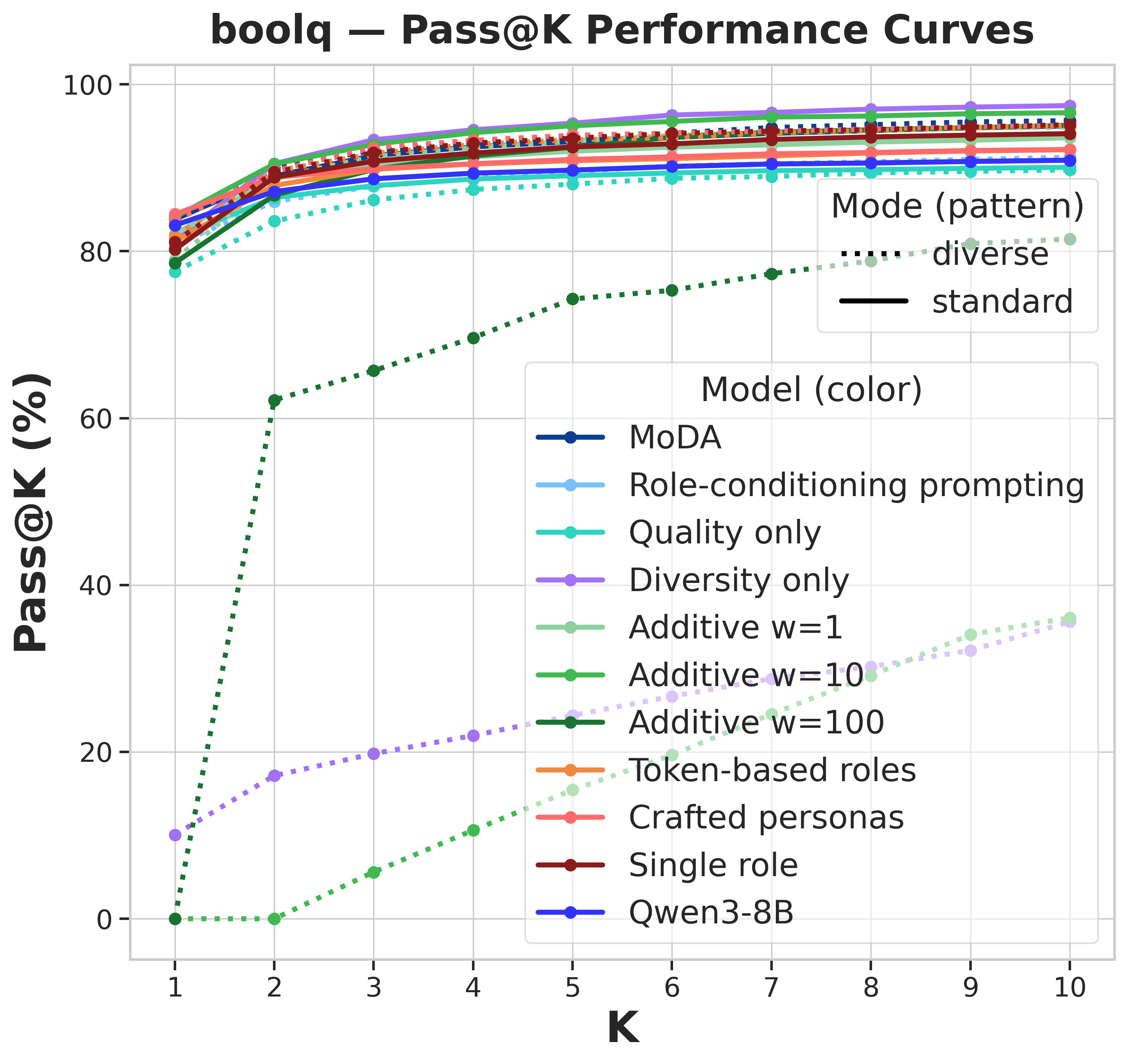}
  \small Pass@k, BoolQ
  \label{fig: ablation pass@k BoolQ}
\end{minipage}\hfill
\centering\begin{minipage}{0.32\textwidth}
  \centering
  \includegraphics[width=\linewidth]{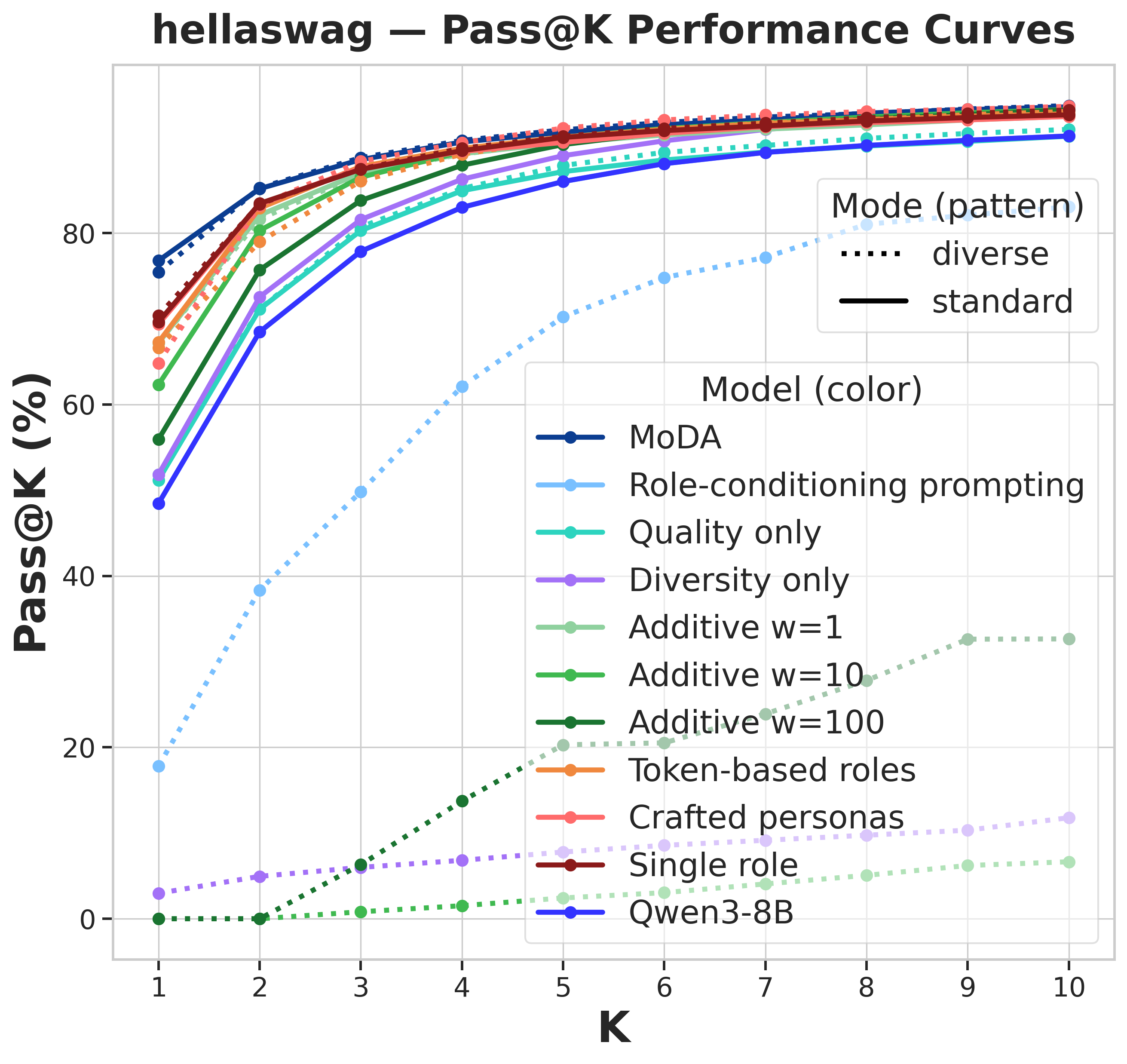}
  \small Pass@k, HellaSwag
  \label{fig: ablation pass@k HellaSwag}
\end{minipage}\hfill
\centering
\begin{minipage}{0.32\textwidth}
  \centering
  \includegraphics[width=\linewidth]{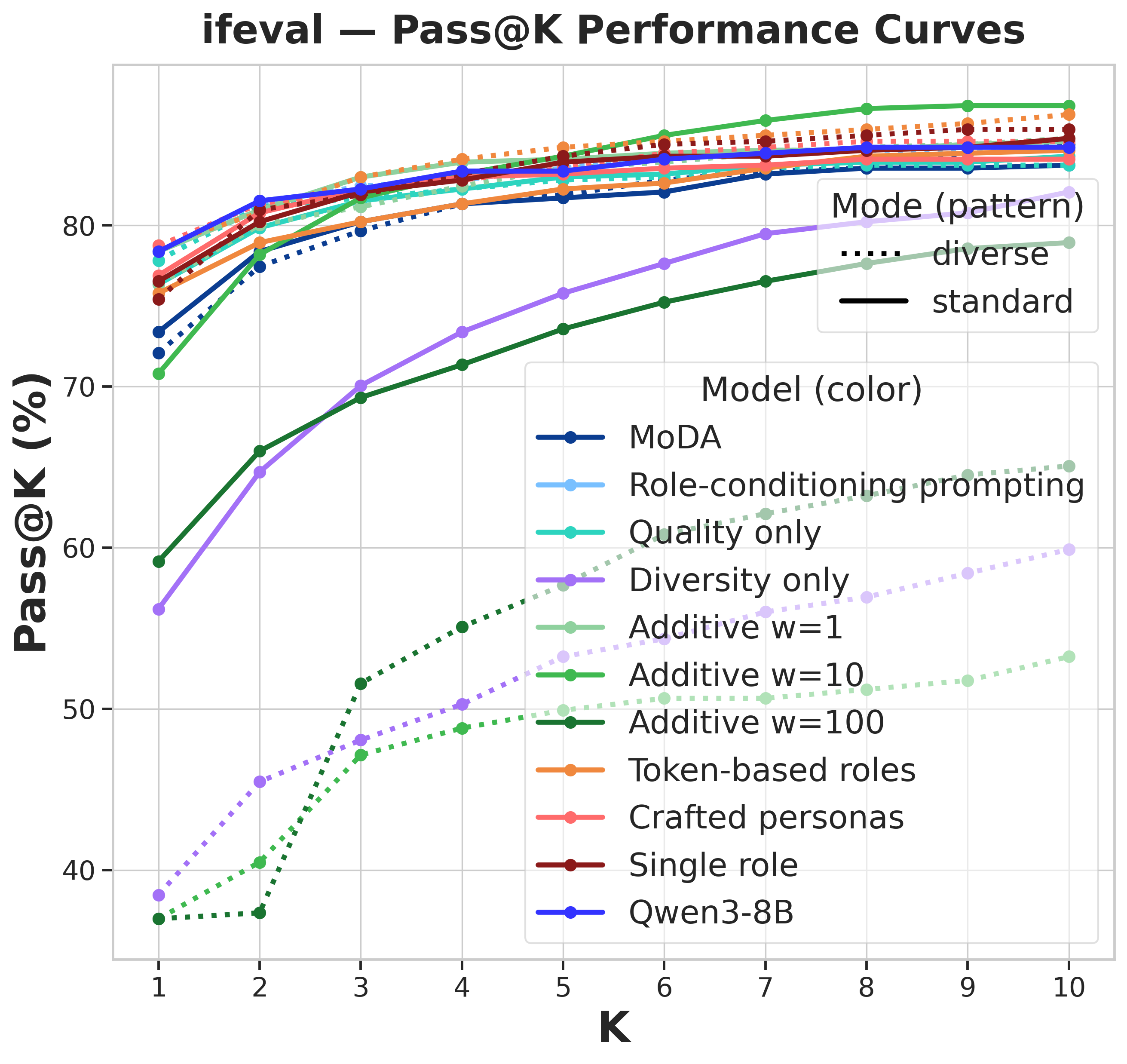}
  \small Pass@k, IF-Eval
  \label{fig: ablation pass@k IF-Eval}
\end{minipage}\hfill
\caption{\textbf{General capability retention for ablated models}: pass@k accuracy by benchmarks, standard vs diverse decoding mode.}
\label{fig:general capability pass@k ablation by benchmark}
\end{figure}

%% file: notes_arxiv/appendix/diversity_main_error_bar.tex
\section*{Infinite-Chat}

\begin{table}[H]
\centering
\begin{tabular}{lccccc}
\toprule
Model & Disc. & S & EV & Qual. \\
\midrule
\multicolumn{5}{l}{\textsc{\underline{Qwen3-8B base, thinking disabled}}} \\
Qwen3-8B & 0.400 $\pm$ 0.001 & 0.132 $\pm$ 0.001 & 1.83 $\pm$ 0.01 & 62.9\% $\pm$ 0.8\% \\
DARLING (Mix) & 0.430 $\pm$ 0.001 & 0.203 $\pm$ 0.001 & 2.33 $\pm$ 0.00 & 57.9\% $\pm$ 1.0\% \\
DARLING (WildChat) & 0.416 $\pm$ 0.001 & 0.185 $\pm$ 0.002 & 2.15 $\pm$ 0.01 & 62.7\% $\pm$ 1.0\% \\
DivPO & 0.410 $\pm$ 0.001 & 0.274 $\pm$ 0.004 & 2.86 $\pm$ 0.03 & 66.2\% $\pm$ 0.9\% \\
\method & \textbf{0.472} $\pm$ 0.001 & \textbf{0.482} $\pm$ 0.003 & \textbf{4.40} $\pm$ 0.01 & \textbf{73.2\%} $\pm$ 0.9\% \\
\midrule
\multicolumn{5}{l}{\textsc{\underline{Qwen3-8B base, thinking enabled}}} \\
Qwen3-8B & 0.400 $\pm$ 0.001 & 0.136 $\pm$ 0.001 & 1.85 $\pm$ 0.01 & 75.8\% $\pm$ 0.7\% \\
SSoT & 0.397 $\pm$ 0.000 & 0.178 $\pm$ 0.002 & 2.09 $\pm$ 0.01 & 76.3\% $\pm$ 0.7\% \\
DivPO & 0.414 $\pm$ 0.002 & 0.200 $\pm$ 0.006 & 2.20 $\pm$ 0.03 & 71.7\% $\pm$ 0.8\% \\
\method & \textbf{0.600} $\pm$ 0.001 & \textbf{0.909} $\pm$ 0.001 & \textbf{9.04} $\pm$ 0.01 & \textbf{76.8\%} $\pm$ 0.7\% \\
\midrule
\multicolumn{5}{l}{\textsc{\underline{Llama-3.1-8B base, thinking disabled}}} \\
Llama-3.1-8B & 0.414 $\pm$ 0.001 & 0.217 $\pm$ 0.002 & 2.40 $\pm$ 0.01 & 46.7\% $\pm$ 0.7\% \\
\method{} (Llama-3.1-8B) & \textbf{0.441} $\pm$ 0.002 & \textbf{0.373} $\pm$ 0.003 & \textbf{3.77} $\pm$ 0.02 & \textbf{51.9\%} $\pm$ 0.9\% \\
\midrule
\multicolumn{5}{l}{\textsc{\underline{GLM-4-9B base, thinking disabled}}} \\
GLM-4-9B & 0.421 $\pm$ 0.001 & 0.234 $\pm$ 0.002 & 2.60 $\pm$ 0.01 & 44.9\% $\pm$ 0.9\% \\
\method{}(GLM-4-9b) & \textbf{0.481} $\pm$ 0.001 & \textbf{0.528} $\pm$ 0.001 & \textbf{5.14} $\pm$ 0.01 & \textbf{60.7\%} $\pm$ 0.8\% \\
\bottomrule
\end{tabular}
\end{table}

\section*{NoveltyBench}

\begin{table}[H]
\centering
\begin{tabular}{lccccc}
\toprule
Model & Disc. & S & EV & Qual. \\
\midrule
\multicolumn{5}{l}{\textsc{\underline{Qwen3-8B base, thinking disabled}}} \\
Qwen3-8B & 0.493 $\pm$ 0.002 & 0.224 $\pm$ 0.003 & 2.36 $\pm$ 0.02 & 4.58 $\pm$ 0.04 \\
DARLING (Mix) & 0.513 $\pm$ 0.001 & 0.281 $\pm$ 0.001 & 2.96 $\pm$ 0.01 & 5.24 $\pm$ 0.07 \\
DARLING (WildChat) & 0.480 $\pm$ 0.001 & 0.335 $\pm$ 0.004 & 3.11 $\pm$ 0.02 & 4.00 $\pm$ 0.06 \\
DivPO & 0.519 $\pm$ 0.001 & 0.380 $\pm$ 0.003 & 3.58 $\pm$ 0.02 & 4.87 $\pm$ 0.02 \\
\method & \textbf{0.537} $\pm$ 0.001 & \textbf{0.470} $\pm$ 0.003 & \textbf{4.19} $\pm$ 0.01 & \textbf{5.47} $\pm$ 0.03 \\
\midrule
\multicolumn{5}{l}{\textsc{\underline{Qwen3-8B base, thinking enabled}}} \\
Qwen3-8B & 0.510 $\pm$ 0.001 & 0.258 $\pm$ 0.003 & 2.67 $\pm$ 0.03 & 4.90 $\pm$ 0.04 \\
SSoT & 0.438 $\pm$ 0.001 & 0.239 $\pm$ 0.002 & 2.57 $\pm$ 0.01 & 0.46 $\pm$ 0.01 \\
DivPO & 0.521 $\pm$ 0.001 & 0.327 $\pm$ 0.003 & 3.12 $\pm$ 0.02 & \textbf{5.06} $\pm$ 0.04 \\
\method & \textbf{0.589} $\pm$ 0.002 & \textbf{0.800} $\pm$ 0.004 & \textbf{7.63} $\pm$ 0.07 & 3.39 $\pm$ 0.05 \\
\midrule
\multicolumn{5}{l}{\textsc{\underline{Llama-3.1-8B base, thinking disabled}}} \\
Llama-3.1-8B & 0.512 $\pm$ 0.001 & 0.312 $\pm$ 0.001 & 3.04 $\pm$ 0.01 & 4.80 $\pm$ 0.08 \\
\method{} (Llama-3.1-8B) & \textbf{0.532} $\pm$ 0.001 & \textbf{0.392} $\pm$ 0.001 & \textbf{3.75} $\pm$ 0.01 & \textbf{6.17} $\pm$ 0.04 \\
\midrule
\multicolumn{5}{l}{\textsc{\underline{GLM-4-9B base, thinking disabled}}} \\
GLM-4-9B & \textbf{0.544} $\pm$ 0.002 & 0.408 $\pm$ 0.004 & 4.08 $\pm$ 0.04 & \textbf{4.14} $\pm$ 0.01 \\
\method{}(GLM-4-9b) & 0.502 $\pm$ 0.002 & \textbf{0.521} $\pm$ 0.005 & \textbf{5.01} $\pm$ 0.04 & 2.70 $\pm$ 0.03 \\
\bottomrule
\end{tabular}
\end{table}

\section*{HypoBench}

\begin{table}[H]
\centering
\begin{tabular}{lccccc}
\toprule
Model & Disc. & S & EV & Qual. \\
\midrule
\multicolumn{5}{l}{\textsc{\underline{Qwen3-8B base, thinking disabled}}} \\
Qwen3-8B & 0.384 $\pm$ 0.005 & 0.144 $\pm$ 0.005 & 1.91 $\pm$ 0.03 & \textbf{4.03} $\pm$ 0.01 \\
DARLING (Mix) & 0.406 $\pm$ 0.005 & 0.222 $\pm$ 0.004 & 2.48 $\pm$ 0.02 & 3.44 $\pm$ 0.03 \\
DARLING (WildChat) & 0.415 $\pm$ 0.004 & 0.205 $\pm$ 0.006 & 2.34 $\pm$ 0.04 & 3.73 $\pm$ 0.02 \\
DivPO & 0.427 $\pm$ 0.006 & 0.198 $\pm$ 0.005 & 2.30 $\pm$ 0.03 & 3.89 $\pm$ 0.01 \\
\method & \textbf{0.480} $\pm$ 0.003 & \textbf{0.547} $\pm$ 0.010 & \textbf{4.91} $\pm$ 0.05 & 3.80 $\pm$ 0.02 \\
\midrule
\multicolumn{5}{l}{\textsc{\underline{Qwen3-8B base, thinking enabled}}} \\
Qwen3-8B & 0.399 $\pm$ 0.006 & 0.164 $\pm$ 0.005 & 2.05 $\pm$ 0.03 & \textbf{3.93} $\pm$ 0.01 \\
SSoT & 0.419 $\pm$ 0.002 & 0.220 $\pm$ 0.010 & 2.33 $\pm$ 0.04 & 3.75 $\pm$ 0.02 \\
DivPO & 0.409 $\pm$ 0.002 & 0.162 $\pm$ 0.005 & 2.04 $\pm$ 0.03 & 3.89 $\pm$ 0.01 \\
\method & \textbf{0.579} $\pm$ 0.013 & \textbf{0.875} $\pm$ 0.018 & \textbf{8.06} $\pm$ 0.25 & 3.71 $\pm$ 0.04 \\
\midrule
\multicolumn{5}{l}{\textsc{\underline{Llama-3.1-8B base, thinking disabled}}} \\
Llama-3.1-8B & \textbf{0.469} $\pm$ 0.005 & 0.207 $\pm$ 0.006 & 2.38 $\pm$ 0.04 & 3.60 $\pm$ 0.02 \\
\method{} (Llama-3.1-8B) & 0.410 $\pm$ 0.001 & \textbf{0.324} $\pm$ 0.014 & \textbf{3.27} $\pm$ 0.10 & \textbf{3.65} $\pm$ 0.01 \\
\midrule
\multicolumn{5}{l}{\textsc{\underline{GLM-4-9B base, thinking disabled}}} \\
GLM-4-9B & 0.402 $\pm$ 0.003 & 0.225 $\pm$ 0.005 & 2.52 $\pm$ 0.03 & \textbf{3.75} $\pm$ 0.01 \\
\method{}(GLM-4-9b) & \textbf{0.452} $\pm$ 0.003 & \textbf{0.485} $\pm$ 0.012 & \textbf{4.64} $\pm$ 0.09 & 3.68 $\pm$ 0.01 \\
\bottomrule
\end{tabular}
\end{table}

\section*{PreScience}

\begin{table}[H]
\centering
\begin{tabular}{lccccc}
\toprule
Model & Disc. & S & EV & Qual. \\
\midrule
\multicolumn{5}{l}{\textsc{\underline{Qwen3-8B base, thinking disabled}}} \\
Qwen3-8B & 0.431 $\pm$ 0.002 & 0.155 $\pm$ 0.004 & 1.94 $\pm$ 0.02 & \textbf{4.25} $\pm$ 0.03 \\
DARLING (Mix) & \textbf{0.559} $\pm$ 0.005 & \textbf{0.541} $\pm$ 0.018 & \textbf{4.73} $\pm$ 0.15 & 2.33 $\pm$ 0.07 \\
DARLING (WildChat) & 0.446 $\pm$ 0.002 & 0.192 $\pm$ 0.003 & 2.21 $\pm$ 0.03 & 4.03 $\pm$ 0.06 \\
DivPO & 0.455 $\pm$ 0.002 & 0.233 $\pm$ 0.001 & 2.41 $\pm$ 0.01 & 3.96 $\pm$ 0.05 \\
\method & 0.440 $\pm$ 0.002 & 0.176 $\pm$ 0.002 & 2.10 $\pm$ 0.01 & 3.96 $\pm$ 0.01 \\
\midrule
\multicolumn{5}{l}{\textsc{\underline{Qwen3-8B base, thinking enabled}}} \\
Qwen3-8B & 0.441 $\pm$ 0.003 & 0.136 $\pm$ 0.001 & 1.83 $\pm$ 0.01 & 3.79 $\pm$ 0.04 \\
SSoT & \textbf{0.461} $\pm$ 0.002 & \textbf{0.281} $\pm$ 0.017 & \textbf{2.39} $\pm$ 0.05 & 3.30 $\pm$ 0.05 \\
DivPO & 0.448 $\pm$ 0.003 & 0.185 $\pm$ 0.015 & 2.04 $\pm$ 0.07 & \textbf{3.87} $\pm$ 0.06 \\
\method & 0.439 $\pm$ 0.001 & 0.194 $\pm$ 0.005 & 2.10 $\pm$ 0.01 & 3.68 $\pm$ 0.00 \\
\midrule
\multicolumn{5}{l}{\textsc{\underline{Llama-3.1-8B base, thinking disabled}}} \\
Llama-3.1-8B & 0.461 $\pm$ 0.005 & \textbf{0.514} $\pm$ 0.020 & 3.67 $\pm$ 0.07 & \textbf{2.62} $\pm$ 0.05 \\
\method{} (Llama-3.1-8B) & \textbf{0.472} $\pm$ 0.003 & 0.470 $\pm$ 0.009 & \textbf{3.89} $\pm$ 0.07 & 2.54 $\pm$ 0.04 \\
\midrule
\multicolumn{5}{l}{\textsc{\underline{GLM-4-9B base, thinking disabled}}} \\
GLM-4-9B & 0.461 $\pm$ 0.004 & 0.438 $\pm$ 0.010 & 4.15 $\pm$ 0.08 & 2.57 $\pm$ 0.05 \\
\method{}(GLM-4-9b) & \textbf{0.464} $\pm$ 0.002 & \textbf{0.448} $\pm$ 0.010 & \textbf{4.28} $\pm$ 0.09 & \textbf{2.58} $\pm$ 0.10 \\
\bottomrule
\end{tabular}
\end{table}

%% file: notes_arxiv/appendix/general_capability_error_bar.tex
\begin{table}[H]
\setlength{\tabcolsep}{2.5pt}
\centering
\small
\begin{tabular}{lcccccccc}
\toprule
Model & GSM8K & MMLU & GPQA & BoolQ & HellaSwag & TruthfulQA & IFEval & Avg. \\
\midrule
\multicolumn{9}{l}{\textsc{\underline{Qwen3-8B base, thinking disabled}}} \\
Qwen3-8B & 84.5 $\pm$ 1.0 & \textbf{82.0} $\pm$ 3.9 & 41.4 $\pm$ 3.5 & 83.1 $\pm$ 0.7 & 48.4 $\pm$ 0.5 & 22.2 $\pm$ 1.5 & \textbf{78.4} $\pm$ 1.8 & 62.9 $\pm$ 0.8 \\
DARLING (Mixture) & 63.5 $\pm$ 1.3 & 63.0 $\pm$ 4.9 & 34.3 $\pm$ 3.4 & 76.7 $\pm$ 0.7 & 62.2 $\pm$ 0.5 & 52.4 $\pm$ 1.7 & 53.0 $\pm$ 2.1 & 57.9 $\pm$ 1.0 \\
DARLING (WildChat) & 77.9 $\pm$ 1.1 & 55.0 $\pm$ 5.0 & 34.3 $\pm$ 3.4 & \textbf{86.4} $\pm$ 0.6 & 68.8 $\pm$ 0.5 & 43.2 $\pm$ 1.7 & 73.0 $\pm$ 1.9 & 62.7 $\pm$ 1.0 \\
DivPO & 80.5 $\pm$ 1.1 & 81.0 $\pm$ 3.9 & 40.4 $\pm$ 3.5 & 83.5 $\pm$ 0.6 & 62.8 $\pm$ 0.5 & 37.7 $\pm$ 1.7 & 77.3 $\pm$ 1.8 & 66.2 $\pm$ 0.9 \\
\method & \textbf{86.7} $\pm$ 0.9 & 81.0 $\pm$ 3.9 & \textbf{52.0} $\pm$ 3.6 & 84.8 $\pm$ 0.6 & \textbf{73.4} $\pm$ 0.4 & \textbf{57.9} $\pm$ 1.7 & 76.9 $\pm$ 1.8 & \textbf{73.2} $\pm$ 0.9 \\
\midrule
\multicolumn{9}{l}{\textsc{\underline{Qwen3-8B base, thinking enabled}}} \\
Qwen3-8B & \textbf{90.3} $\pm$ 0.8 & \textbf{94.0} $\pm$ 2.4 & 43.4 $\pm$ 3.5 & \textbf{87.2} $\pm$ 0.6 & 70.9 $\pm$ 0.5 & 64.9 $\pm$ 1.7 & \textbf{79.7} $\pm$ 1.7 & 75.8 $\pm$ 0.7 \\
SSoT & 84.3 $\pm$ 1.0 & 93.0 $\pm$ 2.6 & \textbf{55.1} $\pm$ 3.5 & \textbf{87.2} $\pm$ 0.6 & \textbf{80.1} $\pm$ 0.4 & \textbf{72.8} $\pm$ 1.6 & 61.9 $\pm$ 2.1 & 76.3 $\pm$ 0.7 \\
DivPO & 84.9 $\pm$ 1.0 & 91.0 $\pm$ 2.9 & 44.4 $\pm$ 3.5 & 83.2 $\pm$ 0.7 & 65.0 $\pm$ 0.5 & 55.3 $\pm$ 1.7 & 77.6 $\pm$ 1.8 & 71.7 $\pm$ 0.8 \\
\method & 83.9 $\pm$ 1.0 & 93.0 $\pm$ 2.6 & 50.5 $\pm$ 3.6 & \textbf{87.2} $\pm$ 0.6 & 76.2 $\pm$ 0.4 & 67.2 $\pm$ 1.6 & 79.5 $\pm$ 1.7 & \textbf{76.8} $\pm$ 0.7 \\
\midrule
\multicolumn{9}{l}{\textsc{\underline{Llama-3.1-8B base, thinking disabled}}} \\
Llama-3.1-8B & 71.3 $\pm$ 1.2 & 16.0 $\pm$ 3.7 & 6.1 $\pm$ 1.7 & \textbf{81.6} $\pm$ 0.7 & 30.2 $\pm$ 0.5 & 53.4 $\pm$ 1.7 & 68.8 $\pm$ 2.0 & 46.7 $\pm$ 0.7 \\
\method(Llama-3.1-8B) & \textbf{74.9} $\pm$ 1.2 & \textbf{30.0} $\pm$ 4.6 & \textbf{18.7} $\pm$ 2.8 & 75.6 $\pm$ 0.8 & \textbf{38.5} $\pm$ 0.5 & \textbf{54.7} $\pm$ 1.7 & \textbf{70.8} $\pm$ 2.0 & \textbf{51.9} $\pm$ 0.9 \\
\midrule
\multicolumn{9}{l}{\textsc{\underline{GLM-4-9B base, thinking disabled}}} \\
GLM-4-9B & 21.8 $\pm$ 1.1 & 35.0 $\pm$ 4.8 & 19.7 $\pm$ 2.8 & 74.5 $\pm$ 0.8 & \textbf{68.0} $\pm$ 0.5 & \textbf{44.8} $\pm$ 1.7 & 50.3 $\pm$ 2.2 & 44.9 $\pm$ 0.9 \\
\method(GLM-4-9B) & \textbf{84.9} $\pm$ 1.0 & \textbf{83.0} $\pm$ 3.8 & \textbf{33.3} $\pm$ 3.4 & \textbf{81.2} $\pm$ 0.7 & 52.3 $\pm$ 0.5 & 34.5 $\pm$ 1.7 & \textbf{55.5} $\pm$ 2.1 & \textbf{60.7} $\pm$ 0.8 \\
\bottomrule
\end{tabular}
\vspace{4pt}
\caption{General capability retention (accuracy \%, mean ± std error; bold = best per block)}
\label{tab:general-capability-pass1-error-bar}
\end{table}

%% file: notes_arxiv/table/ablation_table.tex
\begin{table}[H]
\centering
\setlength{\tabcolsep}{3pt}
\resizebox{\textwidth}{!}{%
\begin{tabular}{lccccc}
\toprule
& \multicolumn{2}{c}{General capability (Avg., \%)} & \multicolumn{3}{c}{Infinite-Chat Diversity} \\
\cmidrule(lr){2-3} \cmidrule(lr){4-6}
Ablation & Standard mode & Diverse mode & Disc. & SBERT & E-Vendi \\
\midrule
Role-conditioned prompting & 62.9  & 56.6  & 0.403  & 0.145  & 1.90  \\
Quality only & 63.2  & 61.8  & 0.401  & 0.131  & 1.82  \\
Diversity only & 61.1  & 10.7  & \textbf{0.656}  & \textbf{0.993}  & \textbf{9.77}  \\
Additive $w{=}1$ & 67.4  & 68.0  & 0.422  & 0.204  & 2.33  \\
Additive $w{=}10$ & 64.5  & 5.7  & 0.632  & 0.976  & 9.64  \\
Additive $w{=}100$ & 61.6  & 5.4  & 0.621  & 0.970  & 8.75  \\
Token-based roles & 69.7  & 67.9 & 0.419 & 0.197 & 2.04 \\
Crafted personas & 70.1  & 66.1  & 0.421  & 0.194  & 2.24  \\
Single role & 69.0  & 69.9  & 0.428  & 0.239  & 2.57  \\
\midrule
Qwen3-8B & 62.9 & - & 0.400 & 0.132 & 1.83 \\
\midrule
\textbf{\method{}} & \textbf{73.2}  & \textbf{71.6}  & 0.444  & 0.257  & 2.69  \\
\bottomrule
\end{tabular}
}
\vspace{4pt}
\caption{\textbf{Ablation results} for general capability (Pass@1) under standard and diverse decoding (accuracy \%) and Infinite-Chat diversity ($n{=}3$ random seeds). Bold marks the best value per column.}
\label{tab:ablation}
\vspace{-0.6cm}
\end{table}

%% file: notes_arxiv/appendix/benchmarks.tex
\section{Benchmark Descriptions}
\label{appx:benchmark}

\subsection{Diversity Metrics}
\label{app:diversity-metrics}

For each prompt \(x\), we generate a response set
\(Y_x=\{y_1,\ldots,y_K\}\) using the decoding configuration in
Appendix~\ref{appx:implementation}. Unless otherwise stated, all diversity
metrics are computed at the prompt level and then averaged over prompts.
For thinking-enabled models, metrics are computed only on the final-answer
span after removing the reasoning trace.

We report both surface-form and semantic diversity metrics.

For application-domain experiments, we additionally report semantic and
learned pairwise diversity, including SBERT diversity, E-Vendi score and learned discriminator diversity score. \textbf{SBERT diversity}\citep{reimers2019sentencebertsentenceembeddingsusing} is the mean pairwise
cosine distance between sentence embeddings:
\[
D_{\mathrm{SBERT}}(Y_x)
=
\frac{2}{K(K-1)}
\sum_{i<j}
\left(1-\cos(e_i,e_j)\right),
\]
where \(e_i\) is the normalized sentence embedding of \(y_i\). We report
the exact embedding model in Appendix~\ref{appx:implementation}.

\textbf{E-Vendi score} stands for embedding Vendi score. The Vendi score is a diversity metric inspired by ecology and quantum statistical mechanics. Given a response set \(Y_x=\{y_1,\ldots,y_K\}\), we
first compute normalized sentence embeddings \(e_i\) using SBERT for each response and form
the pairwise similarity matrix and then normalize the matrix by its trace,
\[
\tilde{S} = \frac{S}{\mathrm{tr}(S)},
\]
and let \(\lambda_1,\ldots,\lambda_K\) denote the eigenvalues of \(\tilde{S}\).
The E-Vendi score is then defined as
\[
D_{\mathrm{E\text{-}Vendi}}(Y_x)
=
\exp\left(
-\sum_{i=1}^{K} \lambda_i \log \lambda_i
\right).
\]
Intuitively, the score measures the effective number of distinct semantic
responses in the set: it is close to \(1\) when all responses are nearly
identical, and increases as the responses become more semantically diverse. In
practice, we add a small \(\epsilon\) inside the logarithm for numerical
stability.

\textbf{Discriminator diversity} uses a trained discriminator
\(f_\psi(x,y_i,y_j)\) that scores whether two responses are meaningfully
distinct for the same prompt. We compute
\[
D_{\mathrm{disc}}(Y_x)
=
\frac{2}{K(K-1)}
\sum_{i<j}
f_\psi(x,y_i,y_j).
\]
The discriminator architecture, training data, held-out split, and scoring
calibration are described in Appendix~\ref{appx:discriminator}. Notably, the
discriminator is not trained on the evaluation outputs.

\subsection{General Capability Retention}
\label{appx:general capability}

We evaluate general capability retention to test whether diversity-oriented training preserves standard reasoning, knowledge, truthfulness, and instruction-following behavior. Capability retention is evaluated in standard mode (no role injection). Max token length is 2048 when thinking is disabled, 8192 when thinking is enabled. We sample 10 responses for each prompt.

The evaluation covers seven benchmarks: \textbf{GSM8K} \citep{cobbe2021gsm8k} for grade-school mathematical reasoning, \textbf{MMLU} \citep{hendryckstest2021} for broad multitask knowledge, \textbf{GPQA} \citep{rein2024gpqa} for graduate-level scientific reasoning, \textbf{BoolQ} \citep{clark2019boolq} for yes/no reading comprehension, \textbf{HellaSwag} \citep{zellers2019hellaswag} for commonsense sentence completion, \textbf{TruthfulQA-MC1} \citep{lin2022truthfulqameasuringmodelsmimic} for robustness to imitative
falsehoods, and \textbf{IFEval} \citep{zhou2023instructionfollowingevaluationlargelanguage} for instruction-following under explicit formatting constraints.

We use each benchmark's native scoring rule: extracted-answer accuracy for GSM8K, multiple-choice accuracy for BoolQ, MMLU, GPQA, HellaSwag, and TruthfulQA-MC1, and strict prompt-level accuracy for IFEval.

\subsection{Domain Application Tasks}
\label{app:domain application}

The domain application suite evaluates whether diversity gains transfer to open-ended generation settings (scientific ideation and creative writing) where multiple distinct high-quality outputs are critical. The suite consists of four tasks. We sample 10 responses per model. For models with role conditioning, we rotate through the available roles to generate these samples.

\textbf{HypoBench} \citep{liu2026hypobenchsystematicprincipledbenchmarking} evaluates scientific hypothesis generation. Given a set of labeled data-science observations, the model is asked to generate hypotheses that explain possible underlying patterns. We evaluate on both real-world datasets, including \texttt{deceptive\_reviews}, \texttt{dreaddit}, and \texttt{headline\_binary}, and synthetic datasets, including \texttt{admission\//level\_1\//base}, \texttt{election\//level1}, and \texttt{shoe}. Every parsed hypothesis is scored individually by an LLM judge (gpt-4o-mini by default) on 3 dimensions, 1–5 each:
\begin{itemize}
    \item Clarity — how precisely/testably it's stated
    \item  Novelty — how non-obvious it is, compared against any known/published hypotheses
       supplied for that dataset
     \item Plausibility — how scientifically well-reasoned it is given the task. We report the average of the 3 scores across every hypothesis scored as the quality metric. Max token number is set to 4096 when thinking is disabled, 16384 when thinking is enabled.
\end{itemize}

\textbf{PreScience} \citep{Ajith2026} evaluates scientific follow-up prediction using the Contribution Generation task. Given prior research context, the model generates a plausible future title and abstract. We use the benchmark's native LACER (Lattice of Automatically Constructed Exemplars for Reference)-style scoring pipeline. Each generated (title, abstract) is paired with the actual/ground-truth follow-up paper as the reference. Both are fed into the judge model (gpt-4o-2024-11-20 by default) with a fixed few-shot prompt template, and the judge model is asked to score the reference-generation pair on the scale of 1–10 based on how similar is the generated paper to the real one. We evaluate each model on 10 research contexts, and report the average LACER score on all the generations as the quality metric. Max token number is set to 1500 when thinking is disabled, 6000 when thinking is enabled.

\textbf{NoveltyBench} \citep{zhang2025noveltybenchevaluatinglanguagemodels} evaluates creative and novelty-seeking generation prompts. It presents each model with a single open-ended creative/generative prompt (e.g. "Tell me a story in five sentences about a girl and her dog") and samples multiple generations. Every generation is scored 1–10 by \texttt{Skywork/Skywork-Reward-Gemma-2-27B-v0.2}. For each prompt, utility is the sum of generation scores, weighted by the original sampling order. This rewards a model for surfacing several distinct good ideas early, and discounts the marginal value of yet another (possibly redundant) generation later in the batch. We use the benchmark's official quality or utility scoring pipeline where available and report diversity over the generated response sets. We report the average utility across all prompts as the quality metric. The Max token number is set to 1024 when thinking is disabled, 4096 when thinking is enabled.

\textbf{Infinite-Chat} \citep{jiang2025artificialhivemindopenendedhomogeneity} comprises 100 open-ended prompts randomly sampled from the held-out set of Infinite-Chat dataset. Since there's no native quality metric for Infinite-Chat, we use the average pass@1 accuracy of the general capability suite as the quality metric. Max token number is set to 1024 when thinking is disabled, 4096 when thinking is enabled.

For each domain, we report native task quality when available and diversity metrics over the generated response sets. Raw diversity tables first average each metric over prompts within a domain and then average the four domain means.

%% file: notes_arxiv/appendix/generation_examples.tex
\section{Generation Examples}
\label{appx: generation examples}

\input{notes_arxiv/appendix/generation_chess}
\input{notes_arxiv/appendix/generation_5_word}
\input{notes_arxiv/appendix/generation_cocktail}

%% file: notes_arxiv/appendix/generation_chess.tex
\label{appx:generation-example-chess}

\begin{figure}[t]
\centering

\begin{promptbox}
\textit{Give a random chess move in response to 1. b3.}
\end{promptbox}

\begin{minipage}[t]{0.32\linewidth}
\begin{respbox}{Sample 1}
Sure! A random but plausible chess move in response to \textbf{1. b3} could be:

\textbf{1... d5}

This is a natural move for Black, developing the queen's knight to d5, putting pressure on the center and responding to the unusual opening move 1. b3. It's a solid and common choice in many openings, such as the King's Indian Defense or the Queen's Gambit Declined.
\end{respbox}
\end{minipage}\hfill%
\begin{minipage}[t]{0.32\linewidth}
\begin{respbox}{Sample 2}
Sure! In response to the opening 1.\textbf{b3}, there are a number of possible moves for Black, and we can pick a random one. Let's go with:

\textbf{1... d5}

This is a common and solid move for Black, aiming to challenge White's pawn on b3 and develop the queen's knight. It's a natural response and leads to a variety of strategic games.
\end{respbox}
\end{minipage}\hfill%
\begin{minipage}[t]{0.32\linewidth}
\begin{respbox}{Sample 3}
Sure! The move \textbf{1.b3} is a rare opening move, not commonly used at the top level, but it's a valid and interesting choice that leads to a quiet, positional game. It prepares to develop the bishop and can be part of a Catalan or a King's Indian Attack style, depending on how the game develops.

A random, reasonable response to \textbf{1.b3} could be:

\textbf{1...e5}

This is a solid and natural reply, aiming to control the center and prepare to develop the light-squared bishop. It's flexible and gives Black good options for future play.

So, the random chess move in response to \textbf{1.b3} is: \textbf{1...e5}.
\end{respbox}
\end{minipage}

\vspace{6pt}

\begin{minipage}[t]{0.32\linewidth}
\begin{respbox}{Sample 4}
Sure! Here's a random chess move in response to \textbf{1. b3}:

\textbf{1... d5}

This is a common and solid move that challenges White's central control and prepares to develop pieces. It's a natural response for Black and leads to a variety of positions, including the Queen's Gambit Declined and other strategic lines. Let me know if you'd like to explore a specific variation!
\end{respbox}
\end{minipage}\hfill%
\begin{minipage}[t]{0.32\linewidth}
\begin{respbox}{Sample 5}
Sure! Here's a random chess move in response to \textbf{1. b3}:

\textbf{1... d5}

This move is solid and common, developing the knight and preparing to challenge the center. It's a natural continuation in many openings, such as the Queen's Gambit Declined or the Modern Defense. Let me know if you'd like more options!
\end{respbox}
\end{minipage}\hfill%
\begin{minipage}[t]{0.32\linewidth}
\begin{respbox}{Sample 6}
Sure! Here's a random chess move in response to \texttt{1. b3}:

\textbf{1... d5}

This is a solid and aggressive reply that aims to quickly develop the queen's knight and take control of the center. It leads to sharp and complex positions that can suit a variety of playing styles.
\end{respbox}
\end{minipage}

\vspace{6pt}

\begin{minipage}[t]{0.32\linewidth}
\begin{modarespbox}{Role 1}
\textbf{Domain:} Game Theory / Chess Strategy \quad \textbf{Focus:} Opening Play / Positional Awareness

\textbf{Context:} The move \texttt{1.b3} is a classical opening in chess, often considered part of the English Opening or a preparatory move toward the Modern Benoni or other structures. It aims to fianchetto the light-squared bishop, control the center, and prepare for development......

\textbf{Conclusion:} The move \texttt{1...e5} is a reasonable and strategically sound response to \texttt{1.b3}, offering Black a solid foundation for counterplay while maintaining flexibility in development. It aligns with the principle of developing pieces quickly and controlling the center, which are fundamental to sound chess strategy.

\end{modarespbox}
\end{minipage}\hfill%
\begin{minipage}[t]{0.32\linewidth}
\begin{modarespbox}{Role 2}
\textbf{Move:} \textbf{d5}

\textbf{Reasoning:} Central Control: playing d5 is a classic response to 1.b3, aiming to challenge White's central pawn structure and prepare for development. Dynamic Counterplay: it opens lines for Black's pieces (e.g., knight to c6 or c5) and invites tactical exchanges. Flexibility: d5 supports future pawn breaks like e5 or c5, depending on the game's trajectory.

\textbf{Example Continuation:} 1.b3 d5 2.Bb2 Nf6 3.e3 c6 4.c4 dxc4 5.d4 c5 (a common line in the Sicilian Defense variant).

This move is purposeful yet aligns with ``random'' as it's a natural, well-known response. Let me know if you'd like more options!

\end{modarespbox}
\end{minipage}\hfill%
\begin{minipage}[t]{0.32\linewidth}
\begin{modarespbox}{Role 3}
\textbf{1. Understanding the Position:} White moves first: \texttt{1. b3}. Evaluation: b3 is a flexible opening that aims for a solid but unambitious game. It prepares to develop the light-squared bishop, controls the center, and can lead to a fianchetto setup. Counterplay: Black can choose between several sound responses, each with different strategic goals (solid, attacking, dynamic, etc.).

\textbf{2. Potential Responses (as of 2024):} \texttt{1... d5}, a classic Black pawn break aiming for control in the center; \texttt{1... e5}, daring, attacking, and takes the initiative immediately; \texttt{1... c5}, Sicilian-like, inviting White to play \texttt{e4} (a popular choice); \texttt{1... Nf6}, develops a piece and prepares castling; \texttt{1... g6}, semi-Sicilian setup, often leading to a fianchetto.

\textbf{3. Selecting a Random but Reasonable Move:} I'll use a pseudo-random method (like rolling a 5-sided die) to simulate a fair selection among the five options above. Result: option \textbf{c. 1... c5}.

\textbf{4. Final Answer:} \textbf{1. b3 c5}

Let me know if you'd like to see a full game continuation or analysis!
\end{modarespbox}
\end{minipage}

\vspace{6pt}

\begin{minipage}[t]{0.32\linewidth}
\begin{modarespbox}{Role 4}
\textbf{Response to 1.b3:} \textbf{2. d4}

\textbf{Rationale:} The move \textbf{2.d4} is a natural and solid choice after \textbf{1.b3}, aiming to control the center and develop pieces. It keeps the game dynamic and opens up options for both sides. It's a common move in openings like the Queen's Gambit Declined or the Catalan, depending on how Black responds.

\textbf{Example continuation:} If Black plays \textbf{2...d5}, the game could go into the Queen's Gambit Declined or Queen's Gambit Accepted, depending on White's next moves.

\textbf{Alternative moves for Black after 1.b3 could include:} \textbf{2...e5} (leading to the Petrov Defense); \textbf{2...d5} (Queen's Gambit Declined or Accepted); \textbf{2...Nf6} (Modern Defense); \textbf{2...g6} (King's Indian Defense).

Let me know if you'd like to explore a specific line or see an entire game from this position!

\end{modarespbox}
\end{minipage}\hfill%
\begin{minipage}[t]{0.32\linewidth}
\begin{modarespbox}{Role 5}
Certainly! Here's a random valid move from White following \textbf{1. b3}:

\textbf{1. ... d5}

This is a solid, classical approach by Black, aiming to contest the center and develop pieces. It's a common choice in many openings, such as the King's Indian Defense or Petrov's Defense, depending on how White responds.

Let me know if you'd like to continue the game or see other possible moves!
\end{modarespbox}
\end{minipage}\hfill%
\begin{minipage}[t]{0.32\linewidth}
\begin{modarespbox}{Role 6}
\textbf{1. Understanding the Position:} White moves first: \texttt{1. b3}. Evaluation: b3 is a flexible opening that aims for a solid but unambitious game. It prepares to develop the light-squared bishop, controls the center, and can lead to a fianchetto setup. Counterplay: Black can choose between several sound responses, each with different strategic goals (solid, attacking, dynamic, etc.).

\textbf{2. Potential Responses (as of 2024):} \texttt{1... d5}, a classic Black pawn break aiming for control in the center; \texttt{1... e5}, daring, attacking, and takes the initiative immediately; \texttt{1... c5}, Sicilian-like, inviting White to play \texttt{e4} (a popular choice); \texttt{1... Nf6}, develops a piece and prepares castling; \texttt{1... g6}, semi-Sicilian setup, often leading to a fianchetto.

\textbf{3. Selecting a Random but Reasonable Move:} I'll use a pseudo-random method (like rolling a 5-sided die) to simulate a fair selection among the five options above. Result: option \textbf{c. 1... c5}.

\textbf{4. Final Answer:} \textbf{1. b3 c5}
\end{modarespbox}
\end{minipage}

\vspace{6pt}
\caption{\textbf{Qualitative example of diverse generation: \textit{Give a random chess move in response to 1. b3.}} Upper: Responses sampled from Qwen3-8B. Bottom: Responses sampled from \method{} under distinct numbered roles for the same prompt, illustrating conceptual diversity across generations.}

\label{fig:qualitative-example-chess}
\vspace{-0.4cm}
\end{figure}

%% file: notes_arxiv/appendix/generation_5_word.tex
\label{appx:generation-example-passphrase}

\begin{figure}[t]
\centering

\begin{promptbox}
\textit{Generate a 5 word passphrase separated by hyphens.}
\end{promptbox}

\begin{minipage}[t]{0.32\linewidth}
\begin{respbox}{Sample 1}
Moonlight-Stars-Dreams-Wind-Sky
\end{respbox}
\end{minipage}\hfill%
\begin{minipage}[t]{0.32\linewidth}
\begin{respbox}{Sample 2}
Moonlight-Peak-Horse-Forest-Snow
\end{respbox}
\end{minipage}\hfill%
\begin{minipage}[t]{0.32\linewidth}
\begin{respbox}{Sample 3}
Moonlight-Peak-Wind-Whisper-Sky
\end{respbox}
\end{minipage}

\vspace{6pt}

\begin{minipage}[t]{0.32\linewidth}
\begin{respbox}{Sample 4}
moonlight - whisper - shadow - echo - breeze
\end{respbox}
\end{minipage}\hfill%
\begin{minipage}[t]{0.32\linewidth}
\begin{respbox}{Sample 5}
Moonlight - Whisper - Dreams - Dance - Sky
\end{respbox}
\end{minipage}\hfill%
\begin{minipage}[t]{0.32\linewidth}
\begin{respbox}{Sample 6}
Moonlight - Whisper - Journey - Echo - Dreams
\end{respbox}
\end{minipage}

\vspace{6pt}

\begin{minipage}[t]{0.32\linewidth}
\begin{modarespbox}{Role 1}
decide-quickly-act-fearless-now
\end{modarespbox}
\end{minipage}\hfill%
\begin{minipage}[t]{0.32\linewidth}
\begin{modarespbox}{Role 2}
parrot-salt-hike-echo-moon
\end{modarespbox}
\end{minipage}\hfill%
\begin{minipage}[t]{0.32\linewidth}
\begin{modarespbox}{Role 3}
forest-dawn-mist-breeze-echo
\end{modarespbox}
\end{minipage}

\vspace{6pt}

\begin{minipage}[t]{0.32\linewidth}
\begin{modarespbox}{Role 4}
Sunny-Daylight-Forest-People-Apple
\end{modarespbox}
\end{minipage}\hfill%
\begin{minipage}[t]{0.32\linewidth}
\begin{modarespbox}{Role 5}
endless-sunrise-joy-dreams-boundless
\end{modarespbox}
\end{minipage}\hfill%
\begin{minipage}[t]{0.32\linewidth}
\begin{modarespbox}{Role 6}
whispering-pine-breeze-sunrise-moonlight
\end{modarespbox}
\end{minipage}

\vspace{6pt}
\caption{\textbf{Qualitative example of diverse generation: \textit{Generate a 5 word passphrase separated by hyphens.}} Upper: Responses sampled from Qwen3-8B. Bottom: Responses sampled from \method{} under distinct numbered roles for the same prompt, illustrating conceptual diversity across generations.}

\label{fig:qualitative-example-passphrase}
\vspace{-0.4cm}
\end{figure}

%% file: notes_arxiv/appendix/generation_cocktail.tex
\label{appx:generation-example-cocktail}

\begin{figure}[t]
\centering

\begin{promptbox}
\textit{Name one cocktail I can make with rum.}
\end{promptbox}

\begin{minipage}[t]{0.32\linewidth}
\begin{respbox}{Sample 1}
One classic cocktail you can make with rum is the \textbf{Piña Colada}. Here's a simple recipe:

\textbf{Piña Colada} \\
2 oz white rum; 1.5 oz pineapple juice; 0.5 oz coconut cream; ice. Optional: pineapple slice or cherry for garnish.

Mix all ingredients in a blender until smooth, pour into a glass, and enjoy!
\end{respbox}
\end{minipage}\hfill%
\begin{minipage}[t]{0.32\linewidth}
\begin{respbox}{Sample 2}
One cocktail you can make with rum is the \textbf{Piña Colada}.

\textbf{Piña Colada Recipe:}

\textbf{Ingredients:} 1 cup pineapple juice; 3/4 cup coconut milk; 1.5 oz (45 ml) light rum; ice cubes.

\textbf{Instructions:} In a blender, combine pineapple juice, coconut milk, and rum. Add a handful of ice cubes. Blend until smooth and creamy. Serve in a chilled glass, optionally garnished with a pineapple slice or a cherry.

Enjoy your tropical drink!
\end{respbox}
\end{minipage}\hfill%
\begin{minipage}[t]{0.32\linewidth}
\begin{respbox}{Sample 3}
One classic cocktail you can make with rum is the \textbf{Piña Colada}. It's a popular tropical drink that's refreshing and perfect for warm weather.

\textbf{Ingredients:} 1.5 oz light rum; 1.5 oz pineapple juice; 3 oz coconut cream; ice. Optional: a splash of orange juice for extra flavor.

\textbf{Instructions:} Fill a blender with ice. Add the rum, pineapple juice, and coconut cream. Blend until smooth. Pour into a glass and enjoy!

Let me know if you'd like a variation or a non-alcoholic version!
\end{respbox}
\end{minipage}

\vspace{6pt}

\begin{minipage}[t]{0.32\linewidth}
\begin{respbox}{Sample 4}
A classic cocktail you can make with rum is the \textbf{Margarita}.

Here's a simple recipe:

\textbf{Margarita} \\
\textbf{Ingredients:} 2 oz (60 ml) tequila; 1 oz (30 ml) triple sec or orange liqueur; 0.5 oz (15 ml) freshly squeezed lime juice; salt (for rimming the glass, optional).

\textbf{Instructions:} Rim a glass with salt (optional). In a shaker, combine tequila, triple sec, and lime juice. Shake well with ice. Strain into the prepared glass. Serve with a lime wedge as garnish.

Enjoy your Margarita!
\end{respbox}
\end{minipage}\hfill%
\begin{minipage}[t]{0.32\linewidth}
\begin{respbox}{Sample 5}
Sure! One classic cocktail you can make with rum is the \textbf{Margarita}. Here's a simple recipe:

\textbf{Margarita} \\
2 oz (60 ml) tequila (or light rum); 1 oz (30 ml) triple sec or orange liqueur; 0.5 oz (15 ml) lime juice; 0.5 oz (15 ml) fresh lime juice (for a sour version); salt for rimming the glass (optional).

\textbf{Instructions:} Rim a glass with salt (optional). Fill the glass with ice. Add the tequila, orange liqueur, and lime juice. Stir well. Garnish with a lime wedge, and enjoy!

If you're using \textbf{rum} instead of tequila, you can make a \textbf{Rum Margarita} with similar ingredients.
\end{respbox}
\end{minipage}\hfill%
\begin{minipage}[t]{0.32\linewidth}
\begin{respbox}{Sample 6}
A classic cocktail you can make with rum is the \textbf{Margarita}. Here's a simple recipe:

\textbf{Ingredients:} 2 oz (60 ml) tequila; 1 oz (30 ml) triple sec or orange liqueur; 0.5 oz (15 ml) fresh lime juice; salt (for rimming the glass, optional); lime wedge (for garnish).

\textbf{Instructions:} Rim a glass with salt (optional). In a shaker, combine tequila, triple sec, and lime juice. Add ice, and shake well. Strain into the prepared glass. Garnish with a lime wedge and enjoy!

Let me know if you'd like a non-alcoholic version or another recipe!
\end{respbox}
\end{minipage}

\begin{minipage}[t]{0.32\linewidth}
\begin{modarespbox}{Role 1}
Sure! Here's a simple and classic cocktail you can make with rum:

\textbf{The Mojito}

\textbf{Ingredients:} 2 oz light rum (such as Havana Club or Bacardi); 1.5 oz fresh lime juice; 0.5 oz fresh lemon juice; 1.5 oz simple syrup (or sugar + water); a handful of fresh mint leaves (about 6--8); soda water; ice cubes.

\textbf{Instructions:} In a tall glass, muddle the mint leaves with the lime and lemon juices and simple syrup until the mint is slightly crushed and the flavors are released. Add a handful of ice cubes. Pour in the rum. Top with more ice and club soda. Garnish with a lime wedge or a sprig of mint.

Enjoy your refreshing Mojito!
\end{modarespbox}
\end{minipage}\hfill%
\begin{minipage}[t]{0.32\linewidth}
\begin{modarespbox}{Role 2}
Ah, the sweet nectar of rum---oh, how I've danced with it in the moonlight of many a bar. Let me conjure a drink that's as smooth as a whispered secret and as bold as a sailor's tale.

\textbf{The ``Jungle Drifter''}

1.5 oz dark rum; 0.5 oz pineapple juice; 0.5 oz lime juice; 0.5 oz simple syrup; 1 dash Angostura bitters. Garnish: a slice of pineapple and a sprig of mint.

\textit{Method:} Shake all ingredients with ice, then strain into a chilled coupe glass. Sip slowly, and let the tropical whispers of the Caribbean take you away.

What's your preferred flavor profile tonight? I'd love to tailor the recipe for you.
\end{modarespbox}
\end{minipage}\hfill%
\begin{minipage}[t]{0.32\linewidth}
\begin{modarespbox}{Role 3}
Here's a cocktail recipe you can make with rum, served with a touch of whimsy:

\textbf{The Midnight Mallard}

\textbf{Ingredients:} 2 oz dark rum; 1 oz passion fruit juice; 1 oz pineapple juice; 0.5 oz lime juice; 0.5 oz coconut nectar (or honey); 1 dash of Angostura bitters. Garnish: a pineapple wedge with a cherry on top (or a glowing neon feather, if you dare).

\textbf{Instructions:} In a shaker with ice, combine all the ingredients. Shake well until chilled. Strain into a chilled coupe or rocks glass. Garnish with your chosen embellishment---light or surreal, it's your cocktail.

Sip as if you're sipping on the edge of a moonlit lagoon, where the stars are made of rum and the tides are sweet.
\end{modarespbox}
\end{minipage}

\vspace{6pt}

\begin{minipage}[t]{0.32\linewidth}
\begin{modarespbox}{Role 4}
\textbf{Cocktail Suggestion: ``Sunset Serenade''}

\textit{Why it fits:} This cocktail blends structured precision (distilled rum as the base) with playful creativity (unexpected ingredients like elderflower syrup and hibiscus bitters).

\textbf{Ingredients:} 1.5 oz white rum; 0.5 oz elderflower syrup; 0.25 oz hibiscus bitters; 0.5 oz fresh lime juice; 1 splash orange bitters. Garnish: a sprig of lavender.

\textbf{Instructions:} Fill a shaker with ice, then add the rum, elderflower syrup, hibiscus bitters, lime juice, and orange bitters. Shake vigorously, like a scientist eager to uncover the unknown. Strain into a chilled coupe glass. Garnish with a sprig of lavender.

This cocktail is not just a drink---it's an experience. What happens when floral and bold meet?
\end{modarespbox}
\end{minipage}\hfill%
\begin{minipage}[t]{0.32\linewidth}
\begin{modarespbox}{Role 5}
Why not try the \textbf{``Phantom of the Caribbean''}?

\textbf{Ingredients:} 2 oz dark rum; 1 oz coconut cream; 1 oz pineapple juice; 0.5 oz lime juice; 0.5 oz simple syrup. Garnish: a slice of pineapple and a sprig of mint.

\textbf{Method:} Shake all ingredients (except garnish) with ice in a cocktail shaker. Strain into a rocks glass filled with a single ice cube. Garnish with pineapple and mint.

\textbf{Storytime:} This cocktail is inspired by the legend of a ghost ship that sails the Caribbean at midnight, leaving only the scent of rum and the echo of laughter. Some say it's a pirate's farewell, others claim it's a trick played by the moon. Either way, it's best enjoyed with a sense of adventure---and a stiff drink.
\end{modarespbox}
\end{minipage}\hfill%
\begin{minipage}[t]{0.32\linewidth}
\begin{modarespbox}{Role 6}
World: Oasis Prime

``You've entered the realm of Legacy Clients. Your request carries the weight of centuries. In the Grand Library of Oasis Prime, the Codex of Beverages is vast. Let me recall\ldots Rum is the spirit of the tropics, born of the sun and the sea. A drink that celebrates adventure and rebellion.

\textbf{The Righteous Rum Runner}, a cocktail of defiance and delight.

\textbf{Ingredients:} 2 oz dark rum; 1 oz fresh lime juice; 1/2 oz simple syrup; 1/2 oz aged rum (for depth); 1 dash of Angostura bitters. Garnish: orange twist.

\textbf{Method:} In a shaker, combine all ingredients. Shake with ice until well chilled. Strain into a chilled coupe glass. Garnish with an orange twist.

This cocktail is a tribute to those who dared to sail beyond the horizon.''

--- The Archivist of Oasis Prime
\end{modarespbox}
\end{minipage}

\vspace{6pt}
\caption{\textbf{Qualitative example of diverse generation: \textit{Give a random chess move in response to 1. b3.}} Upper: Responses sampled from Qwen3-8B. Bottom: Responses sampled from \method{} under distinct numbered roles for the same prompt, illustrating conceptual diversity across generations.}

\label{fig:qualitative-example-cocktail}
\vspace{-0.4cm}
\end{figure}